\documentclass{article}

\PassOptionsToPackage{numbers}{natbib}

\usepackage[final]{neurips_data_2023}
\usepackage{booktabs}
\usepackage{graphicx}
\usepackage{bold-extra}
\usepackage{adjustbox}
\usepackage{enumitem}   
\usepackage[most]{tcolorbox}
\usepackage[frozencache,cachedir=.]{minted}
\tcbuselibrary{minted, listings, breakable, xparse, skins}
\usepackage{xcolor}         
\usepackage{tabularx}
\usepackage{wrapfig}
\usepackage{amssymb, multirow, svg, boldline, bbm, makecell, adjustbox}
\usepackage{amsthm}
\usepackage{colortbl}
\usepackage{xcolor, soul}
\sethlcolor{teal}
\colorlet{BoxColor}{blue!10!}
\tcbset{on line, 
        boxsep=4pt, left=2pt,right=2pt,top=1pt,bottom=1pt,
        colframe=black, colback=BoxColor,  
        highlight math style={enhanced}
        }

\usepackage[utf8]{inputenc} 
\usepackage[T1]{fontenc}    
\usepackage{hyperref}       
\usepackage{url}            
\usepackage{booktabs}       
\usepackage{amsfonts}       
\usepackage{nicefrac}       
\usepackage{microtype}      

\usepackage{todonotes}

\title{\textit{AQuA}: A Benchmarking Tool for Label Quality Assessment}

\author{%
  Mononito Goswami$^{*}$, Vedant Sanil\thanks{MG and VS contributed equally. MG is the corresponding author.}, 
  Arjun Choudhry$^{\dagger}$,\\ \textbf{Arvind Srinivasan}\thanks{AC and AS have equal contribution.}, 
  \textbf{Chalisa Udompanyawit}, \textbf{Artur Dubrawski} \\
  Auton Lab, School of Computer Science\\
  Carnegie Mellon University\\
  \texttt{\{mgoswami, vsanil, arjuncho, arvindsr, cudompan, awd\}@cs.cmu.edu} \\
  \url{www.github.com/autonlab/aqua}
}

\definecolor{bg}{gray}{0.95}
\DeclareTCBListing{mintedbox}{O{}m!O{}}{%
  breakable=true,
  listing engine=minted,
  listing only,
  minted language=#2,
  minted style=default,
  minted options={%
    gobble=0,
    breaklines=true,
    breakafter=,,
    fontsize=\scriptsize,
    numbersep=0pt,
    #1},
  boxsep=0pt,
  left skip=0pt,
  right skip=0pt,
  left=0pt,
  right=0pt,
  top=0pt,
  bottom=0pt,
  arc=5pt,
  leftrule=0pt,
  rightrule=0pt,
  bottomrule=0pt,
  toprule=1pt,
  colback=bg,
  colframe=blue!20!, 
  enhanced,
  #3}

\begin{document}

\maketitle

\begin{abstract}
\label{sec:abstract}
Machine learning (ML) models are only as good as the data they are trained on. But recent studies have found datasets widely used to train and evaluate ML models, e.g. \textit{ImageNet}, to have pervasive labeling errors. Erroneous labels on the train set hurt ML models' ability to generalize, and they impact evaluation and model selection using the test set. Consequently, learning in the presence of labeling errors is an active area of research, yet this field lacks a comprehensive benchmark to evaluate these methods. Most of these methods are evaluated on a few computer vision datasets with significant variance in the experimental protocols. With such a large pool of methods and inconsistent evaluation, it is also unclear how ML practitioners can choose the right models to assess label quality in their data. To this end, we propose a benchmarking environment \texttt{AQuA} to rigorously evaluate methods that enable machine learning in the presence of label noise. We also introduce a design space to delineate concrete design choices of label error detection models. We hope that our proposed design space and benchmark enable practitioners to choose the right tools to improve their label quality and that our benchmark enables objective and rigorous evaluation of machine learning tools facing mislabeled data.
\end{abstract}

\section{Introduction}
\begin{figure}[!bt]
    \centering
    \includegraphics[width=\textwidth]{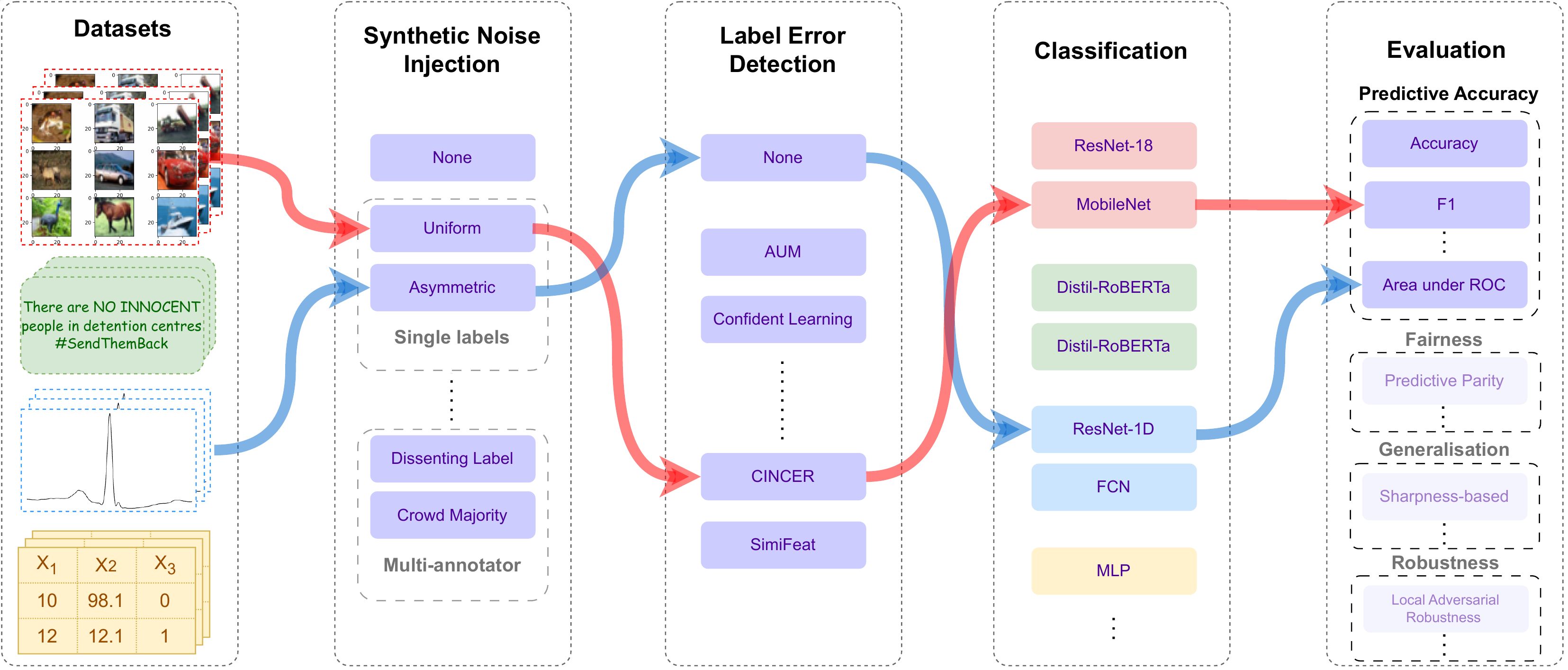}
    \caption{\textit{Overview of the \texttt{AQuA} benchmark framework}. \texttt{AQuA} comprises of datasets from \textbf{4} modalities, \textbf{4} single-label and \textbf{3} multi-annotator label noise injection methods, \textbf{4} state-of-the-art label error detection models, classification models, and several evaluation metrics beyond metrics of predictive accuracy. We are in the process of integrating several fairness, generalization, and robustness metrics into \texttt{AQuA}. The \textcolor{red}{red} and \textcolor{blue}{blue} arrows show two example experimental pipelines for image data and time-series data, respectively.}
    \label{fig:benchmark_overview}
\end{figure}

A lot of machine learning (ML) research is devoted to making efficient and effective use of available data to learn accurate, high-fidelity, and interpretable models, with little to no focus on the quality of the data they are trained and evaluated on. Nonetheless, it is widely recognized that ML models are only as good as the data they rely on, i.e., the quality of data imposes practical limits to what ML models can achieve. Not only are datasets used to train ML models; they also serve as benchmarks to measure the state-of-the-art and validate theoretical findings. Thus, high quality large labeled datasets are the cornerstone of progress in supervised machine learning. However, the data is rarely free of noise, which can both manifest in the features of the data (feature noise) and in labels that categorize them (label noise). Between feature and label noise, the former has been found to be much more harmful to machine learning models \cite{frenay2013classification, zhu2004class, saez2014analyzing}. To make matters worse, label noise is prevalent in popular ML benchmarks. A recent study estimated an average of at least 3.3\% label errors across 10 datasets commonly used for benchmarking computer vision, natural language, and audio classification algorithms \cite{northcutt2021labelerrors}. 
Consequently, a growing body of research is devoted to understanding the harms of label noise and to developing techniques to identify and mitigate labeling errors.

In recent years, over \textbf{50} papers have been written on this topic, including \textbf{6} surveys, yet the literature lacks a comprehensive benchmark to evaluate the available methods. The evaluation of existing methods is lacking along the following dimensions: 

\textbf{Arbitrary choice of datasets and limited data modalities.} To the best of our knowledge, relevant studies have used over \textbf{40} datasets (e.g., ImageNet \cite{ILSVRC15}) and their variations (e.g., Imagenette \cite{imagenette}, ImageNet-100 \cite{gao2022learning}) for evaluation, but mostly on computer vision related tasks, with less than \textbf{15} studies using text data, \textbf{7} using tabular data and only \textbf{1} paper using time-series data. 

\textbf{Arbitrary choice of classification models.} The ultimate goal of identifying labeling errors is to learn a classification model using training data with clean labels. Much like the datasets, relevant studies have used over \textbf{47} different classification architectures (e.g., ResNet~\cite{resnet}, MobileNet~\cite{mobilenet}, ResNeXt~\cite{resnext}, BERT~\cite{bert}, XLM-RoBERTa~\cite{xlm_roberta}, etc) to measure the impact of label cleaning.

\textbf{Inconsistent evaluation protocols and metrics.} Different studies conduct different experiments to measure the efficacy of their proposed methods (e.g., the accuracy of the label cleaning method, or performance of the downstream model before and after label cleaning, etc.) and use various measures of success (e.g., high accuracy, $F_1$-score, or low error rate).

With such diversity and inconsistency in the way in which these methods are evaluated, it is hard to measure the state of the art. To bridge this gap, we propose the \underline{A}nnotation \underline{Qu}ality \underline{A}ssessment, \texttt{AQuA}, the \emph{first} benchmark framework to evaluate machine learning methods in the presence of label noise (Fig.~\ref{fig:benchmark_overview}). We also elucidate the design space for such models, with the hope that it will not only foster future research on detecting labeling errors, but also enable ML practitioners to choose the appropriate label cleaning tools for their specific data and tasks. We run a large-scale experiment (\textbf{> 1000} unique experiments) and make several interesting observations, demonstrating \texttt{AQuA}'s efficacy in benchmarking machine learning models in the presence of label noise.

\section{Background and Problem Formulation}

\begin{wrapfigure}[12]{R}{0.5\textwidth}
\begin{minipage}{0.5\textwidth}
    \centering
    \includegraphics[width=\textwidth]{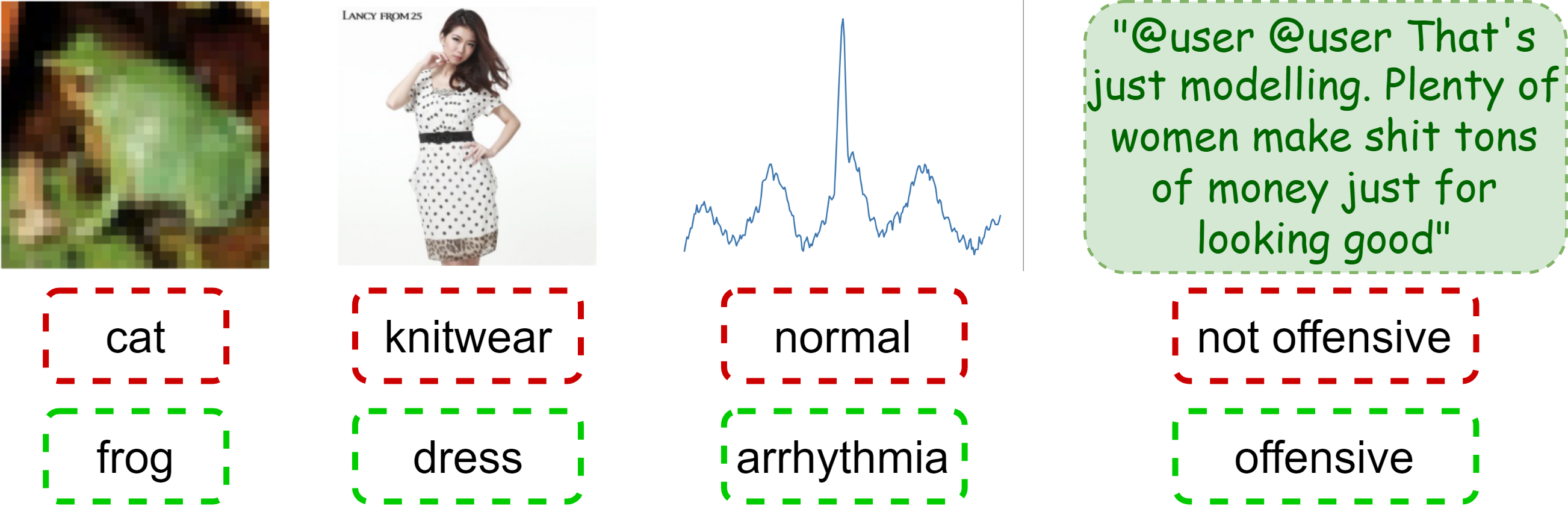}
    \caption{Labeling errors in widely used benchmarks: CIFAR-10, Clothing-100K, MIT-BIH, and TweetEval Hate Speech datasets. Observed labels are in \textcolor{red}{\texttt{red}} and true labels are in \textcolor{green}{\texttt{green}}.}
    \label{fig:enter-label}
\end{minipage}
\end{wrapfigure}

\textbf{Sources of labeling errors.} Labeling errors can arise from automated labeling processes such as crowd-sourcing \cite{yuen2011survey}, programmatic weak supervision \cite{zhang2022survey, goswami2021weak}, and human error (e.g., due to lack of expertise or low confidence in expert assessment)~\cite{peterson2019human}\footnote{The root cause of labeling errors in crowd-sourcing is different from human expert annotation. For instance, errors during crowd-sourcing have been shown to arise from other factors such as gaming the system to maximize monetary gains \citep{yuen2011survey}.}. Errors may also stem from idiosyncrasies of the annotation procedure and the corresponding guidelines themselves~\cite{beyer2020we}. Finally, existing
labels may also become inconsistent with prevailing knowledge due to constantly evolving problem definitions
and domain knowledge leading to concept drift\footnote{For example, sepsis is one of the most sought-after clinical conditions to predict. However, with the constantly evolving definition of sepsis, the labeling process is frequently affected, causing many annotations in legacy benchmark data to become inconsistent with the latest guidelines~\cite{giacobbe2021early}, a very dangerous risk to take in the particular type of application area}. 

\textbf{Impact of labeling errors.} At training time, labeling errors can cripple an ML model's ability to generalize and
introduce undesirable biases in its hypothesis space~\cite{arpit2017closer, zhang2021understanding}. Mislabeled training data is especially problematic for over-parameterized deep neural networks, which can achieve zero training error even on randomly-assigned labels \cite{zhang2021understanding}. At test time, labeling errors can lead to noisy model evaluations
and invalidate common model selection strategies. In safety-critical settings, models trained, evaluated, and selected using mislabeled data can be ineffective at best and can lead to disastrous outcomes at worst. Finally, recent studies in the context of fairness have shown that naively enforcing parity constraints based on noisy labels can harm groups that are unaffected by label noise~\cite{wang2021fair, wu2022fair}.

\textbf{Problem formulation.} Due to the far-reaching consequences that labeling errors can have on model training and evaluation, the literature has attacked multiple different but related problems, for example:
(1) \textit{label error detection}, identify which data points have erroneous labels~\cite{northcutt2021confident, aum_ranking}, 
(2) \textit{label noise estimation}, estimate the proportion of data with noisy labels~\cite{northcutt2017rankpruning}, 
(3) \textit{label noise robust learning}, learn models robust to label noise~\cite{reed2015training, patrini2017making}, and 
(4) \textit{noise transition matrix estimation}, estimate the parameters of the noisy label generation process~\cite{xia2019anchor}. 

In this work, we focus on the \textbf{label error detection problem}, because (a) it is the most \textit{general} of the above problem types, i.e., with knowledge of labeling errors, we can estimate the noise rate, parameters of the noise generation process and train ML models free from label noise, (b) it provides practitioners greater visibility of issues that plague their data, and (c) allows them to directly rectify these errors.

\begin{tcolorbox} 
\small \textbf{Label error detection problem:} Assume a dataset $\mathcal{D}^* = \{(\mathbf{x}_i, y_i^*)\}_{i=1}^N \in (\mathcal{X}, \mathcal{Y})$, where $\mathbf{x}_i$ and $y_i^*$ denote the features and labels, respectively. In practice, we do not have access to $\mathcal{D}^*$, but instead observe a noisy dataset $\mathcal{D} = \{(\mathbf{x}_i, y_i)\}_{i=1}^N \in (\mathcal{X}, \mathcal{Y})$\footnote{We assume that we observe the true features since we are interested in identifying labeling errors and isolating their impact on downstream model performance.}. We call $y_i$ a \emph{labeling error}\footnote{\textit{A note on terminology:} In this paper, we will sometimes refer to \textit{labeling errors} as \textit{noisy labels} or \textit{label noise}, and the process of identifying them as \textit{label error detection}, or loosely as \textit{label cleaning}.} if $y_i \neq y_i^*$, and \textit{correctly labeled}, otherwise. Our goal is to identify all labeling errors in $\mathcal{D}$. 
\end{tcolorbox} 

\section{A Design Space of Labeling Error Detection Models}
\label{sec:designspace}

\begin{wrapfigure}[20]{R}{0.5\textwidth}
\begin{minipage}{0.5\textwidth}
    \centering
    \includegraphics[width=\textwidth]{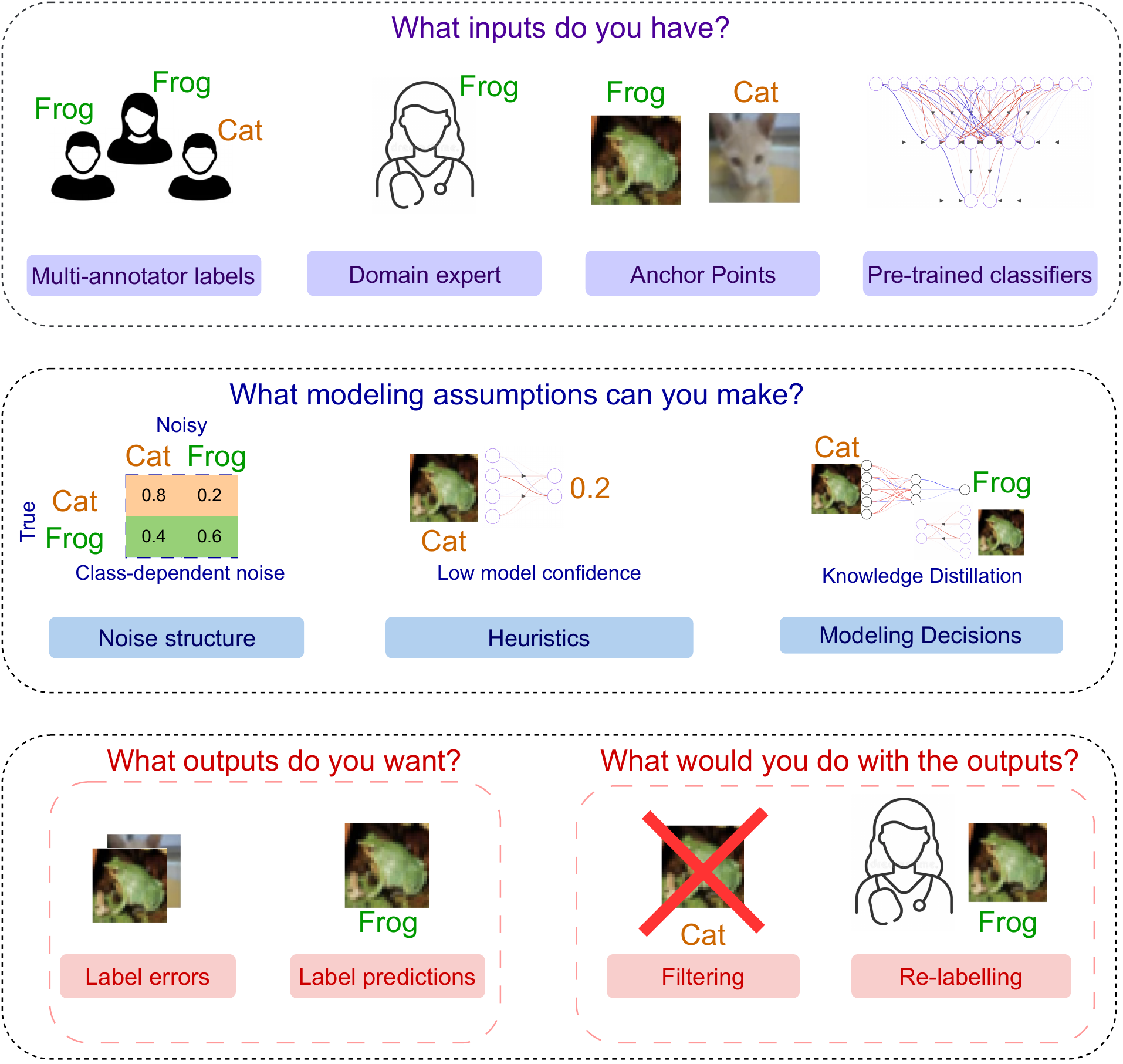}
    \caption{\textit{Design space of labeling error detection models to delineate concrete design choices.} }
    \label{fig:flow_chart}
\end{minipage}
\end{wrapfigure}

In this section, we seek to align the dimensions along which label error detection models vary, with dimensions that can facilitate model selection for ML practitioners. We provide a brief overview of these dimensions below and defer detailed discussions to Appendix~\ref{app:assumptions}.
 
\textbf{What inputs do you have?} All label error detection models take features and noisy labels as input. In most datasets, data points are labeled by multiple experts, but their individual annotations are seldom available. When available, \textit{multi-annotator labels} can be used to identify data points that are inherently ambiguous \cite{active_label_cleaning}, or to model individual annotators to estimate their expertise and propensity for mislabeling examples \cite{gao2022learning}, and using these to identify likely labeling errors. While most methods identify labeling errors and automatically remove or correct them, a few rely on a \textit{human expert} who can be queried to relabel suspicious data points \cite{active_label_cleaning, cincer}. Some other methods assume access to data points called \textit{anchor points}, which most certainly belong to a particular class \cite{classification_by_importance_reweighting, mixture_proportion_estimation}. The number of anchor points required is generally proportional to the number of classes, and quickly becomes prohibitive for multi-class classification problems, and in more complicated noise settings \cite{clusterability_alternative_anchor_points}. Finally, a vast majority of methods assume \textit{access to classification models}, and primarily differ in their \textit{number} (model-free \cite{zhu2022detecting}, one or multiple models \cite{mentornet, coteaching, learn_from_noisy_labeled_data_meta_learning, meta_learning_correction}), \textit{nature of access} (prediction-only \cite{northcutt2021confident} versus access to logits \cite{aum_ranking}, gradients \cite{cincer} etc.), and \textit{extent of pre-training} (no pre-training \cite{northcutt2021confident, aum_ranking} versus large-scale pre-training e.g. large language models \cite{pretrained_lm_2022}). 

\textbf{What modeling assumptions can you make?} Different studies use different assumptions on \textit{data} (noise structure and clusterability), \textit{heuristics} (model self-confidence and perceptual uncertainty), and \textit{modeling decisions} (whether to explicitly model the transition matrix and multi-network training). Most studies in the literature explicitly assume some form of structure in the noise present in the data \cite{northcutt2021confident, zhu2022detecting, robustness_in_deep_discriminative_NN, xia2019anchor}. Most early studies assumed class-dependent noise, i.e., the likelihood of error is only dependent on the latent true class, not on the data \cite{northcutt2021confident, xia2019anchor}. There is growing interest in more realistic forms of noise where the probability of error also depends on the features of a data point (instance-dependent noise) \cite{part_dependent_label_noise, sample_sieve_cores}. To this end, some recent studies have shown promising results by leveraging natural notions of similarity between data points and their labels. For example, \citet{zhu2022detecting} assume that examples with similar features should have similar labels. 

Many studies treat a trained model's low confidence that a data point belongs to its observed label as a heuristic likelihood to identify labeling errors \cite{aum_ranking, northcutt2021confident, label_noise_in_ultrasound_prostate_biopsy}. In a similar vein, a recent study used the loss of a pre-trained large language model on each data point to identify mislabeled examples \cite{pretrained_lm_2022}. When multi-annotator labels are available, as discussed before, some studies have also used them to model the perceptual uncertainty in the annotators to identify labeling errors. 

Finally, studies differ in their modeling decisions. While some explicitly estimate a data structure called the noise transition matrix, which encodes the joint probability of latent true and observed noisy labels \cite{northcutt2021confident, clusterability_alternative_anchor_points, patrini2017making}, others do not \cite{aum_ranking, cincer, zhang2022survey}. Finally, there is a body of work on label noise robust learning using multiple model instances either using knowledge distillation \cite{mentornet, coteaching, coteaching+} or meta-learning \cite{meta_learning_correction, learn_from_noisy_labeled_data_meta_learning}. The key idea is to use a cooperative game between models to identify labeling errors and ensure that the eventually deployed model only learns from clean data.

\textbf{What outputs do you want, and what would you do with them?} All labeling error detection models identify data points that are likely to be labeling errors. With knowledge of the potentially mislabeled data points, most studies simply remove them from consideration\cite{northcutt2021confident, aum_ranking, deep_knn, sample_sieve_cores}. This strategy may be practical for large datasets, where only a small fraction of data is found to be mislabeled and domain experts are unavailable for supervision. We use this strategy by default in \texttt{AQuA}. A smaller number of methods predict the alternate class that the data point is most likely to belong to \cite{meta_learning_correction, zhu2022detecting} and even provide explanations for their predictions \cite{cincer, label_noise_in_context}. CINCER \cite{cincer} is one of the few methods which not only finds labeling errors but also identifies counter-examples in the training data to serve as explanations for its suspicion. Some studies use the label predicted by these models and perform loss re-weighting or correction to learn robust classification models \cite{patrini2017making, robust_loss_functions_2017, combating_noise_using_abstention}. When domain experts are available, some studies also leverage their insight to re-label mislabeled data point~\cite{cincer, active_label_cleaning}.

\section{Benchmark Design}
\label{sec:benchmarkdesign}
\subsection{Real-world, Popular Datasets, and Downstream Classification Models}
\begin{table}[!tb]
\centering
\resizebox{\columnwidth}{!}{
\begin{tabular}{l|lcccccc}
\Xhline{1pt}
\textbf{Modality} & \textbf{Dataset} & \textbf{\# Train / Test} & \textbf{\# Annotators/sample} & \textbf{Label Source} & \textbf{Classification Task} & \textbf{Sample Size} & \textbf{Usage}\\ \midrule
    
    \multirow{4}{*}{Image} & CIFAR-10N\cite{cifar_10n} & 50K / 10K & 3 & Human annotation & Object & $32 \times 32 \times 3$ & \cite{Model_agnostic_cleanlab, zhu2022detecting}\\
    
    & CIFAR-10H\cite{peterson2019human} & 0 / 10K & 47--63 & Human annotation & Object & $32 \times 32 \times 3$ & \cite{active_label_cleaning}\\
    
    & Clothing100K\cite{Clothing1M} & 100K & 1 & Web-labeled & Image & $256 \times 256 \times 3$ & \cite{aum_ranking, northcutt2021labelerrors, zhu2022detecting}\\

    & NoisyCXR\cite{noisycxr} & 26K / 3K & 1--XX & Human expert annotation & Pneumonia & $1024 \times 1024 \times 1$ & \cite{active_label_cleaning} \\ \midrule

    \multirow{2}{*}{Text} & IMDb$^\beta$\cite{IMDB} & 25K / 25K & 1 & Human annotation & Sentiment & - &  \cite{patrini2017making, robust_loss_functions_2017, northcutt2021labelerrors}\\

    & TweetEval\cite{tweeteval} & 10K & 1 & Human annotation & Hate speech & - & - \\ \midrule

    \multirow{6}{*}{Tabular} & Credit Card Fraud$^\beta$\cite{credit_card_dataset} & 284K & 1 & Human annotation & Credit card fraud & $28$ & \cite{clustering_training_losses, reconstruction_error_framework} \\

    & Adult$^\beta$\cite{uci} & 48K & 1 & Rule-based extraction & Salary & $14$ & \cite{cincer, wang2021fair, wu2022fair} \\

    & Dry Bean\cite{dry_bean_dataset} & 13K & 1 & Vision system-based annotation & Bean variety & $17$ & - \\

    & Car Evaluation\cite{hierarchical_decision_model} & 1K & 1 & Hierarchical decision model \cite{hierarchical_decision_model} & Car condition & $6$ & \cite{bootstrap_latent}\\

    & Mushroom$^\beta$\cite{mushroom_dataset} & 8K & 1 & - & Mushroom edibility & $22$ & \cite{clustering_training_losses}\\

    & COMPAS$^\beta$\cite{COMPAS_dataset} & 6K & 1 & - & Recidivism & $28$ & \cite{wang2021fair} \\ \midrule

    \multirow{5}{*}{\begin{tabular}{@{}c@{}}Time \\ Series \end{tabular}} & Crop\cite{dynamic_time_warping} & 7K / 16K & 1 & 
    \begin{tabular}{@{}c@{}}Hierarchical k-means tree \\ with dynamic time warping \cite{dynamic_time_warping} \end{tabular}
    & Crop cover & $46 \times 1$ & - \\

    & ElectricDevices\cite{electric_devices_dataset} & 9K / 7K & 1 & Human annotation & Appliance-type & $96 \times 1$ & - \\

    & MIT-BIH\cite{mit_bih} & 23K / 4K & 1 & Human expert annotation & Arrhythmia & $256 \times 2$ & - \\

    & PenDigits\cite{pendigits_dataset} & 7K / 3K & 1 & Human annotations & Handwritten digit & $16 \times 1$ & - \\

    & WhaleCalls$^\beta$\cite{UEA} & 11K / 2K & 1 & - & Whale call & $4,000 \times 1$ & - \\

\Xhline{1pt}
\end{tabular}}
\caption{\textbf{Summary of datasets.} \texttt{AQuA} currently includes a variety of datasets for different classification problems, varying in the number of classes, sources of annotations, and data modalities. All datasets except those marked with $\beta$ are multi-class.
}
\label{tab:aqua_datasets}
\end{table}

\textbf{Datasets.} \texttt{AQuA} currently comprises of a collection of \textbf{17} popular real-world public datasets from \textbf{4} prevalent data modalities: \textit{image}, \textit{text}, \textit{time-series} and \textit{tabular}. To evaluate label error detection models across various practical scenarios, we carefully choose datasets with diversity in the following characteristics: (1) \textit{classification problems} (\textit{e.g.}, sentiment classification vs.\ hate speech detection), (2) \textit{number of classes} (binary vs multi-class classification), (3) \textit{relative prevalence of classes} (e.g., skewed datasets like Credit Card Fraud~\cite{credit_card_dataset} and balanced ones like IMDb~\cite{IMDB}), (4) \textit{sources of annotations} (e.g., human vs rule-based annotation), and (5) \textit{number of annotations per example} (e.g., CIFAR-10N labeled by 3 annotators). Table~\ref{tab:aqua_datasets}  summarizes the key characteristics of datasets included as a part of \texttt{AQuA}. In particular, to make comparison with prior work easier while maintaining diversity across practical scenarios, we try to include datasets that have been used frequently by prior work (see usage in Table~\ref{tab:aqua_datasets}) and preprocess them in a manner consistent with those works. We do not use any data augmentation during training. App~\ref{app:datasets} provides detailed descriptions of the datasets.

\textbf{Classification models.} The ultimate goal of label cleaning is to train accurate downstream classifiers, but different studies use different classification models to measure the efficacy of their proposed label cleaning methods. To provide a level playing field for all cleaning methods, we include multiple classification model architectures for each data modality. Specifically, we include ResNet-18 \cite{resnet}, MobileNet \cite{mobilenet} and FastViT-T8 \cite{fastvit} for image datasets, \texttt{all-distilroberta-v1} \cite{roberta, distilbert} and \texttt{all-MiniLM-L6-v2} \cite{minilm} for text datasets, ResNet-1D, PatchTST \cite{PatchTST} and LSTM Fully Convolutional Network \cite{FCN} for time-series datasets, and TabTransformer \cite{tabtransformer} and a Multi-Layer Perceptron for tabular datasets. While choosing classification models we prioritized \textit{performant} methods with (1) \textit{different architectures} and \textit{inductive biases}, (2) ideally \textit{pre-trained} using different strategies, and (3) \textit{previously-used} either by label cleaning methods or task-relevant papers. App.~\ref{app:classification_models} and App.~\ref{app:hyperparameters} provide a detailed description of classification models and their hyperparameters, respectively. 

\subsection{Advanced Label Error Detection Methods} 
\begin{wrapfigure}[18]{R}{0.5\textwidth}
\begin{minipage}{0.5\textwidth}
    \centering
    \begin{mintedbox}{python}
    from aqua.models import TrainAqModel, ConvNet
    from aqua.data import Aqdata, load_cifar
    from aqua.reports import generate_report
    
    # Load CIFAR-10 and ResNet-18
    clf = ConvNet('resnet18')
    data = load_cifar() 
    data.add_noise(noise_rate=0.2) # Add uniform noise
    
    # Instantiate a cleaning method and classifier
    cleaner = TrainAqModel(clf, method='aum')
    label_errors = cleaner.find_label_issues(data)

    # Remove data with label issues
    data.clean_data(label_issues) 
    # Train a downstream model on cleaned data
    y_preds = TrainAqModel(clf).fit_predict(data)
    \end{mintedbox}
    \caption{\texttt{AQuA} makes identifying label issues, and evaluating new and existing label error detection models simple.}
    \label{fig:api}
\end{minipage}
\end{wrapfigure}
\texttt{AQuA} provides easy-to-use Application Programmer Interfaces (Fig.~\ref{fig:api}) for \textbf{4} state-of-the-art label error detection methods, namely Area Under Margin ranking (AUM)~\cite{aum_ranking}, Confident Learning~\cite{northcutt2021confident}, Contrastive and Influent Counter Example Strategy (CINCER)~\cite{cincer}, and Model-free Label Error Detection (SimiFeat)~\cite{zhu2022detecting}. Below, we provide a brief overview of these methods and their key ideas.

\textbf{\underline{A}rea \underline{U}nder the \underline{M}argin Ranking (\texttt{AUM}) \cite{aum_ranking}.} Given noisy data and access to the logits of a deep learning model, AUM exploits differences in training dynamics of clean and mislabeled samples to identify labeling errors. The key idea is to identify data points that do not contribute to the generalization of a model as labeling errors by leveraging the delicate tension between the label of a data point (via memorization) and its predicted label (via gradient updates), measured as the margin between the logits of a sample's assigned class and its highest unassigned class. 

\textbf{\underline{Con}fident Learning (\texttt{CON}) \cite{northcutt2021confident}.} Given noisy data, confident learning estimates a data structure called \textit{confident joint}, which is the joint probability distribution of observed noisy and latent true labels. The key idea is to leverage a model trained on held-out data drawn from the same (or similar) distribution to predict the probability that an example $\mathbf{x}_i$ belongs to its observed label $\mathbf{y}_i$. A low probability is then used as a heuristic-likelihood of $\mathbf{y}_i$ being a label error. The confident joint can then be used to identify labeling errors and estimate the noise rate. 

\textbf{\underline{C}ontrastive and \underline{In}fluent \underline{C}ounter \underline{E}xample St\underline{r}ategy (CINCER or \texttt{CIN}) \cite{cincer}.} CINCER treats the problem of identifying labeling errors as a sequential decision making problem where a domain expert can be queried to relabel suspicious examples. CINCER uses the same heuristic as AUM to identify labeling errors, but also identifies counter-examples in the data to serve as explanations of the model's suspicion.

\textbf{Model-free Label Error Detection (SimiFeat) \cite{zhu2022detecting}.} Unlike other methods, SimiFeat does not need a (pre-)trained model to identify labeling errors. Instead, it utilizes labels of the $k$ nearest neighbors to identify labeling errors based on the \textit{clusterability} assumption, \textit{i.e.} data points with similar features should have the same true label with high probability. 

There are many methods to detect labeling errors, but we choose these methods as a \textit{starting point} because they are recent, state-of-the-art, and have different inputs and core assumptions. While all these methods have existing public implementations, through \texttt{AQuA}, our goal is to create a one-stop shop for using and evaluating open-source label error detection models. 

\subsection{Evaluation} 
There is significant variance in the ways that label cleaning methods are evaluated. To rigorously, fairly, and systematically assess these models, we unify the breadth of experimental settings through the following three dimensions of evaluation. 

\textbf{Supervision.} Identifying labeling errors in practice is an \textit{unsupervised} problem since we do not know which data points are mislabeled. Hence, evaluating these methods is a challenging endeavor. Most studies in the literature gather noise labels either from human experts (\textit{human-in-the-loop evaluation}) or by introducing synthetic label noise by design (\textit{synthetic label noise}). 

In human-in-the-loop evaluation, one or more human experts are asked to independently assess the true labels of data points identified as having erroneous labels \cite{pretrained_lm_2022, northcutt2021labelerrors}. While this is a straightforward and precise evaluation method, it is in general unscalable, expensive, time-consuming, and limited to only measuring the \textit{precision} of models (and not \textit{recall}), because the experts are typically only shown data points which a model considers erroneous. 

A much more common and scalable way of evaluating these methods is to introduce various kinds of synthetic label noise and measure a model's ability to detect them. There are many ways of introducing label noise, but injected noise may not always be reflective of the true noise that occurs in natural datasets, and hence identifying realistic noise injection strategies is an active area of research \cite{clusterability_alternative_anchor_points, zhu2022detecting, pretrained_lm_2022, part_dependent_label_noise, algan2020label, jiang2020beyond}. Moreover, model evaluation may still be noisy because there may be mislabeled examples for which our pseudo-noise labels are negative (or \textit{correctly labeled}).

\textbf{Hypotheses.} In general, existing studies evaluate two hypotheses: (1) \textit{cleaning labels on the train set improves the performance of the downstream classifier on the test set}, and (2) \textit{cleaning methods can accurately identify mislabeled data on the train set}. Hypothesis 1 is practical since the primary goal of identifying labeling errors is to train accurate and unbiased classifiers. However, appropriately regularized deep learning models are known to be naturally robust to some label noise. Hence, hypothesis 2 allows researchers to directly measure the efficacy of label cleaning techniques. 

\textbf{Measures of goodness.} Different studies use different measures of predictive accuracy. While some measure error rate \cite{aum_ranking}, others report the accuracy \cite{clusterability_alternative_anchor_points} or ROC-AUC \cite{active_label_cleaning} of their classification models. Similarly, for their cleaning methods, some studies report the $F_1$ score while others report the precision or recall \cite{northcutt2021confident, aum_ranking}. 

\textbf{More gaps in evaluation.} In addition to the lack of consistency, we believe that the experimental settings in many studies are occasionally (1)  \textit{unrealistic}, e.g., adding label noise to more than half (sometimes up to $80\%$) of the data points \cite{aum_ranking, northcutt2021confident}; and (2) \textit{uni-dimensional}, e.g., reporting only one metric of predictive performance.

\textbf{\texttt{AQuA}'s design.} To enable a realistic, multi-faceted and holistic evaluation of label error detection models, we implement \textbf{7} popular label noise injection techniques and multiple metrics of predictive performance. Specifically, for single-label datasets, we implement asymmetric~\cite{zhu2022detecting}, class-dependent~\cite{algan2020label}, instance-dependent~\cite{clusterability_alternative_anchor_points}, and uniform~\cite{algan2020label} noise, and for datasets with labels from multiple annotators, we implement dissenting label, dissenting worker, and crowd majority~\cite{pretrained_lm_2022}. In terms of metrics of predictive accuracy, we implement $F_1$, accuracy, (\textit{weighted}) precision, recall, area under ROC curve (ROC-AUC), average precision (PR-AUC), and error rate. We are in the process of implementing some other metrics beyond predictive accuracy, such as generalization \cite{generalization_measures} and robustness \citep{gisolfi2021model} of models. Our hope is that \texttt{AQuA}'s \textit{config-driven} design will allow non-technical users to integrate it into their labeling workflows and researchers to add new models, datasets, and evaluation pipelines seamlessly. Our choice of datasets and downstream classifiers ensures that the computational complexity of running experiments is not prohibitive. Finally, we make all code, pre-trained models, and experimental logs open-source to enable rigorous and fair evaluation of models.

\section{Experiments, Results and Discussion}
\label{sec:results}

\begin{table}[tbh!]
\centering
    \large
    \resizebox{0.8\columnwidth}{!}{
\begin{tabular}{c|cccc|cccc|cccc|cccc}
\Xhline{1pt}
\multirow{2}{*}{\textbf{Datasets}} & \multicolumn{4}{c}{\textbf{Uniform}} & \multicolumn{4}{c}{\textbf{Asymmetric}} & \multicolumn{4}{c}{\textbf{Class-dependent}} & \multicolumn{4}{c}{\textbf{Instance-dependent}} \\ \cmidrule{2-17}
 & \texttt{\textbf{AUM}} & \texttt{\textbf{CIN}} & \texttt{\textbf{CON}} & \texttt{\textbf{SIM}} & \texttt{\textbf{AUM}} & \texttt{\textbf{CIN}} & \texttt{\textbf{CON}} & \texttt{\textbf{SIM}} & \texttt{\textbf{AUM}} & \texttt{\textbf{CIN}} & \texttt{\textbf{CON}} & \texttt{\textbf{SIM}} & \texttt{\textbf{AUM}} & \texttt{\textbf{CIN}} & \texttt{\textbf{CON}} & \texttt{\textbf{SIM}} \\ \midrule
CIFAR-10 & 73.3 & 74.1 & 45.6 & \textbf{76.7} & 74.3 & 70.8 & 47.7 & \textbf{75.5} & 93.5 & 80.5 & 42.6 & \textbf{93.6} & 68.0 & 69.9 & 44.8 & \textbf{70.9} \\
Clothing-100K & 75.0 & 70.0 & \textbf{76.6} & 76.5 & 74.2 & 68.4 & 73.6 & \textbf{75.7} & 76.3 & 71.2 & 74.0 & \textbf{81.2} & 69.4 & 65.1 & \textbf{72.9} & 71.6 \\
NoisyCXR & \textbf{75.2} & 74.4 & 43.2 & 74.5 & \textbf{73.7} & 71.5 & 39.5 & 73.5 & 84.7 & 78.7 & 31.4 & \textbf{88.4} & 68.0 & 69.8 & 43.3 & \textbf{72.1} \\ \midrule
IMDb & 75.6 & 73.3 & 58.4 & \textbf{78.5} & 75.7 & 74.3 & 59.5 & \textbf{78.7} & 92.1 & 91.0 & 62.8 & \textbf{95.0} & 69.7 & 70.2 & 56.4 & \textbf{74.5} \\
TweetEval & 75.3 & 75.2 & 58.9 & \textbf{77.7} & 75.8 & 76.0 & 57.4 & \textbf{77.6} & 69.2 & 67.9 & 52.4 & \textbf{70.2} & 69.6 & 69.6 & 62.4 & \textbf{73.2} \\ \midrule
Credit Fraud & 75.8 & 75.8 & 73.3 & \textbf{78.1} & 75.7 & 75.8 & \textbf{80.0} & 76.7 & 63.3 & 63.0 & \textbf{87.2} & 72.0 & 69.5 & 69.4 & \textbf{74.3} & 73.5 \\
Adult & 75.7 & 75.8 & 72.9 & \textbf{78.5} & 75.8 & 75.8 & 66.9 & \textbf{77.5} & 63.6 & 64.6 & 61.2 & \textbf{64.9} & 69.6 & 70.2 & 68.7 & \textbf{72.4} \\
Dry Bean & 75.7 & \textbf{91.6} & 42.1 & 82.2 & 75.7 & \textbf{84.9} & 39.0 & 80.3 & 87.2 & \textbf{95.0} & 35.4 & 92.1 & 69.5 & \textbf{83.1} & 35.8 & 77.5 \\
Car Evaluation & 75.3 & 83.5 & 77.4 & \textbf{84.1} & 75.6 & 80.2 & 75.7 & \textbf{81.6} & 77.3 & \textbf{87.5} & 83.2 & 81.2 & 70.1 & \textbf{78.8} & 78.5 & 77.0 \\
Mushrooms & 76.0 & 82.5 & 62.7 & \textbf{85.2} & 75.7 & 80.7 & 66.3 & \textbf{83.0} & 99.3 & \textbf{100} & 75.5 & 99.8 & 69.5 & \textbf{75.4} & 64.1 & 74.3 \\
COMPAS & 75.8 & 74.9 & 63.2 & \textbf{75.9} & 75.8 & 74.8 & 64.6 & \textbf{76.5} & 55.5 & 57.1 & 52.9 & \textbf{57.7} & 69.5 & 69.4 & 61.0 & \textbf{73.1} \\ \midrule
Crop & 76.0 & \textbf{79.0} & 16.3 & 73.1 & \textbf{75.8} & 73.6 & 16.2 & 70.1 & 29.1 & 40.8 & 51.2 & \textbf{63.7} & \textbf{69.5} & 63.2 & 16.3 & 63.8 \\
Electric Devices & 75.8 & \textbf{82.2} & 35.0 & 79.3 & 75.7 & \textbf{78.6} & 35.3 & 75.8 & 37.8 & 50.5 & 55.9 & \textbf{68.3} & 69.9 & \textbf{71.5} & 32.7 & 69.2 \\
MIT-BIH & 75.6 & \textbf{88.4} & 49.7 & 83.3 & 75.7 & \textbf{83.0} & 51.3 & 78.4 & 68.2 & 75.7 & 45.4 & \textbf{80.6} & 69.6 & \textbf{78.4} & 48.1 & 75.2 \\
PenDigits & 75.8 & \textbf{89.0} & 23.1 & 73.4 & 75.7 & \textbf{83.1} & 23.4 & 72.7 & 46.7 & 44.9 & 53.5 & \textbf{78.4} & 69.9 & \textbf{76.0} & 19.8 & 68.1 \\
WhaleCalls & 75.6 & 74.9 & 60.3 & \textbf{77.3} & 75.7 & 75.5 & 61.8 & \textbf{77.2} & 42.3 & 44.7 & \textbf{52.4} & 47.1 & 69.6 & 69.1 & 59.2 & \textbf{71.2} \\
\Xhline{1pt}
\end{tabular}}
\caption{Performance evaluation of cleaning methods to detect erroneous labels across different types of synthetic noise added to the train set in terms of weighted $F_1$, averaged across noise rates and downstream models.
}
\label{tab:Results_2}
\end{table}

\begin{table}[htb!]
\centering
    \large
    \resizebox{\columnwidth}{!}{
\begin{tabular}{c|ccccc|ccccc|ccccc|ccccc|ccccc}
\Xhline{1pt}
\multirow{2}{*}{\textbf{Datasets}} & \multicolumn{5}{c}{\textbf{No   Noise Injected}} & \multicolumn{5}{c}{\textbf{Uniform}} & \multicolumn{5}{c}{\textbf{Asymmetric}} & \multicolumn{5}{c}{\textbf{Class-dependent}} & \multicolumn{5}{c}{\textbf{Instance-dependent}} \\ \cmidrule{2-26}
 & \texttt{\textbf{NON}} & \texttt{\textbf{AUM}} & \texttt{\textbf{CIN}} & \texttt{\textbf{CON}} & \texttt{\textbf{SIM}} & \texttt{\textbf{NON}} & \texttt{\textbf{AUM}} & \texttt{\textbf{CIN}} & \texttt{\textbf{CON}} & \texttt{\textbf{SIM}} & \texttt{\textbf{NON}} & \texttt{\textbf{AUM}} & \texttt{\textbf{CIN}} & \texttt{\textbf{CON}} & \texttt{\textbf{SIM}} & \texttt{\textbf{NON}} & \texttt{\textbf{AUM}} & \texttt{\textbf{CIN}} & \texttt{\textbf{CON}} & \texttt{\textbf{SIM}} & \texttt{\textbf{NON}} & \texttt{\textbf{AUM}} & \texttt{\textbf{CIN}} & \texttt{\textbf{CON}} & \texttt{\textbf{SIM}} \\ \midrule
CIFAR-10 & \textbf{74.3} & 74.1 & 73.0 & 46.0 & 73.5 & 63.2 & 62.6 & \cellcolor{blue!20}\textbf{65.0} & 36.9 & \cellcolor{blue!20}63.4 & 58.1 & \cellcolor{blue!20}\textbf{63.1} & \cellcolor{blue!20}62.2 & 38.3 & \cellcolor{blue!20}\textbf{63.1} & 71.0 & \cellcolor{blue!20}\textbf{71.7} & 70.5 & 46.2 & 67.5 & 57.7 & \cellcolor{blue!20}60.2 & \cellcolor{blue!20}\textbf{62.1} & 34.0 & 56.8 \\
Clothing-100K & \textbf{90.9} & 90.7 & 90.5 & 90.8 & 90.8 & 82.5 & 79.5 & \cellcolor{blue!20}83.2 & \cellcolor{blue!20}\textbf{85.4} & \cellcolor{blue!20}83.2 & 82.8 & 81.4 & 79.2 & 79.4 & \cellcolor{blue!20}\textbf{83.1} & 80.8 & \cellcolor{blue!20}83.3 & \cellcolor{blue!20}82.1 & \cellcolor{blue!20}83.8 & \cellcolor{blue!20}\textbf{86.1} & 78.9 & 74.6 & 71.7 & \cellcolor{blue!20}\textbf{81.6} & \cellcolor{blue!20}\textbf{81.6} \\
NoisyCXR & 56.0 & \cellcolor{blue!20}56.5 & \cellcolor{blue!20}56.7 & 25.2 & \cellcolor{blue!20}\textbf{57.0} & 49.6 & 49.4 & \cellcolor{blue!20}\textbf{52.2} & 19.1 & 48.2 & 49.6 & 48.8 & \cellcolor{blue!20}\textbf{50.6} & 18.0 & 47.9 & 54.2 & \cellcolor{blue!20}54.8 & \cellcolor{blue!20}\textbf{55.8} & 18.6 & 53.8 & 46.4 & \cellcolor{blue!20}46.7 & \cellcolor{blue!20}\textbf{48.3} & 18.9 & \cellcolor{blue!20}46.7 \\ \midrule
IMDb & 84.9 & \cellcolor{blue!20}87.5 & \cellcolor{blue!20}89.2 & 69.6 & \cellcolor{blue!20}\textbf{90.3} & 69.3 & 65.6 & 68.2 & 64.5 & \cellcolor{blue!20}\textbf{70.8} & 74.9 & 67.7 & \cellcolor{blue!20}76.8 & 66.4 & \cellcolor{blue!20}\textbf{80.6} & 87.1 & 85.5 & \cellcolor{blue!20}\textbf{89.1} & 84.4 & \cellcolor{blue!20}87.4 & 65.3 & 62.3 & \cellcolor{blue!20}\textbf{69.2} & 64.4 & 64.5 \\
TweetEval & 73.6 & 73.6 & \cellcolor{blue!20}\textbf{77.1} & 65.1 & \cellcolor{blue!20}76.8 & 71.3 & \cellcolor{blue!20}72.4 & \cellcolor{blue!20}\textbf{74.2} & 53.8 & \cellcolor{blue!20}73.2 & 71.3 & 68.4 & \cellcolor{blue!20}71.9 & 61.8 & \cellcolor{blue!20}\textbf{72.4} & \textbf{77.7} & 74.8 & 70.6 & 49.8 & 67.4 & 68.5 & \cellcolor{blue!20}69.4 & \cellcolor{blue!20}\textbf{71.7} & \cellcolor{blue!20}71.5 & 63.1 \\ \midrule
Credit Fraud & \textbf{100} & 99.9 & 99.9 & 99.9 & 99.9 & \textbf{99.9} & \textbf{99.9} & \textbf{99.9} & 88.8 & \textbf{99.9} & \textbf{99.9} & \textbf{99.9} & \textbf{99.9} & 99.8 & \textbf{99.9} & 99.6 & 66.6 & \cellcolor{blue!20}99.8 & \cellcolor{blue!20}\textbf{99.9} & 66.6 & \textbf{99.9} & 99.8 & \textbf{99.9} & 88.7 & 99.8 \\
Adult & \textbf{84.0} & \textbf{84.0} & \textbf{84.0} & 79.9 & \textbf{84.0} & 82.1 & 82.1 & \cellcolor{blue!20}\textbf{83.0} & 75.0 & 81.9 & \textbf{83.3} & 83.0 & 83.0 & 77.9 & 83.0 & 81.8 & \cellcolor{blue!20}82.3 & \cellcolor{blue!20}\textbf{83.1} & \cellcolor{blue!20}81.9 & 80.4 & \textbf{82.5} & 80.6 & 81.5 & 73.1 & 81.1 \\
Dry Bean & \textbf{91.9} & 90.9 & 91.2 & 57.2 & 90.6 & 89.5 & \cellcolor{blue!20}90.9 & \cellcolor{blue!20}\textbf{91.4} & 60.6 & 78.6 & 85.6 & \cellcolor{blue!20}87.0 & \cellcolor{blue!20}\textbf{89.3} & 55.1 & \cellcolor{blue!20}87.4 & \textbf{91.4} & 91.0 & 90.5 & 29.8 & 90.4 & 87.3 & 83.6 & 84.5 & 40.0 & \cellcolor{blue!20}\textbf{87.5} \\
Car Evaluation & \textbf{93.4} & 92.5 & 85.9 & 57.6 & 92.0 & \textbf{83.8} & 82.5 & 80.0 & 66.5 & 81.1 & \textbf{86.1} & 85.2 & 74.1 & 62.7 & 82.6 & \textbf{86.9} & 83.2 & 78.0 & 57.6 & 83.9 & \textbf{82.6} & 81.5 & 75.3 & 60.2 & 80.5 \\
Mushrooms & 99.7 & \cellcolor{blue!20}\textbf{99.9} & 99.6 & 99.7 & \cellcolor{blue!20}\textbf{99.9} & 98.3 & 98.2 & 97.8 & 89.0 & \cellcolor{blue!20}\textbf{98.7} & 97.9 & 97.5 & \cellcolor{blue!20}\textbf{98.4} & 87.3 & 96.5 & 99.5 & \cellcolor{blue!20}\textbf{100} & 99.3 & 99.1 & \cellcolor{blue!20}\textbf{99.9} & 95.3 & \cellcolor{blue!20}\textbf{96.9} & \cellcolor{blue!20}96.4 & 81.1 & \cellcolor{blue!20}95.8 \\
COMPAS & 67.2 & \cellcolor{blue!20}\textbf{67.3} & 66.2 & 63.7 & 66.6 & 65.6 & 62.1 & \cellcolor{blue!20}\textbf{65.9} & 58.6 & 65.3 & 66.1 & 64.4 & \cellcolor{blue!20}\textbf{66.2} & 46.9 & 65.6 & 54.3 & \cellcolor{blue!20}65.1 & \cellcolor{blue!20}63.8 & 35.5 & \cellcolor{blue!20}\textbf{66.2} & 61.6 & \cellcolor{blue!20}63.0 & \cellcolor{blue!20}61.8 & 48.6 & \cellcolor{blue!20}\textbf{63.7} \\ \midrule
Crop & \textbf{39.1} & 38.7 & 35.5 & 8.4 & 37.8 & 33.1 & \cellcolor{blue!20}37.2 & \cellcolor{blue!20}36.2 & 7.3 & \cellcolor{blue!20}\textbf{37.9} & \textbf{34.1} & 31.5 & 32.8 & 7.2 & 33.4 & \textbf{32.3} & 31.2 & 29.5 & 7.3 & 28.9 & 27.7 & \cellcolor{blue!20}27.8 & \cellcolor{blue!20}29.9 & 5.7 & \cellcolor{blue!20}\textbf{34.5} \\
Electric Devices & 45.3 & \cellcolor{blue!20}\textbf{48.0} & \cellcolor{blue!20}\textbf{48.0} & 29.8 & \cellcolor{blue!20}46.7 & 41.8 & \cellcolor{blue!20}42.6 & \cellcolor{blue!20}\textbf{44.8} & 27.3 & \cellcolor{blue!20}42.1 & 42.5 & 41.3 & 41.3 & 26.9 & \cellcolor{blue!20}\textbf{42.7} & 30.9 & 30.9 & \cellcolor{blue!20}\textbf{32.1} & 24.2 & \cellcolor{blue!20}31.7 & 39.3 & 36.6 & 38.3 & 23.1 & \cellcolor{blue!20}\textbf{40.4} \\
MIT-BIH & 72.7 & 65.1 & \cellcolor{blue!20}\textbf{81.2} & 55.7 & 72.5 & 73.2 & 70.1 & \cellcolor{blue!20}\textbf{80.1} & 61.7 & \cellcolor{blue!20}74.7 & \textbf{71.3} & 68.4 & 69.2 & 46.3 & 69.6 & 72.6 & \cellcolor{blue!20}73.9 & \cellcolor{blue!20}74.4 & 56.9 & \cellcolor{blue!20}\textbf{78.0} & 63.6 & \cellcolor{blue!20}68.1 & \cellcolor{blue!20}70.9 & 52.2 & \cellcolor{blue!20}\textbf{71.5} \\
PenDigits & 64.8 & \cellcolor{blue!20}\textbf{65.2} & 64.3 & 39.5 & 64.5 & 62.6 & \cellcolor{blue!20}\textbf{64.7} & \cellcolor{blue!20}64.4 & 24.6 & \cellcolor{blue!20}64.3 & 58.1 & \cellcolor{blue!20}\textbf{59.1} & 57.8 & 22.9 & \cellcolor{blue!20}59.0 & 43.9 & \cellcolor{blue!20}\textbf{46.5} & \cellcolor{blue!20}46.4 & 15.3 & \cellcolor{blue!20}45.3 & 59.2 & 56.4 & 57.7 & 14.8 & \cellcolor{blue!20}\textbf{59.7} \\
WhaleCalls & \textbf{68.2} & 34.3 & 50.9 & 52.7 & 53.0 & 48.7 & 44.5 & \cellcolor{blue!20}\textbf{51.0} & 43.7 & \cellcolor{blue!20}50.4 & 48.8 & \cellcolor{blue!20}\textbf{53.6} & 47.4 & 45.3 & 47.2 & 42.5 & \cellcolor{blue!20}43.3 & \cellcolor{blue!20}\textbf{47.1} & 41.6 & 42.4 & 48.5 & \cellcolor{blue!20}50.9 & \cellcolor{blue!20}\textbf{58.5} & 44.5 & 47.5 \\

\Xhline{1pt}
\end{tabular}}
\caption{
Impact of label noise and each cleaning method on weighted $F_1$ score of a downstream model for each modality on the test set, averaged across noise rates and downstream models. Highlighted cells indicate better performance than that obtained without label cleaning (\texttt{NON}).
}
\label{tab:Results_1}
\end{table}

\begin{figure}[!thb]
    \setlength{\tabcolsep}{1.5pt}
    \centering
    \begin{tabular}{ccc}
        \includegraphics[width=0.3\textwidth,trim={1.2cm 0.5cm 1.3cm 0.5cm},clip]{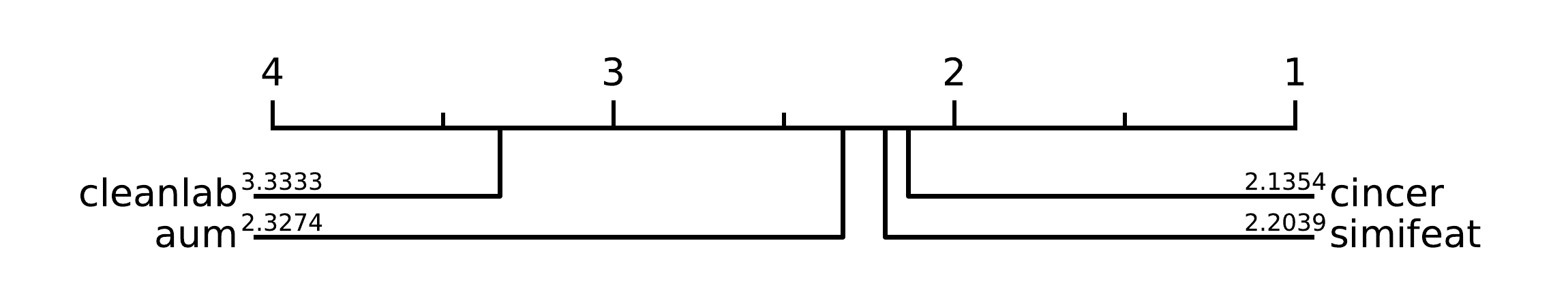} & 
        \includegraphics[width=0.3\textwidth,trim={1.2cm 0.4cm 1.3cm 0.5cm},clip]{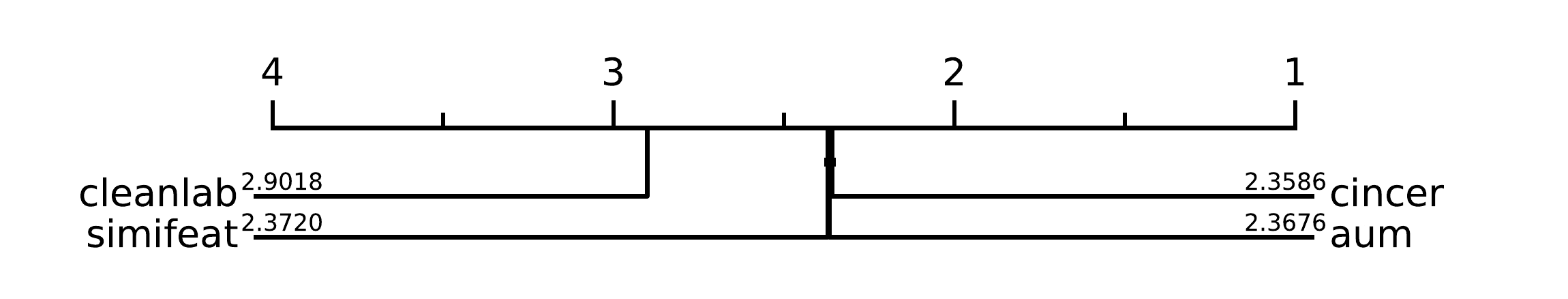} & 
        \includegraphics[width=0.3\textwidth,trim={1.2cm 0.5cm 1.3cm 0.5cm},clip]{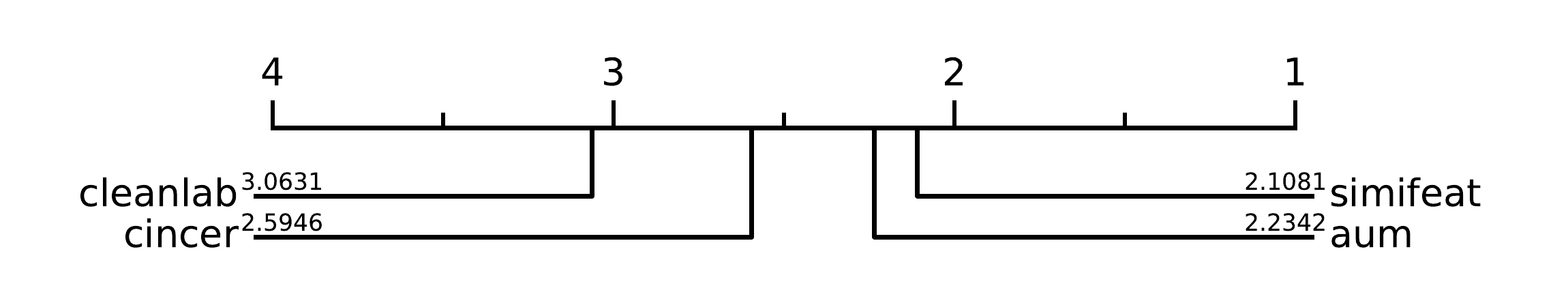} \\
        (\textit{i}) & (\textit{ii}) & (\textit{iii}) \\
        \includegraphics[width=0.3\textwidth,trim={1.2cm 0.5cm 1.3cm 0.5cm},clip]{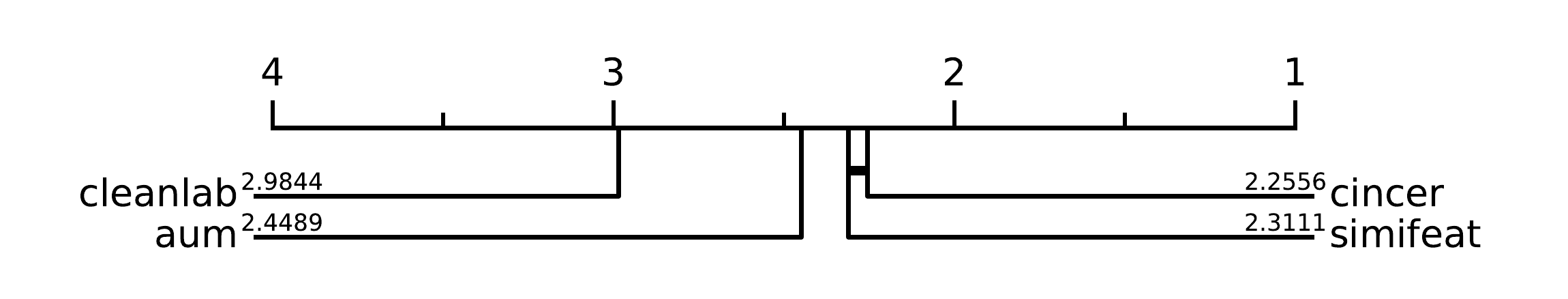} & 
        \includegraphics[width=0.3\textwidth,trim={1.2cm 0.5cm 1.3cm 0.5cm},clip]{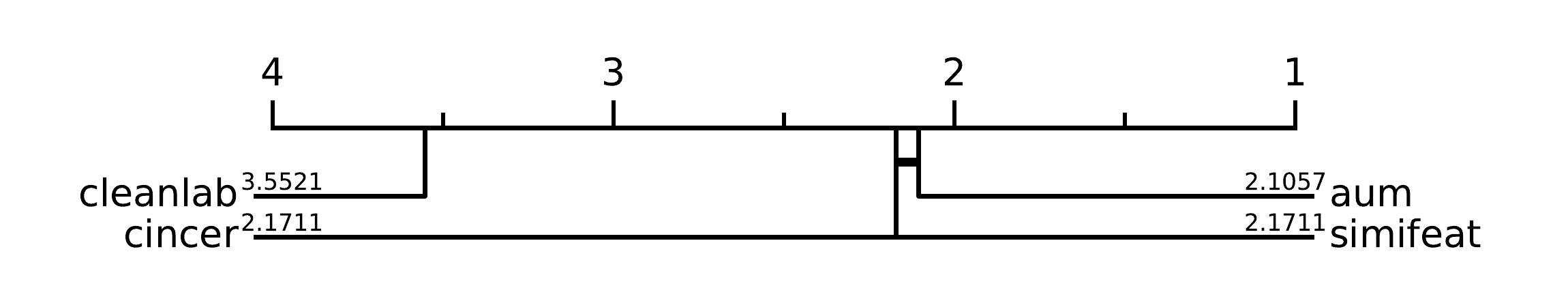} & 
        \includegraphics[width=0.3\textwidth,trim={0.2cm 0.5cm 0.2cm 0.5cm},clip]{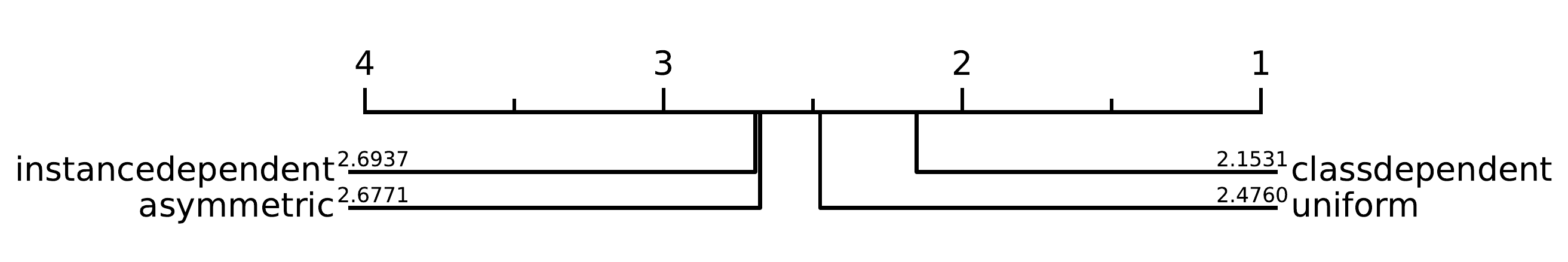} \\
        (\textit{iv}) & (\textit{v}) & (\textit{vi}) \\
    \end{tabular}
    \caption{\textit{Critical difference diagrams representing rankings of cleaning methods} across: (\textit{i}) all datasets,  (\textit{iii}) only image or (\textit{iv}) only text datasets. (\textit{v}) also shows the ranking of cleaning methods across all datasets when accuracy is measured instead of weighted $F_1$ (c.f. \textit{i}). Finally, (\textit{ii}) represents the performance of \textit{downstream models} trained using cleaned labels, and (\textit{vi}) performance of all cleaning methods disaggregated by noise type.}
    \label{fig:CD_overall}
\end{figure}

We conduct several experiments to support \texttt{AQuA}'s design choices and demonstrate its utility in providing a comprehensive and holistic evaluation of machine learning models in the presence of label noise. 

\textbf{Experimental Setup and Hyper-Parameter Tuning.} We run experiments for all combinations of cleaning methods (AUM (\texttt{AUM}), confident learning (\texttt{CON}), CINCER (\texttt{CIN}) and SimiFeat (\texttt{SIM}), including no label cleaning (\texttt{NON}), noise types (\textit{asymmetric}, \textit{class-dependent}, \textit{instance-dependent} and \textit{uniform}); for four different noise rates (0\%, 2\%, 10\% and 40\%), for a total of \textbf{2400 unique experiments}. We conduct experiments using three distinct classification architectures
for image and time-series data, and two different architectures for text and tabular data. To account for class imbalance in some datasets, we report the $F_1$ weighted by the support of each class. Results for all other evaluation metrics can be found in App.~\ref{app:additional_results}. We also adopt critical difference diagrams \citep{demvsar2006statistical} to succinctly represent comparisons between multiple cleaning methods and other independent variables (e.g., data modality and noise type) on multiple datasets. These diagrams represent the average ranks of methods across datasets while grouping those with insignificant difference\footnote{To form cliques, we abandon the posthoc test in favor of pairwise tests with Holm's correction for multiple testing based on prior work \citep{benavoli2016should, bagnall2017great}}. We tuned hyper-parameters of all the classification and cleaning methods till they performed reasonably well on average on all the datasets using hyper-parameter grids used by prior work and reported in App.~\ref{app:hyperparameters}\footnote{We deliberately did not perform extensive hyper-parameter tuning to not overfit to already existing label noise in the original datasets. Also, in practice it is unclear how to tune these cleaning methods well, without explicit knowledge of where the label errors are.}. Finally, all our experiments were carried out on a computing cluster, with a typical machine having 128 AMD EPYC 7502 CPUs, 503 GB of RAM, and 8 NVIDIA RTX A6000 GPUs. 

\textbf{Research Questions.} We aim to answer the following research questions through our experiments:
\begin{itemize}[noitemsep, leftmargin=*, label={\Large\textbullet}]
\vspace{-1mm}
    \item \textit{Which is the best cleaning method in terms of (i) its ability to identify synthetically injected label noise, and (ii) performance of the downstream classifier trained its cleaned labels?}
    \item \textit{Do the rankings of cleaning methods differ across different (i) types of synthetic label noise, (ii) data modalities, and (iii) evaluation metrics (weighted $F_1$ versus accuracy)?} 
\end{itemize}

\subsection{Insights from Large-scale Experiments using \texttt{AQuA}}

Tables~\ref{tab:Results_1}, \ref{tab:Results_2}, and Fig.~\ref{fig:CD_overall} report results from all our experiments aggregated by noise rate, and downstream classification models. Below we highlight some of our key findings. Due to lack of space, we defer finer grained results to App.~\ref{app:additional_results}. 

\textbf{Best cleaning method.} Overall, we found SimiFeat (\texttt{SIM}) \cite{zhu2022detecting} to be the best cleaning method in terms of its ability to identify synthetically injected label noise, closely followed by CINCER (\texttt{CIN}) \cite{cincer} (Fig.~\ref{fig:CD_overall}\textit{(i)}). However, these differences shrink when evaluating cleaning methods using the performance of the downstream model trained using their cleaned labels (Fig.~\ref{fig:CD_overall}\textit{(ii)}). Confident learning (\texttt{CON}) \cite{northcutt2021confident} consistently performed the worst among all the evaluated methods.

\textbf{Deep learning models are inherently robust to label noise.} Perhaps unsurprisingly, we found that most downstream classifiers were reasonably robust to synthetic label noise, as can be seen from the insignificant difference between the setting where datasets were not explicitly cleaned (\texttt{NON}), compared to when they were cleaned using \texttt{SIM}, \texttt{CIN} and \texttt{AUM}. These results also illustrate the importance of measuring both hypotheses (performance of cleaning methods versus downstream models) when evaluating the performance of ML models in the presence of label noise. 

\textbf{Adding label noise can sometimes improve model performance.} In the context of class-dependent or uniform noise, label noise serves as regularization to prevent models from overfitting. This phenomenon is not specific any one modality, but happens for multiple modalities, datasets, and noise types too, for example Electric Devices (time-series) under uniform noise, MIT-BIH (time-series), and Dry Bean (tabular) for class-dependent noise, in Table~\ref{app:results_table}. Moreover, deep learning optimization is highly non-convex, so adding some noise might help the model reach the global minima by traversing an alternative path within the loss landscape.

\textbf{Impact of \texttt{AQuA}'s design choices.} 
We found that cleaning methods perform differently for different data modalities. For instance, all cleaning methods barring \texttt{CON} perform on par on image datasets \textit{(iii)}, but on tabular data \textit{(iv)}, \texttt{AUM} performs significantly worse than \texttt{CIN} and \texttt{SIM}. This may be due to a variety of reasons beyond cleaning methods: size and nature of datasets, inductive biases of downstream classifier, and the quality of feature representations \citep{zhu2022detecting}. We also observed that some types of label noise are easier to detect than others. For example, uniform noise and asymmetric noise were the easiest to detect, cleaning methods found it much hard to detect instance and class-dependent noise \textit{(vi)}. Finally, we noticed differences in model rankings when measuring different evaluation metrics. As an example, the difference between \texttt{CIN} and \texttt{AUM} vanishes when we measure the accuracy \textit{(v)} of the cleaning methods instead of their weighted $F_1$ \textit{(i)}. These findings highlight the need to evaluate label error detection methods across multiple datasets from different modalities, noise types and evaluation metrics.

\section{Conclusion and Future Work}
\label{sec:conclusions}
We propose the first benchmark designed to rigorously evaluate machine learning models in the presence of label noise. We also elucidate the design space of these methods to not only enable ML practitioners to choose the right label cleaning tool for their data, but also foster academic research on the label noise problem. We demonstrate \texttt{AQuA}'s utility by running large-scale experiments to glean several interesting findings. We believe that, as a benchmarking toolkit, \texttt{AQuA} would benefit from more cleaning methods, datasets, synthetic label noise injection strategies, and evaluation metrics. 

Our short-term goals include experimenting with multi-annotator label noise, measuring the impact of feature noise on time-series and image data in comparison to label noise, incorporating several metrics for model generalization, robustness and fairness, and including audio datasets. While other types of noise are beyond the scope of this work, we believe that multi-annotator, multi-class multi-label, and noise in regression problems are exciting avenues of future work, and AQuA’s modular design will enable researchers to experiment with both multi-annotator and multi-class multi-label classification problems easily. We restrict ourselves to multi-class but single-label classification (as opposed to multi-label classification). 

We believe that future work on label error detection should address label issues in the multi-label classification and regression settings. We believe that our work on AQuA can both harness and facilitate the development of foundation models in the two ways: (1) foundation models can be used to identify labeling errors, without explicit supervision, and (2) methods within AQuA can be use to identify labeling errors which can affect foundation model pre-training and fine-tuning. We also believe that future work shoul

\section{Limitations, Biases, and Social Impacts}
\label{sec:limitations_biases_social}
We acknowledge the potential adverse impact of large-scale experimentation on the environment, but believe that our publicly accessible code and experimental findings can significantly reduce resource consumption for ML practitioners in this field. Label error detection models might perpetuate existing biases and impact the fairness of models. We included the Adult dataset, that is frequently used in the fairness literature, in AQuA, to evaluate the impact of label errors on the fairness of models. We would also like to acknowledge that our experiments were carried without extensive hyper-parameter tuning. Moreover, hyper-parameters for cleaning methods and downstream classifiers were chosen based on model performance on the observed training set and fixed throughout the training process. We futher discuss these design choices and their limitations in Appendix~\ref{app:reproducibility}.

\begin{ack}
We would like to thank Cherie Ho and Jack H. Good for their useful comments on initial drafts of the paper. 
This work was partially supported by the National Institutes of Health under awards R01HL141916, 1R01NS124642-01, and 1R01DK131586-01, and by the U.S. Army Research Office and the U.S. Army Futures Command under Contract No. W911NF-20-D-0002.
The content of the information does not necessarily reflect the position or the policy of the government and no official endorsement should be inferred.
\end{ack}

\bibliographystyle{unsrtnat}
\bibliography{bibliography}

\pagebreak




\appendix
\section{Appendix}

\subsection{A Design Space of Labeling Error Detection Models}
\label{app:assumptions}

In this section, we provide some more details on some of the key design decisions of various popular methods which enable machine learning in the presence of label noise. 

\paragraph{Noise Transition Matrix.} Many studies \citep{analyse_noise_model_for_realistic_data, part_dependent_label_noise, clusterability_alternative_anchor_points} explicitly estimate a probabilistic data structure called the noise transition matrix. A noise transition matrix $\mathbf{T}$ encodes the joint \cite{northcutt2021confident}, or more frequently the conditional probability \cite{analyse_noise_model_for_realistic_data, clusterability_alternative_anchor_points} of distribution of latent labels $y_i^*$ and observed noisy labels $y_i$, such that $\mathbf{T}_{ij} \triangleq \mathbb{P}(y = j \mid y^* = i; \mathbf{x})$. The noise transition matrix can be estimated in many different ways, e.g. (using \textit{anchor points}, labels of nearest neighbors (\textit{clusterability}), and pre-trained models). Similarly, the matrix can be either used to identify labeling errors explicitly \citep{northcutt2021confident}, or train robust machine learning models using modified loss functions. We note two key assumptions that a lot of these studies make, which might be violated in practice: (1) noise transition matrix is independent of the features of the data points, and (2) only a small fraction of the labels are noisy. To this end, recent studies have focused on designing novel techniques to estimate noise transition matrix while relax some of these assumptions (e.g., \citep{sample_sieve_cores, part_dependent_label_noise}). Below we briefly discuss three ways in which a noise transition matrix can be estimated, namely using \textit{anchor points}, \textit{nearest neighbours} and \textit{pre-trained models}, and one technique to use these matrices to train robust ML models. 

\paragraph{Estimating $\mathbf{T}$ using Anchor Points.} Intuitively, anchor points are samples in the training data which are highly likely to belong to a certain class. In particular, a data point $\mathbf{x}$ is an anchor for a class $i \in C$ if $\mathbb{P}(y^* = i \mid \mathbf{x}) = 1 - \epsilon$, where $\epsilon \rightarrow 0$. If $\epsilon = 0$, then $\mathbb{P}(Y = j \mid \mathbf{x}) = \sum_{k=1}^C \mathbf{T}_{kj}\mathbb{P}(Y = k \mid \mathbf{x}) = \mathbf{T}_{ij}$. Hence, $\mathbf{T}$ can be derived by evaluating the posterior probability that a anchor point belongs to noisy classes \citep{patrini2017making, classification_by_importance_reweighting}. While intuitive, using anchor points to estimate the transition matrix is not scalable, especially in scenarios where the number of classes is high and training data points is small since training a model which predicts the probability of noisy labels is challenging. Moreover, unavailability and identifiability of anchor points can limit the efficacy of these approaches, even if the posterior distribution can be learned accurately. Lastly, these methods lack the flexibility to extend to more complicated noise settings.

\paragraph{Estimating $\mathbf{T}$ using Clusterability.} These methods assume that data points with similar features should have the same class labels. Unlike previous methods based on anchor points, if good features are available off the shelf, then methods can be considered \textit{model-free}. Otherwise, reasonable features can automatically derived from intermediate-layer representations of deep learning models \cite{clusterability_alternative_anchor_points, deep_knn}. While these methods are intuitive, they rely on finding a good distance metric between the features. Moreover, these models might identify outliers as label noise, preventing the downstream classifier from learning meaningful data points. 

\paragraph{Estimating $\mathbf{T}$ using pre-trained models.} The key idea is to leverage a model trained on held-out data drawn from the same (or similar) distribution to predict the probability that an example $\mathbf{x}_i$ belongs to its observed label $\mathbf{y}_i$. A low probability is then used as a heuristic-likelihood of $\mathbf{y}_i$ being a label error. A careful count of these data points can then be use to estimate $\mathbf{T}$ \cite{northcutt2021confident}. 

But not all studies use pre-trained models to estimate $\mathbf{T}$. With the advent of pre-trained large language models, exploring their utility in detecting labeling errors \citep{pretrained_lm_2022} and studying their performance in the presence of label noise \citep{bert_robustness_to_label_noise} is an active area of research. Recently, \citep{pretrained_lm_2022} used the loss of a large language model to identify labeling errors, under the assumption that these models will exhibit large losses for erroneous data points. Another study demonstrated that unlike classical machine learning models, large language models may already be robust to label noise \citep{bert_robustness_to_label_noise}. 

\paragraph{Using $\mathbf{T}$ to train robust ML models.} We previously discussed how $\mathbf{T}$ can be used to identify labeling errors. There's another body of work which relies on the noise transition matrix to modify loss functions to make train machine learning models robust to label noise \cite{combating_noise_using_abstention, robust_loss_functions_2017, symmetric_cross_entropy_2019}. For example, given the noise transition matrix, \citet{patrini2017making} introduced forward and backward loss corrections, involving simple operations like matrix inversion and multiplication to make existing loss functions robust to noisy labels. 

Next, we provide a brief overview of techniques which do not explicitly estimate the noise transition matrix. We categorize these approches into three categories, primarily based on their key ideas: (1) approaches relying on the \textit{training dynamics} of ML models, (2) \textit{multi-network} approaches, and (3) approaches which leverage labels from \textit{multiple annotators}.

\paragraph{Approaches based on Training Dynamics.} These approaches exploit differences in training dynamics of clean and mislabeled samples to identify labeling errors. For example, Area under Margin Ranking~\citep{aum_ranking} identifies data points that do not contribute to the generalization of a model as labeling errors by leveraging the delicate tension between the label of a data point (via memorization) and its predicted label (via gradient updates), measured as the margin between the logits of a sample's assigned class and its highest unassigned class. On the other hand, \citet{clustering_training_losses} obtain the loss curves for each instance in a dataset from a neural network trained on a noisy training set, and apply clustering on these losses to separate clean and noisy samples. 

\paragraph{Multi-network approaches.} All methods we have discussed thus far use one model to identify labeling errors. But a few studies have leveraged two models to identify labeling errors, using either knowledge distillation~\cite{mentornet, coteaching}, or meta-learning~\cite{learn_from_noisy_labeled_data_meta_learning, meta_learning_correction}. These methods are expected to better identify different types of label errors as they rely on different models of different sizes and inductive biases.

The key idea of methods based on knowledge-distillation is to use a larger teacher network to supervise the training of a smaller student network. The teacher model identifies correctly labeled data points, and trains the student network on these samples only \citep{mentornet}. Instead of training the student and teacher models sequentially, some other studies propose to train the models simultaneously~\citep{coteaching, coteaching+}. 

A few studies utilize similar ideas to knowledge-distillation, instead using meta-learning to train robust machine learning models. For example, \citet{meta_learning_correction} propose a Meta Label Correction framework, where a label correction network acts as a meta-model to correct noisy labels, while the main model leverages these corrected labels. Some other methods re-weight training samples based on their gradient directions. These approaches generally comprise of a target and a meta-deep neural network, where the latter is trained on a clean validation set, and guides the training of the target network via sample re-weighting\citep{learn_from_noisy_labeled_data_meta_learning}. 

\paragraph{Multi-annotator labels.} 
These approaches are based on the premise that certain annotation tasks are inherently ambiguous, and even domain experts find it difficult to correctly label such instances. These methods aim to use multiple annotator labels to better model the noise transition matrix using the correlation between labels from different annotators to better estimate ground-truth consensus. These approaches are particularly useful for the healthcare domain due to the limited number of annotators but high variability of annotations\cite{variability_in_annotations_in_healthcare}. \citet{active_label_cleaning} introduce active label cleaning based on ``re-active learning", where they allow for re-annotation of already labeled instances in an active learning training scheme. Their proposed framework determines relabelling priority on the basis of the predicted posteriors from a classification model. Label cleaning is done over multiple iterations, and within each iteration, samples are initially ranked according to label prediction correctness and annotation difficulty. Each prioritized label is reviewed by multiple annotators until a consensus is formed using all generated labels. Drawing a leaf out of the crowd-sourcing literature, some other studies explicitly model the confusion matrix of each annotator to identify mislabeled data \citep{gao2022learning}.

\subsection{Relation with Weakly Supervised Learning}\label{app:weak_supervision}

AQuA serves two purposes: (1) as a benchmarking tool to evaluate methods that identify labeling errors, (2) and generally as a tool to identify labeling errors in a dataset and choose an appropriate cleaning method. Weakly supervised learning is a class of methods that learn from imperfect and weak sources of supervision to label datasets (see \citet{weak_super1} and \citet{weak_super2} as examples). The labels arising from these methods are indeed noisy. Methods in AQuA can therefore be used to clean datasets labeled using weakly supervised methods.
\subsection{Datasets and their characteristics}
\label{app:datasets}

\texttt{AQuA} currently comprises of a collection of \textbf{17} popular real-world public datasets from \textbf{4} prevalent data modalities: \textit{image}, \textit{text}, \textit{time-series} and \textit{tabular}. To evaluate label error detection models across various practical scenarios, we carefully choose datasets with diversity in the following characteristics: (1) \textit{classification problems} (\textit{e.g.}, sentiment classification vs.\ hate speech detection), (2) \textit{number of classes} (binary vs multi-class classification), (3) \textit{relative prevalence of classes} (e.g., skewed datasets like Credit Card Fraud~\cite{credit_card_dataset} and balanced ones like IMDb~\cite{IMDB}), (4) \textit{sources of annotations} (e.g., human vs rule-based annotation), and (5) \textit{number of annotations per example} (e.g., CIFAR-10N labeled by 3 annotators). Table~\ref{tab:aqua_datasets_app}  summarizes the key characteristics of datasets included as a part of \texttt{AQuA}. In particular, to make comparison with prior work easier while maintaining diversity across practical scenarios, we try to include datasets that have been used frequently by prior work (see usage in Table~\ref{tab:aqua_datasets_app}). Below we provide a brief description of datasets included in \texttt{AQuA}:

\begin{table}[!tb]
\centering
\resizebox{\columnwidth}{!}{
\begin{tabular}{l|lcccccc}
\Xhline{1pt}
\textbf{Modality} & \textbf{Dataset} & \textbf{\# Train / Test} & \textbf{\# Annotators/sample} & \textbf{Label Source} & \textbf{Classification Task} & \textbf{Sample Size} & \textbf{Usage}\\ \midrule
    
    \multirow{4}{*}{Image} & CIFAR-10N\cite{cifar_10n} & 50K / 10K & 3 & Human annotation & Object & $32 \times 32 \times 3$ & \cite{Model_agnostic_cleanlab, zhu2022detecting}\\
    
    & CIFAR-10H\cite{peterson2019human} & 0 / 10K & 47--63 & Human annotation & Object & $32 \times 32 \times 3$ & \cite{active_label_cleaning}\\
    
    & Clothing100K\cite{Clothing1M} & 100K & 1 & Web-labeled & Image & $256 \times 256 \times 3$ & \cite{aum_ranking, northcutt2021labelerrors, zhu2022detecting}\\

    & NoisyCXR\cite{noisycxr} & 26K / 3K & 1--XX & Human expert annotation & Pneumonia & $1024 \times 1024 \times 1$ & \cite{active_label_cleaning} \\ \midrule

    \multirow{2}{*}{Text} & IMDb$^\beta$\cite{IMDB} & 25K / 25K & 1 & Human annotation & Sentiment & - &  \cite{patrini2017making, robust_loss_functions_2017, northcutt2021labelerrors}\\

    & TweetEval\cite{tweeteval} & 10K & 1 & Human annotation & Hate speech & - & - \\ \midrule

    \multirow{6}{*}{Tabular} & Credit Card Fraud$^\beta$\cite{credit_card_dataset} & 284K & 1 & Human annotation & Credit card fraud & $28$ & \cite{clustering_training_losses, reconstruction_error_framework} \\

    & Adult$^\beta$\cite{uci} & 48K & 1 & Rule-based extraction & Salary & $14$ & \cite{cincer, wang2021fair, wu2022fair} \\

    & Dry Bean\cite{dry_bean_dataset} & 13K & 1 & Vision system-based annotation & Bean variety & $17$ & - \\

    & Car Evaluation\cite{hierarchical_decision_model} & 1K & 1 & Hierarchical decision model \cite{hierarchical_decision_model} & Car condition & $6$ & \cite{bootstrap_latent}\\

    & Mushroom$^\beta$\cite{mushroom_dataset} & 8K & 1 & - & Mushroom edibility & $22$ & \cite{clustering_training_losses}\\

    & COMPAS$^\beta$\cite{COMPAS_dataset} & 6K & 1 & - & Recidivism & $28$ & \cite{wang2021fair} \\ \midrule

    \multirow{5}{*}{\begin{tabular}{@{}c@{}}Time \\ Series \end{tabular}} & Crop\cite{dynamic_time_warping} & 7K / 16K & 1 & 
    \begin{tabular}{@{}c@{}}Hierarchical k-means tree \\ with dynamic time warping \cite{dynamic_time_warping} \end{tabular}
    & Crop cover & $46 \times 1$ & - \\

    & ElectricDevices\cite{electric_devices_dataset} & 9K / 7K & 1 & Human annotation & Appliance-type & $96 \times 1$ & - \\

    & MIT-BIH\cite{mit_bih} & 23K / 4K & 1 & Human expert annotation & Arrhythmia & $256 \times 2$ & - \\

    & PenDigits\cite{pendigits_dataset} & 7K / 3K & 1 & Human annotations & Handwritten digit & $16 \times 1$ & - \\

    & WhaleCalls$^\beta$\cite{UEA} & 11K / 2K & 1 & - & Whale call & $4,000 \times 1$ & - \\

\Xhline{1pt}
\end{tabular}}
\caption{\textbf{Summary of datasets.} \texttt{AQuA} currently includes a variety of datasets for different classification problems, varying in the number of classes, sources of annotations, and data modalities. All datasets except those marked with $\beta$ are multi-class.
}
\label{tab:aqua_datasets_app}
\end{table}

    \paragraph{\textbf{CIFAR-10N \cite{cifar_10n}:}} CIFAR-10N is a human-annotated dataset built upon the CIFAR-10 dataset, which is a $10$-class image dataset consisting of $32 \times 32$ color images, with each class containing a total of 6000 images. The classes are airplanes, cars, birds, cats, deer, dogs, frogs, horses, ships, and trucks, and they are all mutually exclusive. CIFAR-10N enables researchers to evaluate inter-annotator agreement-based metrics, since it contains 3 human-annotated labels per sample obtained from Amazon Mechanical Turk. The training set of the CIFAR-10N datasets consists of a ``clean label" along with three human-annotated labels on the training set of CIFAR-10.

    \paragraph{\textbf{CIFAR-10H \cite{peterson2019human}:}} Like CIFAR-10N, the CIFAR-10H data also comprises of multiple human annotations of the CIFAT-10 data. But unlike, CIFAR-10N, only the test set samples are annotated by crowd workers in Amazon Mechanical Turks. Each data point is annotated by 47 to 63 human annotators, making CIFAR-10H a repository of human perceptual uncertainty on the labels of CIFAR-10's testing data. 

    \paragraph{\textbf{Clothing100K \cite{Clothing1M, aum_ranking}:}} Clothing100K is a subset of the Clothing1M dataset, which includes over 1 million clothing images belonging to 14 different classes. The labels of data points are obtained by crawling online shopping websites, and therefore expected to reflect real-world noise. Due to the presence of real-world noise, most recently proposed studies evaluate their methods on Clothing1M or its subsets. 
    To speed up our experiments, we only use a subset of 100,000 samples to train and evaluate models in \texttt{AQuQ} \cite{aum_ranking}. 

    \paragraph{\textbf{NoisyCXR \cite{noisycxr}:}} NoisyCXR dataset is a multi-class dataset comprising of chest X-rays, with the primary goal of detecting pneumonia in lungs. Like CIFAR-10N and CIFAR-10H, this dataset too comprises of one or more expert-annotated labels. We included NoisyCXR since many data points have more than one expert labels and the dataset presents practical challenges prevalent in deploying machine learning in the real world such as ambiguously labels and vague samples.

    \paragraph{\textbf{IMDb \cite{IMDB}:}} The IMDb dataset consists of 50,000 highly polarized textual movie reviews from IMDb with labels for binary sentiment classification. Each sample is labeled either negative or positive. Using the 10-score rating system on IMDb, the review text is labeled negative when its star rating is $<= 4$, and it is considered positive when the star rating is $>= 7$. Any sample with scores greater than $4$ but less than $7$ is considered neither positive nor negative and excluded from the dataset. The training and testing splits contain 25,000 samples each, and each contains an equal number of positive and negative reviews.

    \paragraph{\textbf{TweetEval \cite{tweeteval}:}} TweetEval is a multi-task textual benchmark comprising of labels for seven different tasks including topic classification, sentiment analysis, irony detection, hate speech detection, offensive language detection, emoji prediction, and emotion analysis. For our benchmark, we chose the hate speech detection task primarily due to its size (i.e. the number of data points associated with hate speech labels was much larger than some other task), and real-world impact. These data points are obtained from Twitter and focus on the detection of hateful tweets targeting women and immigrants. The dataset contains an even number of training, validation, and testing samples.

    \paragraph{\textbf{Credit Card Fraud Detection Dataset \cite{credit_card_dataset}:}} This is a real-world binary classifcation tabular dataset obtained from European credit card holders' transactions in September 2013. We included this dataset due to its highly unbalanced class distribution: only a small fraction of 0.172\% of the samples are labeled as fraud. The attribute values for each sample are obtained after principle components analysis transformation to protect users' transaction information. Only the time and amount are not transformed and used as is. 

    \paragraph{\textbf{Adult \cite{uci}:}} The Adult dataset, also known as the ``Census Income" dataset, is a tabular binary class classification dataset used to predict whether or not an individual has an annual salary of $>=$ USD $50,000$. The data is collected and extracted from the 1994 Consensus database under the conditions: \texttt{((AAGE>16) \&\& (AGI>100) \&\& (AFNLWGT>1) \&\& (HRSWK>0))}. It contains attributes like age, work class, fnlwgt (the final weight, i.e., the number of people each row represents), education, education number, marital status, occupation, relationship, race, sex, capital gain, capital loss, hours per week, and native country. We included this dataset since it is widely used to evaluate advances in the context of the fairness of machine learning models. 

    \paragraph{\textbf{Dry Bean \cite{dry_bean_dataset}:}} This is a tabular multi-class classification dataset for classifying a sample into one of seven types of beans. It was created by clicking high-resolution images of 13,611 bean grains, and these images were subjected to segmentation and feature extraction, resulting in a total of 16 attributes: 12 based on dimensions and 4 based on shape form.

    \paragraph{\textbf{Car Evaluation \cite{hierarchical_decision_model}:}} This is a tabular multi-class classification dataset for evaluating a car's condition. It has class values ``unacceptable", ``acceptable", ``good" and ``very good". It was generated using a hierarchical decision model which evaluated cars based on three intermediate concepts: \texttt{TECH}, \texttt{PRICE}, and \texttt{COMFORT}. These intermediate concepts were further linked to 6 lower level concepts. Owing to this underlying structure, this dataset can be used for testing constructive induction and structure discovery methods.

    \paragraph{\textbf{Mushroom \cite{mushroom_dataset}:}} The Mushroom dataset is a tabular binary class classification dataset, created from descriptions of hypothetical records of 23 species of gilled mushrooms belonging to the Lepiota and Agaricus families. These 22 attribute, mushroom records were derived from The Audubon Society Field Guide to North American Mushrooms. Each species was originally labeled as \texttt{definitely poisonous}, \texttt{definitely edible}, or \texttt{unknown edibility}. However, the dataset creators merged the definitely poisonous and unknown edibility classes into one \texttt{poisonous} class. 

    \paragraph{\textbf{COMPAS \cite{COMPAS_dataset}:}} The Correctional Offender Management Profiling for Alternative Sanctions (COMPAS) dataset is obtained from pretrial COMPAS algorithm jurisdiction from Broward County Sheriff's Office in Florida to evaluate recidivism in cases in a two-year span. In COMPAS jurisdiction, each defendant receives three scores which include ``Risk of Recidivism," ``Risk of Violence" and ``Risk of Failure to Appear", which are based on the answers in the COMPAS survey \cite{COMPAS_dataset}. The data was compiled using the person's name, date of birth, and race, which sometimes could be incorrectly labeled and portray a wrong COMPAS score corresponding to the criminal records. Like Adult, the COMPAS dataset is also one of the most commonly used datasets to evaluate the fairness of machine learning models. 

    \paragraph{\textbf{Crop \cite{dynamic_time_warping}:}} The Crop dataset is a multi-class tabular dataset, obtained from the European Space Agency Sentinel-2 and NASA Landsat-8 program to demonstrate the change of landscape through its pixel over a period of time data. The change is observed through the change in the colors of the geographic coordinate shown in pixels over the time series. The dataset includes ``wheat crop", ``broad-leaved tree" and ``urban" classes. With the given pixels changing over the time series, they can be used to generate land-cover maps with different classes.

    \paragraph{\textbf{ElectricDevices \cite{electric_devices_dataset}:}} This is a multi-class time-series dataset for detecting the type of appliance from their electricity usage patterns. The dataset was created from the data recorded as part of a UK government study \emph{Powering the Nation}, conducted with the intention of collecting data about consumers' electricity use within the home to reduce the national carbon footprint. The dataset comprises of electricity readings from 251 households, taken over a month in 2-minute intervals.

    \paragraph{\textbf{MIT-BIH \cite{mit_bih}:}} The Massachusetts Institute of Technology - Beth Israel Hospital (MIT-BIH) dataset is a multi-class dataset comprising of electrocardiograms primarily used to evaluate automated arrhythmia detection algorithms \cite{goswami2021weak}. It is collected from a mixed population of 47 in-patients and out-patients. The analog output of the playback unit was filtered using a bandpass of 0.1--100 Hz and digitized with 360 Hz. Each record is 30 min long and was annotated by a simple QRS detector with revisited domain expert annotations. 

    \paragraph{\textbf{PenDigits \cite{pendigits_dataset}:}} This is a multi-class time-series handwritten digit classification dataset. It was created by tracing the pen used by 44 writers to draw digits across a digital screen. Then, the authors re-sampled the data spatially to generate attributes having a constant spatial step and variable time step. The data was further re-sampled to 8 spatial points, where each instance is 2 dimensions of 8 points.

    \paragraph{\textbf{WhaleCalls \cite{UEA}:}} The WhaleCalls dataset is a binary class time-series classification dataset for evaluating whether an audio signal is a right whale's up-call. Up-calls are right whale vocalizations in the acoustic range of 60--250Hz. They are often difficult to hear due to increased congestion in the low-frequency band with anthropogenic sounds like piling, naval operations, or ship noise. Thus, detecting right whale up-calls is a critical task, since it further enables maritime navigation technologies. 
\subsection{Classification Models Used in our Benchmark}\label{app:classification_models}
The ultimate goal of label cleaning is to train accurate downstream classifiers, but different studies use different classification models to measure the efficacy of their proposed label cleaning methods. To provide a level playing field for all cleaning methods, we include at least two classification model architectures for each data modality. Specifically, we include ResNet-18 \cite{resnet}, MobileNet \cite{mobilenet} and FastViT-T8 \cite{fastvit} for image datasets, \texttt{all-distilroberta-v1} \cite{roberta, distilbert} and \texttt{all-MiniLM-L6-v2} \cite{minilm} for text datasets, ResNet-1D, PatchTST \cite{PatchTST} and LSTM Fully Convolutional Network \cite{FCN} for time-series datasets, and TabTransformer \cite{tabtransformer} and a Multi-Layer Perceptron for tabular datasets. While choosing classification models we prioritized \textit{performant} methods with (1) \textit{different architectures} and \textit{inductive biases}, (2) ideally \textit{pre-trained} using different strategies, and (3) \textit{previously-used} either by label cleaning methods or task-relevant papers. We do not use tree-based models in our experiments, even though they are easy to integrate into AQuA, since they are incompatible with some of the label error detection methods like AUM. We provide a brief descriptions of all classification models included in \texttt{AQuA} below.
    
\paragraph{\textbf{ResNet-18 \cite{resnet}:}} ResNet is a commonly used computer vision architecture aimed at reducing the vanishing gradient problem in deep networks using jumping connections between layers and activating the previous layers. Our benchmark uses ResNet-18, which consists of 18 deep layers with a $7 \times 7$ kernel in the first layer, 4 identical ConvNet layers, and a fully connected layer with softmax activation. Each ConvNet layer has two blocks, each composed of two weight layers. Variants of ResNet are frequently used in the evaluation pipeline of popular label error detection models \citep{northcutt2021confident, aum_ranking}.

\paragraph{\textbf{MobileNet \cite{mobilenet}:}} MobileNet is a 53-layer deep convolutional neural network (CNN) used for mobile vision applications owing to its low computational intensity. It is implemented on the idea of depth-wise separable convolutions to create a light deep CNN having fewer parameters. Each depth-wise separable convolution is further composed of a depth-wise convolution and a point-wise convolution. Thus, MobileNet consists of a total of 28 layers, when accounting for the depth-wise and point-wise layers. After each convolutional layer, batch normalization and ReLU activation are applied. We include MobileNet because it has been shown to be performant and light-weight, enabling us to speed up our experiments. 

\paragraph{\textbf{FastViT-T8 \cite{fastvit}:}} FastViT-T8 is a hybrid vision transformer model that achieves state-of-the-art accuracy-latency tradeoff. It is trained using a novel token mixing operator, RepMixer, that uses structural reparameterization for lowering memory access costs by eliminating skip-connections in the network. To reduce latency, FastViT replaces dense $k x k$ convolutions with their factorised versions. The FastViT-T8 model has an expansion ratio less than 4 and a total of 8 FastViT blocks. It consists of a total of 3.6M parameters. We include it in our experiments since it adds a different architecture for evaluation and achieves a good balance between computational cost and accuracy.

\paragraph{\textbf{DistilRoBERTa \cite{distilbert,roberta}:}} We use the \texttt{all-distilroberta-v1} model, which is a pre-trained \texttt{distilroberta-base} model, further fine-tuned on a 1 billion sentence pairs dataset using a self-supervised contrastive learning objective, where the model is tasked with predicting one sentence out of a randomly sampled set of sentences which can be paired with an input sentence. It was trained to map sentences and paragraphs into 768-dimensional vector space and can be further used for clustering and semantic search. all-distilroberta-v1's ancestor BERT and RoBERTa have been frequently used by studies in natural language processing and detecting labeling errors \cite{pretrained_lm_2022} alike. 

\paragraph{\textbf{MiniLM-L6 \cite{minilm}:}} We also use the \texttt{all-MiniLM-L6-v2} model, which is a pre-trained \texttt{MiniLM-L6-H384-uncased} model, further fine-tuned on a 1 billion sentence pairs dataset using a self-supervised contrastive learning objective, where the model is tasked with predicting one sentence out of a randomly sampled set of sentences which can be paired with an input sentence. It was trained to map sentences and paragraphs into 384-dimensional vector space and can be further used for clustering and information retrieval applications. We included this model because it has a different inductive bias in comparison to all-distilroberta-v1 and is one of the fastest open-source pre-trained language models. 

\paragraph{\textbf{Multi-layer Perceptron:}} Multi-layer perceptron is a fully-connected multi-layer feed-forward connection of neurons, producing a set of output from a set of inputs. It typically consists of at least one hidden layer, which is any layer between the input and the output layer. Each layer consists of artificial neurons which apply activation function from the calculated sum from its inputs and forward it to the output. While it is frequently used for image classification, we apply it tabular data in our benchmark as a standard evaluation model to compare cleaning methods.

\paragraph{\textbf{TabTransformer \cite{tabtransformer}:}} TabTransformer is a deep data modeling architecture for tabular data built upon self-attention based transformer architecture for supervised and semi-supervised learning. It transforms categorical features into contextualized embeddings, outperforming other deep networks for tabular data while matching the performance of tree-based ensemble methods. The contextualized embeddings enable interpretability compared to context-free embeddings from competing approaches and are robust against noisy and missing data.

\paragraph{\textbf{ResNet-1D \cite{resnet}:}} While the ResNet architecture has classically been used in computer vision tasks, one-dimensional convolutional neural networks have been shown to be state-of-the-art from time series classification \citep{ismail2019deep}. In the healthcare domain, specifically in settings where there often are multiple channels of time series data, ResNet-1D can be implemented with channel attention to improve the model's learning efficiency from multi-feature channels. 

\paragraph{\textbf{PatchTST \cite{PatchTST}:}} PatchTST is a transformer model designed for multivariate time-series forecasting. It has two key design elements: patching and channel-independence. During patching, we segment the time-series into sub-series to be fed into the transformer as tokens. This aids in local semantic information retention in the embeddings, reduced computation and memory usage for attention maps, and enables the model to learn a longer sequence. Channel independence refers to individual channels containing a univariate time series with the same embedding and transformer weights, and enables PatchTST to surpass the long-term forecasting accuracy compared to state-of-the-art time-series transformer-based models.

\paragraph{\textbf{Fully Convolutional Network \cite{FCN}:}} A fully convolutional network (FCN) is a deep learning architecture primarily consisting of convolutional layers, pooling, and upsampling, and is commonly used for semantic segmentation. Since it typically lacks a dense layer, it is quick to train. In an FCN, a 1x1 convolutional layer replaces the conventional fully-connected convolutional layer and dense layers. In particular, we use an LSTM-FCN to evaluate cleaning methods on time-series classification tasks. Like ResNet-1D, FCNs too have been shown to perform well for time-series classification problems \citep{ismail2019deep}.
\subsection{Hyperparameters and Hyperparameter Grids}\label{app:hyperparameters}
We tuned hyper-parameters of all the classification and cleaning methods till they performed reasonably well on average on all the datasets using hyper-parameter grids in used by prior work and reported in Tables~\ref{tab:hp_cleaning_method} and \ref{tab:hp_base_model}. During training, we reduce the learning rate by a factor of $10$ if the loss does not improve for a ``patience" number of epochs.

We deliberately did not perform extensive hyper-parameter tuning so as to not overfit to already existing label noise in the original datasets. Also, in practice it is unclear how to tune these cleaning methods well, without explicit knowledge of where the label errors are. We also did not tune hyper-parameters for downstream classifiers so that differences in their performance could be directly attributed to the cleaning methods, rather than differences in their own hyper-parameters.

In the case of SimiFeat and CINCER, we selected hyperparameter grids based on the parameters outlined in the original papers that introduced these methods. However, for AUM, we had to define the hyperparameter grid ourselves, as the authors did not provide specific recommendations in their publication. Notably, Confident Learning did not involve any hyperparameters as part of its configuration.

\begin{table}[h!]
\centering
\resizebox{0.7\columnwidth}{!}{
\begin{tabular}{r|rl}
\Xhline{1pt}
\textbf{Label Error Detection Method} & \multicolumn{2}{c}{\textbf{Hyper-parameters}} \\ \midrule
\texttt{AUM} & \texttt{alpha} : & $\{\mathbf{0.01}, 0.05, 0.1, 0.15, 0.2\}$ \\ \midrule
\multirow{4}{*}{\texttt{CINCER}} & \texttt{threshold} : & $\{0.05, 0.1, 0.15, 0.2, \mathbf{0.25}\}$ \\
 & \texttt{inspector} : & $\{\textbf{margin}\}$ \\
 & \texttt{negotiator} : & $\{\textbf{random}\}$ \\
 & \texttt{nfisher radius} : & $\{\mathbf{0.1}\}$ \\ \midrule
\texttt{Confident Learning} & \multicolumn{2}{c}{$-$} \\ \midrule
\multirow{3}{*}{\texttt{SimiFeat}} & \texttt{max iter} : & $\{600, \mathbf{1000}\}$ \\
 & \texttt{min similarity} : & $\{\mathbf{0.45}, 0.5\}$ \\
 & \texttt{Tii offset} : & $\{0.1, 1.0, \mathbf{2.5}\}$ \\ \Xhline{1pt}
\end{tabular}}
\caption{Hyper-parameter grids for label error detection models. The final hyper-parameters chosen for our experiments are in bold. The exhaustive set of hyperparameters for all downstream classification models can be found in \texttt{https://github.com/autonlab/aqua/tree/main/aqua/configs/models/cleaning}.}
\label{tab:hp_cleaning_method}
\end{table}

\begin{table}[h!]
\centering
\resizebox{0.7\columnwidth}{!}{
\begin{tabular}{r|rl}
\Xhline{1pt}
\textbf{Model} & \multicolumn{2}{c}{\textbf{Hyper-parameters}} \\ \midrule
\multirow{5}{*}{ResNet-18} & \texttt{batch size} : & $\{64, 128, \mathbf{256}\}$ \\
 & \texttt{epochs} : & $\{\mathbf{20}\}$ \\
 & \texttt{learning rate} : & $\{0.005, \mathbf{0.01}, 0.1\}$ \\
 & \texttt{momentum} : & $\{\mathbf{0.8}, 0.9\}$ \\
 & \texttt{weight decay} : & $\{1e-5, \mathbf{1e-4}\}$ \\ \midrule
\multirow{5}{*}{MobileNet} & \texttt{batch size} : & $\{64, 128, \mathbf{256}\}$ \\
 & \texttt{epochs} : & $\{\mathbf{20}\}$ \\
 & \texttt{learning rate} : & $\{0.005, \mathbf{0.01}, 0.1\}$ \\
 & \texttt{momentum} : & $\{0.8, \mathbf{0.9}\}$ \\
 & \texttt{weight decay} : & $\{1e-5, \mathbf{1e-4}\}$ \\ \midrule
\multirow{5}{*}{FastViT-T8} & \texttt{batch size} : & $\{64, \mathbf{128}, 256\}$ \\
 & \texttt{epochs} : & $\{\mathbf{20}\}$ \\
 & \texttt{learning rate} : & $\{0.005, \mathbf{0.01}, 0.1\}$ \\
 & \texttt{momentum} : & $\{\mathbf{0.8}, 0.9\}$ \\
 & \texttt{weight decay} : & $\{1e-5, \mathbf{1e-4}\}$ \\ \midrule
\multirow{3}{*}{DistilRoBERTa} & \texttt{batch size} : & $\{\mathbf{64}, 128\}$ \\
 & \texttt{epochs} : & $\{1, \mathbf{2}, 3\}$ \\
 & \texttt{learning rate} : & $\{1e-5, 5e-5, \mathbf{1e-4}\}$ \\ \midrule
\multirow{3}{*}{MiniLM-L6} & \texttt{batch size} : & $\{\mathbf{64}, 128\}$ \\
 & \texttt{epochs} : & $\{1, 2, \mathbf{3}\}$ \\
 & \texttt{learning rate} : & $\{1e-5, 5e-5, \mathbf{1e-4}\}$ \\ \midrule
\multirow{4}{*}{Multi-layer Perceptron} & \texttt{batch size} : & $\{\mathbf{64}\}$ \\
 & \texttt{dropout rate} : & $\{\mathbf{0.0}, 0.1, 0.2\}$ \\
 & \texttt{epochs} : & $\{\mathbf{15}, 30\}$ \\
 & \texttt{learning rate} : & $\{\mathbf{0.001}, 0.005\}$ \\ \midrule
\multirow{5}{*}{TabTransformer} & \texttt{batch size} : & $\{\mathbf{64}\}$ \\
 & \texttt{momentum} : & $\{0.01, \mathbf{0.02}\}$ \\
 & \texttt{epochs} : & $\{5, \mathbf{10}, 20\}$ \\
 & \texttt{learning rate} : & $\{0.005, \mathbf{0.01}, 0.02\}$ \\
 & \texttt{mask type} : & $\{\mathbf{sparsemax}\}$ \\ \midrule
\multirow{3}{*}{ResNet-1D} & \texttt{batch size} : & $\{\mathbf{32}, 64, 128\}$ \\
 & \texttt{epochs} : & $\{\mathbf{5}, 10\}$ \\
 & \texttt{learning rate} : & $\{0.005, \mathbf{0.01}\}$ \\
 \midrule
\multirow{3}{*}{Fully Convolutional Network} & \texttt{batch size} : & $\{\mathbf{16}, 32, 64\}$ \\
 & \texttt{epochs} : & $\{5, \mathbf{10}\}$ \\
 & \texttt{learning rate} : & $\{0.005, \mathbf{0.01}\}$ \\
 \midrule
\multirow{4}{*}{PatchTST} & \texttt{batch size} : & $\{32, 64, \mathbf{128}\}$ \\
 & \texttt{epochs} : & $\{10, 20, \mathbf{40}, 80\}$ \\
 & \texttt{learning rate} : & $\{0.00005, \mathbf{0.0001}, 0.0002\}$ \\
 & \texttt{patch length} : & $\{8, \mathbf{16}, 32\}$ \\ \Xhline{1pt}
\end{tabular}}
\caption{Hyper-parameter grids for downstream classification models. The final hyper-parameters chosen for our experiments are in \textbf{bold}. The exhaustive set of hyperparameters for all downstream classification models can be found in \texttt{https://github.com/autonlab/aqua/tree/main/aqua/configs/models/base}.}
\label{tab:hp_base_model}
\end{table}

\subsection{Reproducibility and Replicability}\label{app:reproducibility}

\paragraph{Data cards.} A data card is a CSV file for a given dataset, random seed, noise rate, and noise type, where rows and columns correspond to data points and predictions of cleaning methods, respectively. Each data card also has two additional columns for corrupted (i.e. the static copy) and original labels of data points. All the cleaning methods are evaluated on the same labeling errors. All the data cards from out experiments are uploaded here\footnote{https://drive.google.com/drive/folders/1RHczHDUUilTOhcPyF5JSDvkO-rhiUKgb}.

\paragraph{Randomness.} We try to control all randomness in our experiments stemming from \texttt{PyTorch}, \texttt{random}, \texttt{numpy}, and CUDA. All our experiments are run with the random seed 42. For tabular data, we run two independent experiments with random seeds 42 and 43 for the multi-layer perception model.

\paragraph{Hyper-parameter tuning.} For each cleaning method and downstream classification model, for a given dataset, hyper-parameters were chosen based on model performance on the observed training set, measured using weighted $F_1$ score. Once chosen, hyper-parameters were frozen for all noise experiments (noise type + noise rate). However, this evaluation setup has the following limitations:
\begin{itemize}
    \item Tuning hyper-parameters based on the observed training set presents an advantage to the baseline method. In the ideal world, we should conduct extensive hyper-parameter tuning in each experiment setting, i.e. for each combination of dataset, noise rate, noise type, and cleaning method. However, that would be prohibitively expensive. Besides, we believe that insensitivity to hyper-parameters would be a hallmark of a good cleaning method.
    \item Tuning hyper-parameters based on a held-out validation set with no label errors prior to and after label cleaning. But this ideal scenario is contingent on a guaranteed error-free validation set and at least twice as much compute, which are prohibitive assumptions.
\end{itemize}

There were two primary reasons behind this design decision: (1) Our goal was to identify hyper-parameters that led to reasonable performance on the training set. Fine-grained tuning of hyper-parameters based on any dataset, whether held-out or in-domain, is tricky because the impact of label errors on model evaluation is hard to predict. We believe that evaluating model performance in the presence of label noise is a hard but important research direction that warrants a dedicated study. (2) Furthermore, it may not be important to pick the ``best" model that performs well on a held-out dataset, when in fact most if not all of the considered label cleaning methods utilize these downstream models (primarily trained on the training set) to learn representations of training data points. Once erroneous labels are identified, they are removed and the same model is re-trained on the ``cleaned" training data, and their performance is measured on the test data.

\subsection{Synthetic Label Noise}
\label{app:noise_types}
To enable a realistic, multi-faceted and holistic evaluation of label error detection models, we implement \textbf{7} popular label noise injection techniques and multiple metrics of predictive performance. Specifically, for single-label datasets, we implement asymmetric~\cite{zhu2022detecting}, class-dependent~\cite{algan2020label}, instance-dependent~\cite{clusterability_alternative_anchor_points}, and uniform~\cite{algan2020label} noise, and for datasets with labels from multiple annotators, we implement dissenting label, dissenting worker, and crowd majority~\cite{pretrained_lm_2022}.

    \paragraph{\textbf{Uniform Noise \cite{algan2020label}:}} For this type of noise, each entry in the noise transition matrix, except the diagonal ones, is equal. Specifically, for a noise rate $p \in [0, 1]$, 
    $$\mathbf{T}_{ij} = \begin{cases} 1 - p, & i = j \\ \frac{p}{M - 1}, & \text{otherwise} \end{cases}$$

    \paragraph{\textbf{Class-dependent Noise \cite{algan2020label}:}}  In this setting, similar classes have a higher probability of being mislabeled with each other. For any given dataset, we define the noise transition matrix as the confusion matrix derived from of a model that has been trained and evaluated on the dataset's training set. 

    \paragraph{\textbf{Asymmetric Label Noise \cite{zhu2022detecting}:}}
    We generate asymmetric noise by pair-wise flipping, i.e., for dataset with $K$ classes, we randomly flip the observed label $i$ to the next class $(i + 1) \ \text{mod} \ K$.

    \paragraph{\textbf{Instance-dependent Label Noise \cite{part_dependent_label_noise}:}} Unlike the previous settings, instance-dependent noise depends both on the data features and class labels to introduce realistic noise into a dataset. We follow Algorithm 2 in \citep{part_dependent_label_noise} to generate instance-dependent label noise. 

    We also implement three kinds of label noise for datasets which comprise of labels from multiple annotators following \citet{pretrained_lm_2022}. 

    \paragraph{\textbf{Dissenting Label :}} This approach randomly replaces the final labels with disagreeing labels to simulate a situation of imperfect quality control.

    \paragraph{\textbf{Dissenting Worker \cite{pretrained_lm_2022}:}} The dissenting worker approach simulates gaps in annotator training by randomly selecting an annotator and replacing the final labels with labels from the given annotator which do not match the final labels. This process is repeated for different annotators till the required noise rate is achieved. 

    \paragraph{\textbf{Crowd Majority \cite{pretrained_lm_2022}:}} The crowd majority approach can introduce systematic errors into a dataset by aggregating all individual annotations to produce a label other than the final label.

\pagebreak
\subsection{Additional Results}\label{app:additional_results}

\subsubsection{Performance of Cleaning Methods Across Different Synthetic Noise Types}

\begin{table}[h!]
\centering
\large
\resizebox{0.75\columnwidth}{!}{
\begin{tabular}{c|cccc|cccc|cccc|cccc}
\Xhline{1pt}
\multirow{2}{*}{\textbf{Datasets}} & \multicolumn{4}{c}{\textbf{Uniform}} & \multicolumn{4}{c}{\textbf{Asymmetric}} & \multicolumn{4}{c}{\textbf{Class-dependent}} & \multicolumn{4}{c}{\textbf{Instance-dependent}}  \\ \cmidrule{2-17}
 & \texttt{\textbf{AUM}} & \texttt{\textbf{CIN}} & \texttt{\textbf{CON}} & \texttt{\textbf{SIM}} & \texttt{\textbf{AUM}} & \texttt{\textbf{CIN}} & \texttt{\textbf{CON}} & \texttt{\textbf{SIM}} & \texttt{\textbf{AUM}} & \texttt{\textbf{CIN}} & \texttt{\textbf{CON}} & \texttt{\textbf{SIM}} & \texttt{\textbf{AUM}} & \texttt{\textbf{CIN}} & \texttt{\textbf{CON}} & \texttt{\textbf{SIM}} \\ \midrule
CIFAR-10 & 84.8 & 80.7 & 17.9 & 89.1 & 85.0 & 80.1 & 18.0 & 88.8 & 98.9 & 88.2 & 18.1 & 97.2 & 78.5 & 77.3 & 18.8 & 83.3 \\
Clothing-100K & 85.2 & 84.2 & 80.6 & 85.4 & 84.9 & 84.0 & 94.3 & 85.2 & 92.4 & 91.2 & 70.2 & 92.7 & 77.9 & 74.0 & 79.1 & 78.0  \\ 
NoisyCXR & 85.0 & 79.2 & 13.2 & 85.2 & 85.0 & 79.3 & 14.5 & 85.2 & 99.5 & 86.5 & 14.1 & 100 & 78.6 & 76.9 & 16.5 & 78.7 \\
\midrule
IMDb & 84.8 & 90.8 & 59.7 & 91.6 & 85.0 & 91.2 & 66.1 & 91.8 & 93.4 & 94.8 & 63.4 & 96.3 & 78.6 & 87.3 & 56.9 & 86.2 \\
TweetEval & 85.6 & 86.7 & 62.6 & 87.4 & 84.9 & 86.6 & 56.8 & 87.3 & 70.2 & 70.9 & 63.7 & 71.8 & 78.5 & 78.9 & 59.9 & 81.3  \\ \midrule

Credit Fraud & 85.0 & 85.3 & 81.5 & 96.1 & 85.0 & 85.2 & 97.7 & 92.3 & 76.6 & 76.6 & 88.5 & 92.9 & 78.6 & 78.7 & 93.4 & 94.5\\
\end{tabular}}
\caption{Performance evaluation of cleaning methods to detect erroneous labels across different types of synthetic noise added to the train set in terms of weighted $F_1$, for noise rate $= 0.1$. The classification models used for images, text, tabular, and time series datasets are ResNet-18, \texttt{all-distilroberta-v1}, Multi-layer perception (random seed 42), and ResNet-1D, respectively.
}
\end{table}

\begin{table}[h!]
\centering
\large
\resizebox{0.75\columnwidth}{!}{
\begin{tabular}{c|cccc|cccc|cccc|cccc}
\Xhline{1pt}
\multirow{2}{*}{\textbf{Datasets}} & \multicolumn{4}{c}{\textbf{Uniform}} & \multicolumn{4}{c}{\textbf{Asymmetric}} & \multicolumn{4}{c}{\textbf{Class-dependent}} & \multicolumn{4}{c}{\textbf{Instance-dependent}} \\ \cmidrule{2-17}
 & \texttt{\textbf{AUM}} & \texttt{\textbf{CIN}} & \texttt{\textbf{CON}} & \texttt{\textbf{SIM}} & \texttt{\textbf{AUM}} & \texttt{\textbf{CIN}} & \texttt{\textbf{CON}} & \texttt{\textbf{SIM}} & \texttt{\textbf{AUM}} & \texttt{\textbf{CIN}} & \texttt{\textbf{CON}} & \texttt{\textbf{SIM}} & \texttt{\textbf{AUM}} & \texttt{\textbf{CIN}} & \texttt{\textbf{CON}} & \texttt{\textbf{SIM}} \\ \midrule
CIFAR-10 & 84.8 & 81.3 & 18.2 & 88.9 & 84.9 & 83.2 & 17.8 & 88.4 & 90.5 & 84.0 & 17.6 & 91.6 & 78.4 & 80.8 & 18.7 & 80.0 \\
Clothing-100K & 85.2 & 83.1 & 94.4 & 85.4 & 84.9 & 83.0 & 94.3 & 85.2 & 84.8 & 78.1 & 94.2 & 85.0 & 77.9 & 68.7 & 74.4 & 78.0 \\ 
NoisyCXR & 85.0 & 78.3 & 13.8 & 85.2 & 84.9 & 77.0 & 15.1 & 85.2 & 90.9 & 80.8 & 16.3 & 91.6 & 78.6 & 75.0 & 15.2 & 79.0 \\
\midrule
IMDb & 84.8 & 80.3 & 59.5 & 90.4 & 84.9 & 81.9 & 59.6 & 90.8 & 88.0 & 83.9 & 60.8 & 92.5 & 79.0 & 81.8 & 57.3 & 85.7 \\
TweetEval & 85.6 & 86.2 & 56.7 & 86.6 & 85.1 & 85.9 & 62.6 & 86.3 & 73.0 & 73.6 & 46.2 & 75.0 & 78.6 & 78.7 & 60.0 & 81.8 \\ \midrule
Credit Fraud & 85.1 & 85.4 & 92.8 & 91.1 & 85.0 & 85.2 & 67.9 & 85.2 & 79.2 & 79.3 & 93.6 & 94.4 & 78.5 & 78.6 & 65.8 & 86.5 \\
\end{tabular}}
\caption{Performance evaluation of cleaning methods to detect erroneous labels across different types of synthetic noise added to the train set in terms of weighted $F_1$, for noise rate $= 0.1$. The classification models used for images, text, tabular, and time series datasets are MobileNet-v2, \texttt{all-MiniLM-L6-v2}, Multi-layer perception (random seed 43), and Fully convolutional network, respectively.
}
\end{table}

\begin{table}[h!]
\centering
\large
\resizebox{0.75\columnwidth}{!}{
\begin{tabular}{c|cccc|cccc|cccc|cccc}
\Xhline{1pt}
\multirow{2}{*}{\textbf{Datasets}} & \multicolumn{4}{c}{\textbf{Uniform}} & \multicolumn{4}{c}{\textbf{Asymmetric}} & \multicolumn{4}{c}{\textbf{Class-dependent}} & \multicolumn{4}{c}{\textbf{Instance-dependent}}  \\ \cmidrule{2-17}
 & \texttt{\textbf{AUM}} & \texttt{\textbf{CIN}} & \texttt{\textbf{CON}} & \texttt{\textbf{SIM}} & \texttt{\textbf{AUM}} & \texttt{\textbf{CIN}} & \texttt{\textbf{CON}} & \texttt{\textbf{SIM}} & \texttt{\textbf{AUM}} & \texttt{\textbf{CIN}} & \texttt{\textbf{CON}} & \texttt{\textbf{SIM}} & \texttt{\textbf{AUM}} & \texttt{\textbf{CIN}} & \texttt{\textbf{CON}} & \texttt{\textbf{SIM}} \\ \midrule
CIFAR-10 & 85.0 & 64.2 & 85.1 & 85.1 & 85.0 & 66.0 & 85.3 & 85.3 & 91.8 & 69.8 & 92.2 & 92.2 & 78.6 & 62.5 & 78.8 & 78.8 \\
Clothing-100K & 85.3 & 85.0 & 85.5 & 89.6 & 85.1 & 61.0 & 85.3 & 89.6 & 49.2 & 41.5 & 48.8 & 63.5 & 78.4 & 77.9 & 78.6 & 84.3 \\
NoisyCXR & 85.1 & 71.5 & 85.4 & 86.4 & 85.1 & 71.1 & 85.3 & 85.7 & 63.6 & 68.6 & 63.4 & 73.5 & 78.6 & 66.3 & 78.7 & 81.1 \\ \midrule
Credit Fraud & 85.0 & 85.2 & 97.7 & 85.2 & 85.0 & 85.3 & 95.4 & 85.4 & 34.8 & 33.9 & 79.9 & 33.9 & 78.7 & 78.8 & 65.9 & 78.8 \\
Adult & 85.0 & 86.0 & 85.3 & 87.0 & 85.0 & 85.9 & 85.3 & 87.2 & 64.6 & 65.6 & 64.4 & 67.2 & 78.7 & 79.9 & 78.7 & 81.3 \\
Dry Bean & 85.1 & 94.7 & 22.4 & 92.0 & 85.0 & 94.2 & 25.0 & 90.2 & 86.7 & 95.3 & 32.4 & 92.5 & 78.3 & 93.0 & 27.8 & 83.6 \\
\end{tabular}}
\caption{Performance evaluation of cleaning methods to detect erroneous labels across different types of synthetic noise added to the train set in terms of weighted $F_1$, for noise rate $= 0.1$. The classification models used for images, tabular, and time series datasets are  Fast-ViT-T8, TabTransformer, and PatchTST, respectively.}
\end{table}

\begin{table}[h!]
\centering
\large
\resizebox{0.75\columnwidth}{!}{
\begin{tabular}{c|cccc|cccc|cccc|cccc}
\Xhline{1pt}
\multirow{2}{*}{\textbf{Datasets}} & \multicolumn{4}{c}{\textbf{Uniform}} & \multicolumn{4}{c}{\textbf{Asymmetric}} & \multicolumn{4}{c}{\textbf{Class-dependent}} & \multicolumn{4}{c}{\textbf{Instance-dependent}}  \\ \cmidrule{2-17}
 & \texttt{\textbf{AUM}} & \texttt{\textbf{CIN}} & \texttt{\textbf{CON}} & \texttt{\textbf{SIM}} & \texttt{\textbf{AUM}} & \texttt{\textbf{CIN}} & \texttt{\textbf{CON}} & \texttt{\textbf{SIM}} & \texttt{\textbf{AUM}} & \texttt{\textbf{CIN}} & \texttt{\textbf{CON}} & \texttt{\textbf{SIM}} & \texttt{\textbf{AUM}} & \texttt{\textbf{CIN}} & \texttt{\textbf{CON}} & \texttt{\textbf{SIM}} \\ \midrule
CIFAR-10 & 45.9 & 67.2 & 32.1 & 47.3 & 45.8 & 59.2 & 32.6 & 48.5 & 98.9 & 88.2 & 18.1 & 97.2 & 40.4 & 59.4 & 35.0 & 48.0  \\
Clothing-100K & 45.2 & 36.0 & 59.1 & 44.7 & 45.5 & 36.2 & 24.9 & 45.0 & 92.4 & 91.2 & 70.2 & 92.7 & 40.2 & 31.0 & 58.6 & 39.6 \\ 
NoisyCXR & 45.7 & 70.8 & 31.6 & 45.1 & 45.7 & 58.1 & 30.8 & 45.0 & 99.5 & 86.5 & 14.1 & 100 & 39.8 & 62.8 & 33.8 & 39.0 \\
\midrule
IMDb & 45.6 & 44.9 & 50.6 & 47.6 & 45.6 & 45.0 & 50.7 & 47.6 & 93.4 & 94.8 & 63.4 & 96.3 & 40.5 & 39.6 & 50.7 & 43.7 \\
TweetEval & 44.2 & 43.4 & 47.3 & 49.7 & 45.8 & 45.0 & 46.9 & 53.1 & 70.2 & 70.9 & 63.7 & 71.8 & 39.5 & 39.0 & 53.4 & 51.8 \\ \midrule
Credit Fraud & 45.6 & 45.1 & 40.6 & 45.0 & 45.6 & 45.0 & 60.3 & 45.0 & 76.6 & 76.6 & 88.5 & 92.9 & 40.0 & 39.3 & 50.7 & 39.3 \\
\end{tabular}}
\caption{Performance evaluation of cleaning methods to detect erroneous labels across different types of synthetic noise added to the train set in terms of weighted $F_1$, for noise rate $= 0.4$. The classification models used for images, text, tabular, and time series datasets are ResNet-18, \texttt{all-distilroberta-v1}, Multi-layer perception (random seed 42), and ResNet-1D, respectively.
}
\end{table}

\begin{table}[h!]
\centering
\large
\resizebox{0.75\columnwidth}{!}{
\begin{tabular}{c|cccc|cccc|cccc|cccc}
\Xhline{1pt}
\multirow{2}{*}{\textbf{Datasets}} & \multicolumn{4}{c}{\textbf{Uniform}} & \multicolumn{4}{c}{\textbf{Asymmetric}} & \multicolumn{4}{c}{\textbf{Class-dependent}} & \multicolumn{4}{c}{\textbf{Instance-dependent}} \\ \cmidrule{2-17}
 & \texttt{\textbf{AUM}} & \texttt{\textbf{CIN}} & \texttt{\textbf{CON}} & \texttt{\textbf{SIM}} & \texttt{\textbf{AUM}} & \texttt{\textbf{CIN}} & \texttt{\textbf{CON}} & \texttt{\textbf{SIM}} & \texttt{\textbf{AUM}} & \texttt{\textbf{CIN}} & \texttt{\textbf{CON}} & \texttt{\textbf{SIM}} & \texttt{\textbf{AUM}} & \texttt{\textbf{CIN}} & \texttt{\textbf{CON}} & \texttt{\textbf{SIM}} \\ \midrule
CIFAR-10 & 46.0 & 72.3 & 32.2 & 63.7 & 45.7 & 60.8 & 32.4 & 48.7 & 90.5 & 84.0 & 17.6 & 91.6 & 40.0 & 64.9 & 34.9 & 47.8 \\
Clothing-100K & 45.2 & 33.5 & 78.4 & 44.7 & 45.5 & 39.1 & 69.7 & 45.0 & 84.8 & 78.1 & 94.2 & 85.0 & 40.2 & 35.6 & 82.7 & 39.6 \\ 
NoisyCXR & 45.7 & 68.5 & 31.4 & 45.1 & 45.6 & 57.0 & 28.8 & 45.2 & 90.9 & 80.8 & 16.3 & 91.6 & 40.4 & 59.9 & 33.5 & 41.2 \\
\midrule
IMDb & 45.6 & 44.9 & 50.4 & 47.8 & 45.7 & 45.0 & 50.8 & 48.6 & 88.0 & 83.9 & 60.8 & 92.5 & 40.2 & 39.5 & 49.9 & 42.7 \\
TweetEval & 44.2 & 43.4 & 40.7 & 53.1 & 45.6 & 45.0 & 54.3 & 48.2 & 73.0 & 73.6 & 46.2 & 75.0 & 40.8 & 40.0 & 53.8 & 44.1 \\ \midrule
Credit Fraud & 45.7 & 45.1 & 20.6 & 45.1 & 45.6 & 45.0 & 52.0 & 45.0 & 79.2 & 79.3 & 93.6 & 94.4 & 40.1 & 39.4 & 60.0 & 39.4 \\
\end{tabular}}
\caption{Performance evaluation of cleaning methods to detect erroneous labels across different types of synthetic noise added to the train set in terms of weighted $F_1$, for noise rate $= 0.4$. The classification models used for images, text, tabular, and time series datasets are MobileNet-v2, \texttt{all-MiniLM-L6-v2}, Multi-layer perception (random seed 43), and Fully convolutional network, respectively.
}
\end{table}

\begin{table}[h!]
\centering
\large
\resizebox{0.75\columnwidth}{!}{
\begin{tabular}{c|cccc|cccc|cccc|cccc}
\Xhline{1pt}
\multirow{2}{*}{\textbf{Datasets}} & \multicolumn{4}{c}{\textbf{Uniform}} & \multicolumn{4}{c}{\textbf{Asymmetric}} & \multicolumn{4}{c}{\textbf{Class-dependent}} & \multicolumn{4}{c}{\textbf{Instance-dependent}}  \\ \cmidrule{2-17}
 & \texttt{\textbf{AUM}} & \texttt{\textbf{CIN}} & \texttt{\textbf{CON}} & \texttt{\textbf{SIM}} & \texttt{\textbf{AUM}} & \texttt{\textbf{CIN}} & \texttt{\textbf{CON}} & \texttt{\textbf{SIM}} & \texttt{\textbf{AUM}} & \texttt{\textbf{CIN}} & \texttt{\textbf{CON}} & \texttt{\textbf{SIM}} & \texttt{\textbf{AUM}} & \texttt{\textbf{CIN}} & \texttt{\textbf{CON}} & \texttt{\textbf{SIM}} \\ \midrule
CIFAR-10 & 45.4 & 56.2 & 44.6 & 44.6 & 45.7 & 52.4 & 45.0 & 45.0 & 91.8 & 69.8 & 92.2 & 92.2 & 40.6 & 52.7 & 39.7 & 39.7 \\
Clothing-100K & 45.4 & 35.4 & 44.7 & 61.0 & 45.6 & 38.2 & 45.0 & 62.3 & 49.2 & 41.5 & 48.8 & 63.5 & 41.1 & 28.9 & 40.6 & 53.7 \\
NoisyCXR & 45.1 & 60.9 & 44.6 & 58.3 & 45.6 & 56.8 & 45.0 & 54.8 & 63.6 & 68.6 & 63.4 & 73.5 & 40.2 & 53.7 & 39.5 & 52.3 \\ \midrule
Credit Fraud & 45.9 & 45.2 & 40.6 & 45.2 & 45.6 & 45.0 & 60.4 & 45.0 & 34.8 & 33.9 & 79.9 & 33.9 & 40.0 & 39.2 & 40.2 & 39.2 \\
Adult & 45.7 & 45.0 & 45.0 & 64.7 & 45.8 & 45.6 & 45.0 & 45.2 & 64.6 & 65.6 & 64.4 & 67.2 & 40.8 & 40.0 & 40.0 & 53.7 \\
Dry Bean & 46.2 & 87.6 & 39.8 & 59.9 & 45.7 & 60.8 & 35.9 & 48.9 & 86.7 & 95.3 & 32.4 & 92.5 & 39.4 & 61.4 & 42.0 & 55.3 \\
\end{tabular}}
\caption{Performance evaluation of cleaning methods to detect erroneous labels across different types of synthetic noise added to the train set in terms of weighted $F_1$, for noise rate $= 0.4$. The classification models used for images, tabular, and time series datasets are  Fast-ViT-T8, TabTransformer, and PatchTST, respectively.}
\end{table}

\begin{table}[h!]
\centering
\large
\resizebox{0.75\columnwidth}{!}{
\begin{tabular}{c|cccc|cccc|cccc|cccc}
\Xhline{1pt}
\multirow{2}{*}{\textbf{Datasets}} & \multicolumn{4}{c}{\textbf{Uniform}} & \multicolumn{4}{c}{\textbf{Asymmetric}} & \multicolumn{4}{c}{\textbf{Class-dependent}} & \multicolumn{4}{c}{\textbf{Instance-dependent}}  \\ \cmidrule{2-17}
 & \texttt{\textbf{AUM}} & \texttt{\textbf{CIN}} & \texttt{\textbf{CON}} & \texttt{\textbf{SIM}} & \texttt{\textbf{AUM}} & \texttt{\textbf{CIN}} & \texttt{\textbf{CON}} & \texttt{\textbf{SIM}} & \texttt{\textbf{AUM}} & \texttt{\textbf{CIN}} & \texttt{\textbf{CON}} & \texttt{\textbf{SIM}} & \texttt{\textbf{AUM}} & \texttt{\textbf{CIN}} & \texttt{\textbf{CON}} & \texttt{\textbf{SIM}} \\ \midrule
CIFAR-10 & 96.5 & 86.9 & 17.9 & 96.5 & 96.5 & 86.8 & 17.9 & 96.6 & 96.6 & 87.1 & 17.8 & 96.2 & 90.0 & 82.9 & 17.5 & 92.7 \\
Clothing-100K & 96.8 & 96.7 & 85.0 & 97.3 & 96.5 & 96.5 & 84.9 & 97.0 & 99.5 & 99.7 & 96.8 & 100 & 90.0 & 89.5 & 94.7 & 90.4 \\
NoisyCXR & 96.4 & 84.8 & 12.8 & 96.9 & 96.5 & 84.3 & 15.9 & 97.0 & 99.4 & 86.6 & 15.6 & 100 & 89.7 & 81.0 & 15.2 & 90.1 \\
\midrule
IMDb & 96.5 & 97.5 & 65.0 & 96.9 & 96.5 & 96.6 & 64.9 & 96.9 & 95.8 & 95.5 & 64.5 & 96.6 & 89.8 & 92.8 & 61.4 & 94.8 \\
TweetEval & 96.2 & 94.6 & 77.9 & 94.9 & 96.6 & 96.6 & 61.9 & 95.2 & 72.0 & 72.2 & 52.1 & 71.9 & 90.2 & 90.6 & 64.9 & 90.3
\\ \midrule
Credit Fraud & 96.5 & 97.0 & 99.0 & 99.2 & 96.6 & 97.0 & 99.0 & 97.3 & 76.2 & 76.3 & 88.3 & 85.1 & 89.9 & 90.2 & 96.9 & 94.5 \\
\end{tabular}}
\caption{Performance evaluation of cleaning methods to detect erroneous labels across different types of synthetic noise added to the train set in terms of weighted $F_1$, for noise rate $= 0.02$. The classification models used for images, text, tabular, and time series datasets are ResNet-18, \texttt{all-distilroberta-v1}, Multi-layer perception (random seed 42), and ResNet-1D, respectively.
}
\end{table}

\begin{table}[h!]
\centering
\large
\resizebox{0.75\columnwidth}{!}{
\begin{tabular}{c|cccc|cccc|cccc|cccc}
\Xhline{1pt}
\multirow{2}{*}{\textbf{Datasets}} & \multicolumn{4}{c}{\textbf{Uniform}} & \multicolumn{4}{c}{\textbf{Asymmetric}} & \multicolumn{4}{c}{\textbf{Class-dependent}} & \multicolumn{4}{c}{\textbf{Instance-dependent}} \\ \cmidrule{2-17}
 & \texttt{\textbf{AUM}} & \texttt{\textbf{CIN}} & \texttt{\textbf{CON}} & \texttt{\textbf{SIM}} & \texttt{\textbf{AUM}} & \texttt{\textbf{CIN}} & \texttt{\textbf{CON}} & \texttt{\textbf{SIM}} & \texttt{\textbf{AUM}} & \texttt{\textbf{CIN}} & \texttt{\textbf{CON}} & \texttt{\textbf{SIM}} & \texttt{\textbf{AUM}} & \texttt{\textbf{CIN}} & \texttt{\textbf{CON}} & \texttt{\textbf{SIM}} \\ \midrule
CIFAR-10 & 74.8 & 85.4 & 65.1 & 77.9 & 83.8 & 76.2 & 83.4 & 81.5 & 90.6 & 84.0 & 17.6 & 91.6 & 74.7 & 79.8 & 69.3 & 77.4 \\
Clothing-100K & 89.9 & 79.0 & 64.6 & 83.1 & 82.8 & 80.1 & 67.1 & 74.7 & 84.9 & 78.1 & 94.2 & 85.0 & 88.2 & 98.3 & 56.3 & 87.0 \\
NoisyCXR & 91.6 & 81.4 & 58.5 & 73.0 & 78.3 & 85.3 & 22.8 & 68.9 & 90.9 & 80.9 & 16.3 & 91.7 & 75.8 & 80.9 & 66.8 & 97.8 \\
\midrule
IMDb & 96.5 & 81.4 & 65.0 & 96.4 & 96.5 & 86.2 & 64.9 & 96.3 & 94.0 & 93.1 & 63.5 & 95.4 & 90.1 & 80.1 & 61.8 & 94.1 \\
TweetEval & 96.2 & 96.7 & 67.7 & 94.6 & 96.6 & 97.0 & 61.8 & 95.1 & 56.8 & 46.3 & 42.3 & 55.6 & 89.7 & 90.1 & 82.3 & 89.8 \\
 \midrule
Credit Fraud & 96.6 & 97.0 & 87.1 & 97.7 & 96.6 & 97.0 & 87.1 & 97.5 & 77.0 & 77.1 & 92.9 & 86.6 & 89.8 & 90.2 & 96.9 & 98.4 \\
\end{tabular}}
\caption{Performance evaluation of cleaning methods to detect erroneous labels across different types of synthetic noise added to the train set in terms of weighted $F_1$, for noise rate $= 0.02$. The classification models used for images, text, tabular, and time series datasets are MobileNet-v2, \texttt{all-MiniLM-L6-v2}, Multi-layer perception (random seed 43), and Fully convolutional network, respectively.
}
\end{table}

\begin{table}[h!]
\centering
\large
\resizebox{0.75\columnwidth}{!}{
\begin{tabular}{c|cccc|cccc|cccc|cccc}
\Xhline{1pt}
\multirow{2}{*}{\textbf{Datasets}} & \multicolumn{4}{c}{\textbf{Uniform}} & \multicolumn{4}{c}{\textbf{Asymmetric}} & \multicolumn{4}{c}{\textbf{Class-dependent}} & \multicolumn{4}{c}{\textbf{Instance-dependent}}  \\ \cmidrule{2-17}
 & \texttt{\textbf{AUM}} & \texttt{\textbf{CIN}} & \texttt{\textbf{CON}} & \texttt{\textbf{SIM}} & \texttt{\textbf{AUM}} & \texttt{\textbf{CIN}} & \texttt{\textbf{CON}} & \texttt{\textbf{SIM}} & \texttt{\textbf{AUM}} & \texttt{\textbf{CIN}} & \texttt{\textbf{CON}} & \texttt{\textbf{SIM}} & \texttt{\textbf{AUM}} & \texttt{\textbf{CIN}} & \texttt{\textbf{CON}} & \texttt{\textbf{SIM}} \\ \midrule
CIFAR-10 & 96.6 & 72.5 & 97.0 & 97.0 & 96.4 & 72.3 & 97.0 & 97.0 & 91.8 & 69.8 & 92.2 & 92.2 & 90.3 & 68.4 & 90.6 & 90.6 \\
Clothing-100K & 96.7 & 97.0 & 97.1 & 97.1 & 96.6 & 97.0 & 97.0 & 97.2 & 49.2 & 41.5 & 48.8 & 63.5 & 90.4 & 82.0 & 90.7 & 93.2 \\
NoisyCXR & 96.6 & 73.5 & 97.1 & 94.8 & 96.6 & 74.1 & 97.0 & 94.4 & 63.6 & 68.6 & 63.4 & 73.5 & 89.8 & 71.4 & 90.2 & 89.8 \\ \midrule
Credit Fraud & 96.6 & 97.0 & 99.8 & 97.6 & 96.6 & 97.0 & 99.8 & 97.5 & 34.8 & 33.9 & 79.9 & 33.9 & 89.8 & 90.1 & 98.4 & 90.3 \\
Adult & 96.6 & 95.8 & 97.0 & 95.6 & 97.0 & 95.9 & 97.0 & 95.3 & 64.6 & 65.6 & 64.4 & 67.2 & 89.9 & 90.2 & 90.2 & 90.9 \\
Dry Bean & 96.4 & 96.1 & 28.5 & 97.1 & 96.6 & 95.9 & 32.4 & 97.0 & 86.7 & 95.3 & 32.4 & 92.5 & 90.4 & 95.0 & 28.6 & 93.3 \\
\end{tabular}}
\caption{Performance evaluation of cleaning methods to detect erroneous labels across different types of synthetic noise added to the train set in terms of weighted $F_1$, for noise rate $= 0.02$. The classification models used for images, tabular, and time series datasets are  Fast-ViT-T8, TabTransformer, and PatchTST, respectively.}
\end{table}

\vfill

\pagebreak

\subsubsection{Impact of Label Noise on Weighted $F_1$ Score}

\begin{table}[h!]
\centering
\large
\resizebox{\columnwidth}{!}{
\begin{tabular}{c|ccccc|ccccc|ccccc|ccccc|ccccc}
\Xhline{1pt}
\multirow{2}{*}{\textbf{Datasets}} & \multicolumn{5}{c}{\textbf{No   Noise Injected}} & \multicolumn{5}{c}{\textbf{Uniform}} & \multicolumn{5}{c}{\textbf{Asymmetric}} & \multicolumn{5}{c}{\textbf{Class-dependent}} & \multicolumn{5}{c}{\textbf{Instance-dependent}} \\ \cmidrule{2-26}
 & \texttt{\textbf{NON}} & \texttt{\textbf{AUM}} & \texttt{\textbf{CIN}} & \texttt{\textbf{CON}} & \texttt{\textbf{SIM}} & \texttt{\textbf{NON}} & \texttt{\textbf{AUM}} & \texttt{\textbf{CIN}} & \texttt{\textbf{CON}} & \texttt{\textbf{SIM}} & \texttt{\textbf{NON}} & \texttt{\textbf{AUM}} & \texttt{\textbf{CIN}} & \texttt{\textbf{CON}} & \texttt{\textbf{SIM}} & \texttt{\textbf{NON}} & \texttt{\textbf{AUM}} & \texttt{\textbf{CIN}} & \texttt{\textbf{CON}} & \texttt{\textbf{SIM}} & \texttt{\textbf{NON}} & \texttt{\textbf{AUM}} & \texttt{\textbf{CIN}} & \texttt{\textbf{CON}} & \texttt{\textbf{SIM}} \\ \midrule
CIFAR-10 & 81.1 & 81.1 & 80.4 & 24.0 & 79.4 & 74.4 & 73.3 & 78.0 & 18.6 & 75.7 & 74.7 & 74.2 & 77.0 & 12.5 & 75.4 & 80.5 & 80.6 & 79.9 & 27.0 & 80.3 & 72.6 & 71.2 & 76.4 & 12.5 & 73.4  \\
Clothing-100K & 90.9 & 91.0 & 91.1 & 90.9 & 90.9 & 87.8 & 86.5 & 81.5 & 89.9 & 88.5 & 90.0 & 89.9 & 89.8 & 90.9 & 89.9 & 90.0 & 89.6 & 88.7 & 90.3 & 90.3 & 87.6 & 77.4 & 86.6 & 86.2 & 84.7  \\
NoisyCXR & 65.4 & 65.3 & 64.7 & 10.4 & 64.5 & 61.6 & 61.0 & 63.3 & 7.3 & 61.8 & 61.3 & 62.1 & 63.8 & 9.5 & 61.5 & 65.0 & 65.7 & 65.0 & 7.3 & 65.8 & 59.4 & 59.3 & 62.0 & 13.0 & 59.1 \\ \midrule
IMDb & 89.1 & 90.5 & 93.1 & 80.0 & 92.1 & 92.5 & 92.0 & 92.3 & 80.5 & 89.4 & 91.3 & 90.4 & 87.8 & 88.9 & 92.2 & 92.4 & 92.3 & 91.1 & 91.3 & 89.6 & 90.0 & 91.6 & 89.1 & 86.9 & 79.2 \\
TweetEval & 82.1 & 80.7 & 81.9 & 60.4 & 81.8 & 82.5 & 76.7 & 80.0 & 78.9 & 82.0 & 81.5 & 66.9 & 81.8 & 79.6 & 81.4 & 81.6 & 82.3 & 77.9 & 66.7 & 76.8 & 79.6 & 78.0 & 78.5 & 80.0 & 78.7 \\
\end{tabular}}
\caption{
Impact of label noise and each cleaning method on weighted $F_1$ score of a downstream model for each modality on the test set for noise rate $= 0.1$. The classification models used for images, text, tabular, and time series datasets are ResNet-18, \texttt{all-distilroberta-v1}, Multi-layer perception (random seed 42), and ResNet-1D, respectively.}
\label{app:results_table}
\end{table}

\begin{table}[h!]
\centering
\large
\resizebox{\columnwidth}{!}{
\begin{tabular}{c|ccccc|ccccc|ccccc|ccccc|ccccc}
\Xhline{1pt}
\multirow{2}{*}{\textbf{Datasets}} & \multicolumn{5}{c}{\textbf{No Noise Injected}} & \multicolumn{5}{c}{\textbf{Uniform}} & \multicolumn{5}{c}{\textbf{Asymmetric}} & \multicolumn{5}{c}{\textbf{Class-dependent}} & \multicolumn{5}{c}{\textbf{Instance-dependent}} \\ \cmidrule{2-26}
 & \texttt{\textbf{NON}} & \texttt{\textbf{AUM}} & \texttt{\textbf{CIN}} & \texttt{\textbf{CON}} & \texttt{\textbf{SIM}} & \texttt{\textbf{NON}} & \texttt{\textbf{AUM}} & \texttt{\textbf{CIN}} & \texttt{\textbf{CON}} & \texttt{\textbf{SIM}} & \texttt{\textbf{NON}} & \texttt{\textbf{AUM}} & \texttt{\textbf{CIN}} & \texttt{\textbf{CON}} & \texttt{\textbf{SIM}} & \texttt{\textbf{NON}} & \texttt{\textbf{AUM}} & \texttt{\textbf{CIN}} & \texttt{\textbf{CON}} & \texttt{\textbf{SIM}} & \texttt{\textbf{NON}} & \texttt{\textbf{AUM}} & \texttt{\textbf{CIN}} & \texttt{\textbf{CON}} & \texttt{\textbf{SIM}} \\ \midrule

CIFAR-10 & 80.3 & 79.9 & 80.1 & 52.6 & 80.3 & 75.1 & 77.4 & 68.3 & 43.6 & 69.6 & 75.7 & 72.2 & 67.7 & 52.4 & 75.9 & 74.9 & 76.5 & 77.0 & 56.9 & 63.4 & 73.0 & 75.8 & 74.3 & 47.9 & 71.9  \\
Clothing-100K & 91.0 & 90.4 & 90.3 & 90.9 & 90.4 & 89.6 & 85.2 & 90.1 & 91.0 & 89.0 & 88.8 & 88.9 & 89.8 & 90.9 & 84.6 & 80.1 & 71.2 & 71.7 & 90.8 & 84.0 & 64.8 & 74.6 & 76.6 & 78.7 & 80.0 \\
NoisyCXR & 63.4 & 65.0 & 65.3 & 19.5 & 63.2 & 60.0 & 58.9 & 64.8 & 12.3 & 58.5 & 60.1 & 60.3 & 63.9 & 3.6 & 59.7 & 61.6 & 60.1 & 65.8 & 8.5 & 61.2 & 57.5 & 56.9 & 59.8 & 13.0 & 56.3 \\ \midrule
IMDb & 80.7 & 84.4 & 85.2 & 59.2 & 88.4 & 78.3 & 65.0 & 84.7 & 86.8 & 77.4 & 78.4 & 73.9 & 82.2 & 85.5 & 83.2 & 81.8 & 77.6 & 87.0 & 79.7 & 84.5 & 78.1 & 71.9 & 81.9 & 66.4 & 76.1  \\
TweetEval & 65.0 & 66.4 & 72.2 & 69.7 & 71.7 & 71.9 & 61.1 & 80.6 & 66.8 & 77.5 & 61.4 & 61.7 & 72.3 & 68.6 & 76.1 & 72.2 & 77.9 & 79.4 & 36.1 & 79.0 & 64.0 & 71.5 & 73.9 & 69.2 & 71.8 \\
\end{tabular}}
\caption{
Impact of label noise and each cleaning method on weighted $F_1$ score of a downstream model for each modality on the test set for noise rate $= 0.1$. The classification models used for images, text, tabular, and time series datasets are MobileNet-v2, \texttt{all-MiniLM-L6-v2}, Multi-layer perception (random seed 43), and Fully convolutional network, respectively.}
\end{table}

\begin{table}[h!]
\centering
\large
\resizebox{\columnwidth}{!}{
\begin{tabular}{c|ccccc|ccccc|ccccc|ccccc|ccccc}
\Xhline{1pt}
\multirow{2}{*}{\textbf{Datasets}} & \multicolumn{5}{c}{\textbf{No Noise Injected}} & \multicolumn{5}{c}{\textbf{Uniform}} & \multicolumn{5}{c}{\textbf{Asymmetric}} & \multicolumn{5}{c}{\textbf{Class-dependent}} & \multicolumn{5}{c}{\textbf{Instance-dependent}} \\ \cmidrule{2-26}
 & \texttt{\textbf{NON}} & \texttt{\textbf{AUM}} & \texttt{\textbf{CIN}} & \texttt{\textbf{CON}} & \texttt{\textbf{SIM}} & \texttt{\textbf{NON}} & \texttt{\textbf{AUM}} & \texttt{\textbf{CIN}} & \texttt{\textbf{CON}} & \texttt{\textbf{SIM}} & \texttt{\textbf{NON}} & \texttt{\textbf{AUM}} & \texttt{\textbf{CIN}} & \texttt{\textbf{CON}} & \texttt{\textbf{SIM}} & \texttt{\textbf{NON}} & \texttt{\textbf{AUM}} & \texttt{\textbf{CIN}} & \texttt{\textbf{CON}} & \texttt{\textbf{SIM}} & \texttt{\textbf{NON}} & \texttt{\textbf{AUM}} & \texttt{\textbf{CIN}} & \texttt{\textbf{CON}} & \texttt{\textbf{SIM}} \\ \midrule
CIFAR-10 & 61.5 & 61.4 & 58.4 & 61.5 & 60.6 & 52.0 & 53.3 & 53.9 & 53.5 & 52.9 & 52.7 & 54.9 & 53.1 & 52.6 & 54.2 & 57.7 & 58.2 & 54.8 & 56.9 & 58.6 & 50.4 & 49.5 & 52.4 & 48.5 & 49.0 \\
Clothing-100K & 90.9 & 90.6 & 90.2 & 90.7 & 90.8 & 81.6 & 69.3 & 88.9 & 87.1 & 85.7 & 84.4 & 88.1 & 85.1 & 87.0 & 90.8 & 72.1 & 88.5 & 84.8 & 70.1 & 83.7 & 74.7 & 84.8 & 36.9 & 83.3 & 87.2 \\
NoisyCXR & 39.7 & 49.3 & 40.4 & 45.5 & 43.3 & 40.8 & 39.2 & 37.4 & 39.8 & 37.5 & 40.0 & 37.8 & 40.7 & 41.8 & 39.8 & 35.7 & 38.7 & 36.3 & 38.3 & 34.5 & 37.3 & 37.8 & 35.9 & 39.7 & 36.9 \\ \midrule
Credit Fraud & 99.9 & 99.9 & 99.7 & 99.7 & 99.9 & 99.9 & 99.9 & 99.7 & 99.7 & 99.9 & 99.9 & 99.9 & 99.8 & 99.8 & 99.9 & 98.9 & 0.0 & 99.4 & 99.9 & 0.0 & 99.8 & 99.8 & 99.8 & 99.7 & 99.9 \\
Adult & 83.1 & 83.2 & 83.4 & 83.5 & 83.4 & 82.1 & 83.5 & 82.9 & 83.5 & 83.6 & 82.5 & 82.1 & 82.4 & 82.9 & 83.4 & 80.4 & 81.5 & 82.6 & 82.7 & 75.8 & 83.3 & 81.4 & 82.9 & 82.7 & 83.3 \\
\end{tabular}}
\caption{
Impact of label noise and each cleaning method on weighted $F_1$ score of a downstream model for each modality on the test set for noise rate $= 0.1$. The classification models used for images, tabular, and time series datasets are Fast-ViT-T8, TabTransformer, and PatchTST, respectively.}
\end{table}

\begin{table}[h!]
\centering
\large
\resizebox{\columnwidth}{!}{
\begin{tabular}{c|ccccc|ccccc|ccccc|ccccc|ccccc}
\Xhline{1pt}
\multirow{2}{*}{\textbf{Datasets}} & \multicolumn{5}{c}{\textbf{No Noise Injected}} & \multicolumn{5}{c}{\textbf{Uniform}} & \multicolumn{5}{c}{\textbf{Asymmetric}} & \multicolumn{5}{c}{\textbf{Class-dependent}} & \multicolumn{5}{c}{\textbf{Instance-dependent}} \\ \cmidrule{2-26}
 & \texttt{\textbf{NON}} & \texttt{\textbf{AUM}} & \texttt{\textbf{CIN}} & \texttt{\textbf{CON}} & \texttt{\textbf{SIM}} & \texttt{\textbf{NON}} & \texttt{\textbf{AUM}} & \texttt{\textbf{CIN}} & \texttt{\textbf{CON}} & \texttt{\textbf{SIM}} & \texttt{\textbf{NON}} & \texttt{\textbf{AUM}} & \texttt{\textbf{CIN}} & \texttt{\textbf{CON}} & \texttt{\textbf{SIM}} & \texttt{\textbf{NON}} & \texttt{\textbf{AUM}} & \texttt{\textbf{CIN}} & \texttt{\textbf{CON}} & \texttt{\textbf{SIM}} & \texttt{\textbf{NON}} & \texttt{\textbf{AUM}} & \texttt{\textbf{CIN}} & \texttt{\textbf{CON}} & \texttt{\textbf{SIM}} \\ \midrule

CIFAR-10 & 81.1 & 81.0 & 80.4 & 24.0 & 79.4 & 50.4 & 50.6 & 64.7 & 8.5 & 53.7 & 50.1 & 50.2 & 53.2 & 19.4 & 50.4 & 80.4 & 80.6 & 79.9 & 26.9 & 80.3 & 47.9 & 46.2 & 51.0 & 16.3 & 49.2 \\
Clothing-100K & 90.9 & 91.0 & 91.1 & 90.9 & 90.9 & 61.3 & 62.3 & 57.7 & 79.7 & 67.2 & 70.9 & 60.8 & 63.8 & 28.9 & 70.9 & 90.0 & 89.6 & 88.7 & 90.2 & 90.3 & 82.4 & 64.3 & 50.4 & 73.8 & 70.3  \\ 
NoisyCXR & 65.4 & 65.3 & 64.7 & 10.4 & 64.5 & 44.2 & 44.1 & 56.9 & 10.1 & 43.4 & 41.1 & 39.8 & 45.5 & 9.5 & 40.6 & 65.0 & 65.7 & 65.0 & 7.2 & 65.8 & 40.5 & 40.4 & 47.6 & 7.4 & 39.3\\ \midrule
IMDb & 89.1 & 90.5 & 93.1 & 80.0 & 92.1 & 33.3 & 33.3 & 33.3 & 33.3 & 33.3 & 33.3 & 33.3 & 77.8 & 33.3 & 66.6 & 92.4 & 92.2 & 91.1 & 91.3 & 89.6 & 33.3 & 33.3 & 33.3 & 33.3 & 33.3 \\
TweetEval & 82.1 & 80.6 & 81.9 & 60.4 & 81.8 & 60.4 & 74.0 & 60.4 & 23.5 & 60.4 & 60.4 & 60.4 & 60.4 & 12.1 & 60.4 & 81.6 & 82.3 & 77.9 & 66.7 & 76.8 & 60.4 & 60.4 & 60.4 & 60.4 & 60.4 \\
\end{tabular}}
\caption{
Impact of label noise and each cleaning method on weighted $F_1$ score of a downstream model for each modality on the test set for noise rate $= 0.4$. The classification models used for images, text, tabular, and time series datasets are ResNet-18, \texttt{all-distilroberta-v1}, Multi-layer perception (random seed 42), and ResNet-1D, respectively.}
\end{table}

\pagebreak

\begin{table}[h!]
\centering
\large
\resizebox{\columnwidth}{!}{
\begin{tabular}{c|ccccc|ccccc|ccccc|ccccc|ccccc}
\Xhline{1pt}
\multirow{2}{*}{\textbf{Datasets}} & \multicolumn{5}{c}{\textbf{No Noise Injected}} & \multicolumn{5}{c}{\textbf{Uniform}} & \multicolumn{5}{c}{\textbf{Asymmetric}} & \multicolumn{5}{c}{\textbf{Class-dependent}} & \multicolumn{5}{c}{\textbf{Instance-dependent}} \\ \cmidrule{2-26}
 & \texttt{\textbf{NON}} & \texttt{\textbf{AUM}} & \texttt{\textbf{CIN}} & \texttt{\textbf{CON}} & \texttt{\textbf{SIM}} & \texttt{\textbf{NON}} & \texttt{\textbf{AUM}} & \texttt{\textbf{CIN}} & \texttt{\textbf{CON}} & \texttt{\textbf{SIM}} & \texttt{\textbf{NON}} & \texttt{\textbf{AUM}} & \texttt{\textbf{CIN}} & \texttt{\textbf{CON}} & \texttt{\textbf{SIM}} & \texttt{\textbf{NON}} & \texttt{\textbf{AUM}} & \texttt{\textbf{CIN}} & \texttt{\textbf{CON}} & \texttt{\textbf{SIM}} & \texttt{\textbf{NON}} & \texttt{\textbf{AUM}} & \texttt{\textbf{CIN}} & \texttt{\textbf{CON}} & \texttt{\textbf{SIM}} \\ \midrule
CIFAR-10 & 80.3 & 79.9 & 80.1 & 52.6 & 80.3 & 64.1 & 55.7 & 65.4 & 29.4 & 65.9 & 13.3 & 63.9 & 58.5 & 29.5 & 59.5 & 74.9 & 76.6 & 77.0 & 56.9 & 63.4 & 35.3 & 57.8 & 59.9 & 22.7 & 60.4 \\
Clothing-100K & 91.0 & 90.5 & 90.3 & 90.9 & 90.5 & 70.0 & 62.4 & 79.2 & 77.7 & 59.1 & 73.8 & 69.7 & 47.3 & 88.3 & 64.9 & 80.1 & 71.3 & 71.8 & 90.9 & 84.0 & 63.6 & 28.0 & 69.1 & 70.4 & 63.1  \\ 
NoisyCXR & 63.4 & 65.0 & 65.4 & 19.6 & 63.3 & 44.9 & 45.4 & 53.0 & 8.6 & 42.4 & 43.1 & 38.9 & 43.5 & 9.6 & 38.9 & 61.7 & 60.2 & 65.9 & 8.5 & 61.2 & 36.1 & 36.8 & 40.2 & 8.8 & 40.8 \\ 
\midrule
IMDb & 80.8 & 84.5 & 85.2 & 59.2 & 88.4 & 33.3 & 33.3 & 33.3 & 33.3 & 43.3 & 77.3 & 33.3 & 45.3 & 33.3 & 63.5 & 81.8 & 77.6 & 87.0 & 79.8 & 84.6 & 33.3 & 33.3 & 33.3 & 33.3 & 33.3 \\
TweetEval & 65.0 & 66.5 & 72.3 & 69.7 & 71.8 & 69.9 & 60.7 & 74.2 & 12.2 & 60.4 & 60.4 & 73.0 & 60.4 & 60.4 & 60.4 & 72.2 & 77.9 & 79.5 & 36.1 & 79.0 & 60.4 & 60.4 & 59.9 & 60.4 & 12.2 \\
\end{tabular}}
\caption{
Impact of label noise and each cleaning method on weighted $F_1$ score of a downstream model for each modality on the test set for noise rate $= 0.4$. The classification models used for images, text, tabular, and time series datasets are MobileNet-v2, \texttt{all-MiniLM-L6-v2}, Multi-layer perception (random seed 43), and Fully convolutional network, respectively.}
\end{table}

\begin{table}[h!]
\centering
\large
\resizebox{\columnwidth}{!}{
\begin{tabular}{c|ccccc|ccccc|ccccc|ccccc|ccccc}
\Xhline{1pt}
\multirow{2}{*}{\textbf{Datasets}} & \multicolumn{5}{c}{\textbf{No Noise Injected}} & \multicolumn{5}{c}{\textbf{Uniform}} & \multicolumn{5}{c}{\textbf{Asymmetric}} & \multicolumn{5}{c}{\textbf{Class-dependent}} & \multicolumn{5}{c}{\textbf{Instance-dependent}} \\ \cmidrule{2-26}
 & \texttt{\textbf{NON}} & \texttt{\textbf{AUM}} & \texttt{\textbf{CIN}} & \texttt{\textbf{CON}} & \texttt{\textbf{SIM}} & \texttt{\textbf{NON}} & \texttt{\textbf{AUM}} & \texttt{\textbf{CIN}} & \texttt{\textbf{CON}} & \texttt{\textbf{SIM}} & \texttt{\textbf{NON}} & \texttt{\textbf{AUM}} & \texttt{\textbf{CIN}} & \texttt{\textbf{CON}} & \texttt{\textbf{SIM}} & \texttt{\textbf{NON}} & \texttt{\textbf{AUM}} & \texttt{\textbf{CIN}} & \texttt{\textbf{CON}} & \texttt{\textbf{SIM}} & \texttt{\textbf{NON}} & \texttt{\textbf{AUM}} & \texttt{\textbf{CIN}} & \texttt{\textbf{CON}} & \texttt{\textbf{SIM}} \\ \midrule
CIFAR-10 & 61.5 & 61.4 & 58.4 & 61.5 & 60.6 & 34.5 & 33.5 & 37.0 & 34.7 & 32.8 & 36.0 & 34.7 & 34.7 & 35.6 & 34.8 & 57.7 & 58.2 & 54.8 & 56.9 & 58.6 & 31.5 & 31.9 & 31.5 & 31.3 & 32.5 \\
Clothing-100K & 90.9 & 90.6 & 90.2 & 90.7 & 90.8 & 80.4 & 77.6 & 78.9 & 70.9 & 87.5 & 66.7 & 63.2 & 65.1 & 59.1 & 76.1 & 72.1 & 88.5 & 84.8 & 70.1 & 83.7 & 68.2 & 72.9 & 64.3 & 69.7 & 80.0\\
NoisyCXR & 31.4 & 32.0 & 32.7 & 45.5 & 43.3 & 23.7 & 24.3 & 23.5 & 24.5 & 17.0 & 27.5 & 29.5 & 27.4 & 27.8 & 23.4 & 35.7 & 38.7 & 36.3 & 38.3 & 34.5 & 22.1 & 23.4 & 24.8 & 24.5 & 22.9 \\ \midrule
Credit Fraud & 99.9 & 99.9 & 99.7 & 99.7 & 99.9 & 99.7 & 99.7 & 99.7 & 99.7 & 99.7 & 99.8 & 99.7 & 99.7 & 99.7 & 99.7 & 98.9 & 0.0 & 99.4 & 99.9 & 0.0 & 99.7 & 99.7 & 99.7 & 0.0 & 99.7 \\
Adult & 83.1 & 83.2 & 83.4 & 83.5 & 83.4 & 71.8 & 70.4 & 78.7 & 76.6 & 66.2 & 80.9 & 80.1 & 80.1 & 68.6 & 79.7 & 80.4 & 81.5 & 82.6 & 82.7 & 75.8 & 76.8 & 66.1 & 66.6 & 69.6 & 66.1 \\
\end{tabular}}
\caption{
Impact of label noise and each cleaning method on weighted $F_1$ score of a downstream model for each modality on the test set for noise rate $= 0.4$. The classification models used for images, tabular, and time series datasets are Fast-ViT-T8, TabTransformer, and PatchTST, respectively.}
\end{table}

\begin{table}[h!]
\centering
\large
\resizebox{\columnwidth}{!}{
\begin{tabular}{c|ccccc|ccccc|ccccc|ccccc|ccccc}
\Xhline{1pt}
\multirow{2}{*}{\textbf{Datasets}} & \multicolumn{5}{c}{\textbf{No Noise Injected}} & \multicolumn{5}{c}{\textbf{Uniform}} & \multicolumn{5}{c}{\textbf{Asymmetric}} & \multicolumn{5}{c}{\textbf{Class-dependent}} & \multicolumn{5}{c}{\textbf{Instance-dependent}} \\ \cmidrule{2-26}
 & \texttt{\textbf{NON}} & \texttt{\textbf{AUM}} & \texttt{\textbf{CIN}} & \texttt{\textbf{CON}} & \texttt{\textbf{SIM}} & \texttt{\textbf{NON}} & \texttt{\textbf{AUM}} & \texttt{\textbf{CIN}} & \texttt{\textbf{CON}} & \texttt{\textbf{SIM}} & \texttt{\textbf{NON}} & \texttt{\textbf{AUM}} & \texttt{\textbf{CIN}} & \texttt{\textbf{CON}} & \texttt{\textbf{SIM}} & \texttt{\textbf{NON}} & \texttt{\textbf{AUM}} & \texttt{\textbf{CIN}} & \texttt{\textbf{CON}} & \texttt{\textbf{SIM}} & \texttt{\textbf{NON}} & \texttt{\textbf{AUM}} & \texttt{\textbf{CIN}} & \texttt{\textbf{CON}} & \texttt{\textbf{SIM}} \\ \midrule
CIFAR-10 & 81.1 & 81.1 & 80.5 & 24.0 & 79.4 & 80.1 & 79.8 & 79.6 & 20.2 & 79.9 & 78.3 & 79.0 & 78.1 & 18.9 & 80.0 & 80.3 & 79.9 & 78.9 & 20.6 & 80.7 & 77.5 & 74.5 & 78.0 & 12.0 & 39.2 \\
Clothing-100K & 90.9 & 91.0 & 91.1 & 90.9 & 91.0 & 90.8 & 90.9 & 90.6 & 90.9 & 90.9 & 90.9 & 90.9 & 90.9 & 90.9 & 90.9 & 90.9 & 91.0 & 91.1 & 90.9 & 90.9 & 90.3 & 90.2 & 90.4 & 90.9 & 90.6 \\
NoisyCXR & 65.5 & 65.3 & 64.7 & 10.4 & 64.6 & 63.7 & 64.3 & 64.8 & 10.3 & 64.3 & 64.3 & 64.3 & 65.0 & 11.8 & 64.5 & 65.3 & 65.3 & 65.3 & 12.8 & 65.4 & 62.5 & 62.5 & 63.8 & 11.0 & 62.3 \\ \midrule
IMDb & 80.8 & 84.5 & 85.2 & 59.2 & 88.4 & 85.7 & 86.1 & 74.1 & 61.2 & 89.0 & 78.7 & 83.1 & 76.1 & 68.6 & 87.0 & 80.9 & 87.2 & 85.5 & 73.1 & 83.4 & 66.7 & 51.9 & 86.3 & 77.8 & 80.8 \\
TweetEval & 65.0 & 66.5 & 72.3 & 69.7 & 71.8 & 61.2 & 79.7 & 67.2 & 66.1 & 78.5 & 81.9 & 69.7 & 73.3 & 67.9 & 73.8 & 74.6 & 47.2 & 26.6 & 12.2 & 12.2 & 66.2 & 66.8 & 77.1 & 78.1 & 74.3  \\
\end{tabular}}
\caption{
Impact of label noise and each cleaning method on weighted $F_1$ score of a downstream model for each modality on the test set for noise rate $= 0.02$. The classification models used for images, text, tabular, and time series datasets are ResNet-18, \texttt{all-distilroberta-v1}, Multi-layer perception (random seed 42), and ResNet-1D, respectively.}
\end{table}

\begin{table}[h!]
\centering
\large
\resizebox{\columnwidth}{!}{
\begin{tabular}{c|ccccc|ccccc|ccccc|ccccc|ccccc}
\Xhline{1pt}
\multirow{2}{*}{\textbf{Datasets}} & \multicolumn{5}{c}{\textbf{No Noise Injected}} & \multicolumn{5}{c}{\textbf{Uniform}} & \multicolumn{5}{c}{\textbf{Asymmetric}} & \multicolumn{5}{c}{\textbf{Class-dependent}} & \multicolumn{5}{c}{\textbf{Instance-dependent}} \\ \cmidrule{2-26}
 & \texttt{\textbf{NON}} & \texttt{\textbf{AUM}} & \texttt{\textbf{CIN}} & \texttt{\textbf{CON}} & \texttt{\textbf{SIM}} & \texttt{\textbf{NON}} & \texttt{\textbf{AUM}} & \texttt{\textbf{CIN}} & \texttt{\textbf{CON}} & \texttt{\textbf{SIM}} & \texttt{\textbf{NON}} & \texttt{\textbf{AUM}} & \texttt{\textbf{CIN}} & \texttt{\textbf{CON}} & \texttt{\textbf{SIM}} & \texttt{\textbf{NON}} & \texttt{\textbf{AUM}} & \texttt{\textbf{CIN}} & \texttt{\textbf{CON}} & \texttt{\textbf{SIM}} & \texttt{\textbf{NON}} & \texttt{\textbf{AUM}} & \texttt{\textbf{CIN}} & \texttt{\textbf{CON}} & \texttt{\textbf{SIM}} \\ \midrule
CIFAR-10 & 80.3 & 79.9 & 80.1 & 52.6 & 80.3 & 79.6 & 80.1 & 80.6 & 64.8 & 81.3 & 81.0 & 80.5 & 80.3 & 63.9 & 79.8 & 74.9 & 76.6 & 77.0 & 56.9 & 63.4 & 76.1 & 78.9 & 79.4 & 59.6 & 79.4 \\
Clothing-100K & 91.0 & 90.5 & 90.3 & 90.9 & 90.5 & 90.6 & 90.9 & 90.9 & 90.9 & 90.5 & 90.9 & 90.5 & 90.5 & 90.9 & 90.5 & 80.1 & 71.3 & 71.8 & 90.9 & 84.0 & 88.2 & 89.3 & 90.1 & 90.9 & 89.3 \\ 
NoisyCXR & 63.4 & 65.0 & 65.4 & 19.6 & 63.3 & 62.6 & 62.8 & 65.3 & 13.2 & 63.2 & 63.1 & 63.2 & 64.2 & 2.1 & 62.4 & 61.7 & 60.2 & 65.9 & 8.5 & 61.2 & 61.3 & 60.2 & 61.8 & 12.4 & 61.5 \\ \midrule
IMDb & 89.1 & 90.6 & 93.1 & 80.0 & 92.2 & 92.4 & 83.4 & 91.2 & 91.5 & 92.3 & 90.5 & 92.2 & 91.4 & 88.8 & 90.9 & 93.0 & 85.9 & 92.6 & 91.2 & 92.6 & 90.3 & 91.7 & 91.1 & 88.3 & 84.4 \\
TweetEval & 82.1 & 80.7 & 81.9 & 60.4 & 81.8 & 81.8 & 82.1 & 82.9 & 75.1 & 80.1 & 81.8 & 78.8 & 83.2 & 81.9 & 82.2 & 84.0 & 80.9 & 82.1 & 80.8 & 80.5 & 80.1 & 79.5 & 80.3 & 81.1 & 81.2 \\
\end{tabular}}
\caption{
Impact of label noise and each cleaning method on weighted $F_1$ score of a downstream model for each modality on the test set for noise rate $= 0.02$. The classification models used for images, text, tabular, and time series datasets are MobileNet-v2, \texttt{all-MiniLM-L6-v2}, Multi-layer perception (random seed 43), and Fully convolutional network, respectively.}
\end{table}

\begin{table}[h!]
\centering
\large
\resizebox{\columnwidth}{!}{
\begin{tabular}{c|ccccc|ccccc|ccccc|ccccc|ccccc}
\Xhline{1pt}
\multirow{2}{*}{\textbf{Datasets}} & \multicolumn{5}{c}{\textbf{No Noise Injected}} & \multicolumn{5}{c}{\textbf{Uniform}} & \multicolumn{5}{c}{\textbf{Asymmetric}} & \multicolumn{5}{c}{\textbf{Class-dependent}} & \multicolumn{5}{c}{\textbf{Instance-dependent}} \\ \cmidrule{2-26}
 & \texttt{\textbf{NON}} & \texttt{\textbf{AUM}} & \texttt{\textbf{CIN}} & \texttt{\textbf{CON}} & \texttt{\textbf{SIM}} & \texttt{\textbf{NON}} & \texttt{\textbf{AUM}} & \texttt{\textbf{CIN}} & \texttt{\textbf{CON}} & \texttt{\textbf{SIM}} & \texttt{\textbf{NON}} & \texttt{\textbf{AUM}} & \texttt{\textbf{CIN}} & \texttt{\textbf{CON}} & \texttt{\textbf{SIM}} & \texttt{\textbf{NON}} & \texttt{\textbf{AUM}} & \texttt{\textbf{CIN}} & \texttt{\textbf{CON}} & \texttt{\textbf{SIM}} & \texttt{\textbf{NON}} & \texttt{\textbf{AUM}} & \texttt{\textbf{CIN}} & \texttt{\textbf{CON}} & \texttt{\textbf{SIM}} \\ \midrule
CIFAR-10 & 61.5 & 61.4 & 58.4 & 61.5 & 60.6 & 58.8 & 59.5 & 57.5 & 58.4 & 58.5 & 60.9 & 58.5 & 57.5 & 59.7 & 58.1 & 57.7 & 58.2 & 54.8 & 56.9 & 58.6 & 55.2 & 55.6 & 56.2 & 54.5 & 55.8 \\
Clothing-100K & 90.9 & 90.6 & 90.2 & 90.7 & 90.8 & 90.5 & 90.1 & 90.8 & 90.8 & 90.3 & 89.1 & 90.8 & 90.5 & 88.0 & 89.5 & 72.1 & 88.5 & 84.8 & 70.1 & 83.7 & 89.9 & 89.5 & 80.8 & 90.1 & 88.8 \\
NoisyCXR & 46.2 & 36.2 & 46.6 & 45.5 & 43.3 & 44.6 & 44.3 & 40.5 & 45.6 & 45.7 & 45.4 & 43.3 & 41.2 & 45.8 & 40.4 & 35.7 & 38.7 & 36.3 & 38.3 & 34.5 & 40.4 & 43.0 & 38.8 & 39.7 & 40.9 \\ \midrule
Credit Fraud & 99.9 & 99.9 & 99.7 & 99.7 & 99.9 & 100 & 99.9 & 99.9 & 99.7 & 99.9 & 99.7 & 99.9 & 99.9 & 99.8 & 100 & 98.9 & 0.0 & 99.4 & 99.9 & 0.0 & 99.9 & 99.8 & 99.9 & 99.5 & 99.7 \\
Adult & 83.1 & 83.2 & 83.4 & 83.5 & 83.4 & 82.6 & 82.6 & 83.4 & 83.5 & 83.7 & 83.8 & 83.5 & 82.8 & 81.8 & 82.9 & 80.4 & 81.5 & 82.6 & 82.7 & 75.8 & 83.8 & 78.9 & 83.1 & 80.7 & 83.6 \\
\end{tabular}}
\caption{
Impact of label noise and each cleaning method on weighted $F_1$ score of a downstream model for each modality on the test set for noise rate $= 0.02$. The classification models used for images, tabular, and time series datasets are Fast-ViT-T8, TabTransformer, and PatchTST, respectively.}
\end{table}

\pagebreak

\subsubsection{Critical Difference Diagrams}

To compare cleaning methods and downstream classifiers across multiple datasets, we follow the recommendations of \citet{demvsar2006statistical}. First, we use the Friedman test \citep{friedman} to evaluate whether a statistically significant difference exists between classifiers’ performance. Then, for classifiers with significantly different performance, we conduct pairwise post-hoc analysis recommended by \citet{benavoli2016should} where the average rank comparison is replaced with the Wilcoxon signed-rank test \citep{wilcoxon1945} with Holm’s alpha correction \citep{holm}. The thick horizontal line in a critical difference diagram shows models that are not significantly different in performance.

\begin{figure}[h!]
\setlength{\tabcolsep}{1.5pt}
\centering
\begin{tabular}{cc}
  \includegraphics[width=0.5\textwidth,trim={1.2cm 0.5cm 1.3cm 0.5cm},clip]{images/new_cd/q1.pdf} &
  \includegraphics[width=0.5\textwidth,trim={1.2cm 0.4cm 1.3cm 0.5cm},clip]{images/new_cd/q2.pdf} \\
  (\textit{i}) & (\textit{ii}) 
\end{tabular}
\caption{Ranking of cleaning methods across all datasets, base classification models, synthetic noise types, noise rates, random seeds in terms of (\textit{i}) their ability to identify labeling errors measured using weighted $F_1$, (\textit{ii}) the weighted $F_1$ of downstream models trained on their cleaned data}
\label{fig:CD_1}
\end{figure}

\begin{figure}[h!]
\setlength{\tabcolsep}{1.5pt}
\centering
\begin{tabular}{cc}
  \includegraphics[width=0.5\textwidth,trim={1.2cm 0.5cm 1.3cm 0.5cm},clip]{images/new_cd/q5c_i2.pdf} &
  \includegraphics[width=0.5\textwidth,trim={1.2cm 0.4cm 1.3cm 0.5cm},clip]{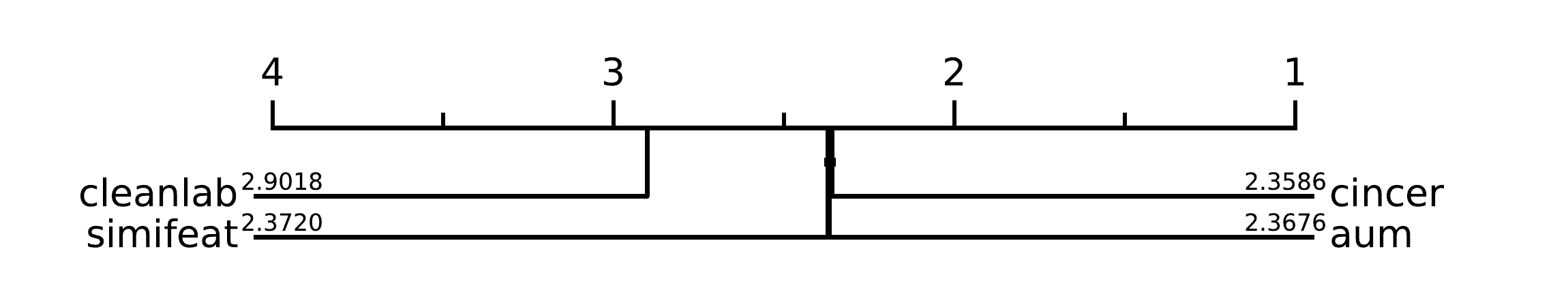} \\
  (\textit{i}) & (\textit{ii}) 
\end{tabular}
\caption{Ranking of cleaning methods across all datasets, base classification models, synthetic noise types, noise rates, random seeds in terms of (\textit{i}) their ability to identify labeling errors measured using \textit{accuracy}, (\textit{ii}) the \textit{accuracy} of downstream models trained on their cleaned data.}
\label{fig:CD_4}
\end{figure}

\begin{figure}[h!]
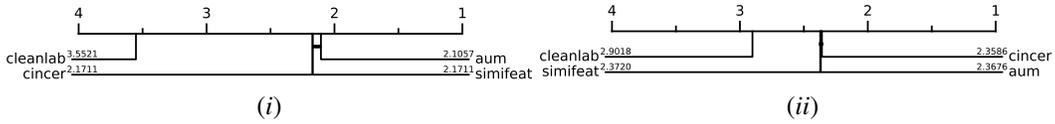

\setlength{\tabcolsep}{1.5pt}
\centering
\begin{tabular}{cc}
  \includegraphics[width=0.5\textwidth,trim={1.2cm 0.5cm 1.3cm 0.5cm},clip]{images/new_cd/q5c_i2.pdf} &
  \includegraphics[width=0.5\textwidth,trim={1.2cm 0.4cm 1.3cm 0.5cm},clip]{images/new_cd/q5c_ii2.pdf} \\
  (\textit{i}) & (\textit{ii}) 
\end{tabular}
\caption{Ranking of synthetic noise types by their ability to impact the (\textit{i}) performance of cleaning methods, (\textit{ii}) \textit{weighted $F_1$} of downstream models trained on cleaned datasets.}
\label{fig:CD_5}
\end{figure}

\begin{figure}[h!]
\setlength{\tabcolsep}{1.5pt}
\centering
\begin{tabular}{cc}

  \includegraphics[width=0.5\textwidth,trim={1.2cm 0.5cm 1.3cm 0.5cm},clip]{images/new_cd/q5ai-image.pdf} &
  \includegraphics[width=0.5\textwidth,trim={1.2cm 0.4cm 1.3cm 0.5cm},clip]{images/new_cd/q5ai-tabular.pdf} \\
  \textit{image} & \textit{tabular} \\
  \includegraphics[width=0.5\textwidth,trim={1.2cm 0.4cm 1.3cm 0.5cm},clip]{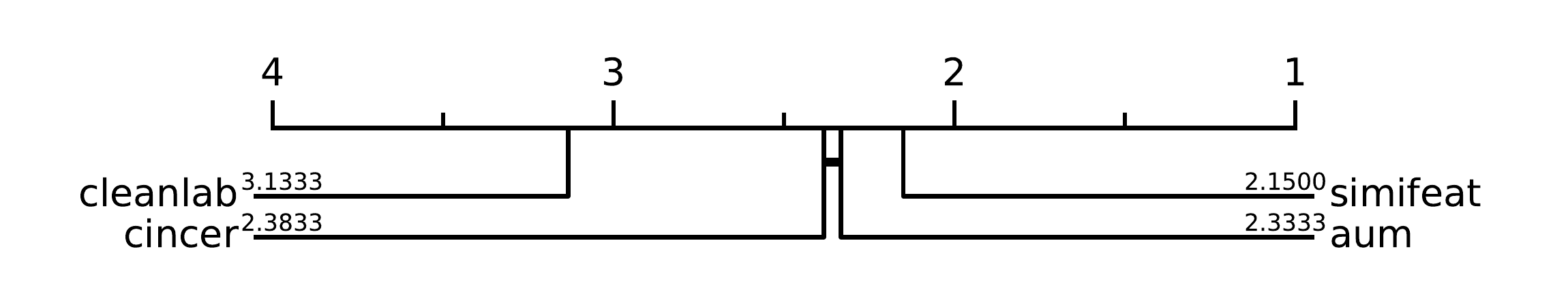} & 
  \includegraphics[width=0.5\textwidth,trim={1.2cm 0.4cm 1.3cm 0.5cm},clip]{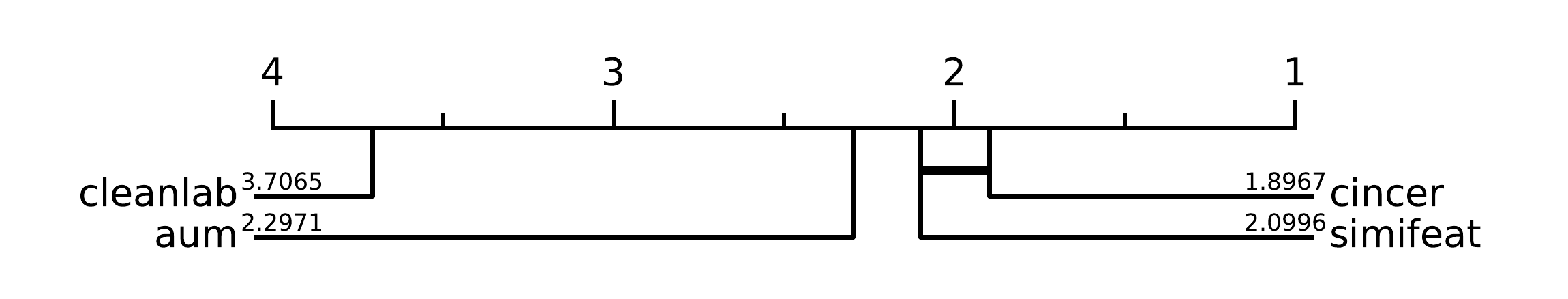} \\
  \textit{text} & \textit{time-series} \\
  \multicolumn{2}{c}{(\textit{i}) Rankings by the ability to identify labeling errors measured using weighted $F_1$} \\
  
  \includegraphics[width=0.5\textwidth,trim={1.2cm 0.5cm 1.3cm 0.5cm},clip]{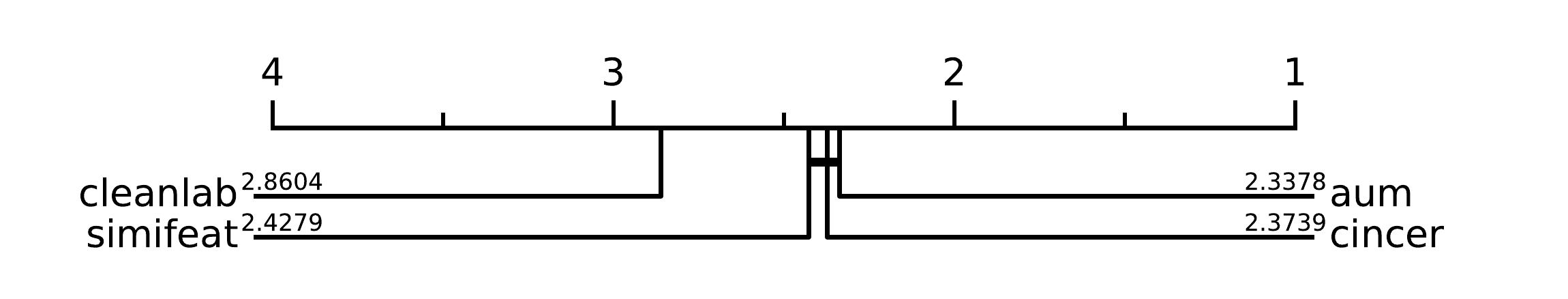} &
  \includegraphics[width=0.5\textwidth,trim={1.2cm 0.4cm 1.3cm 0.5cm},clip]{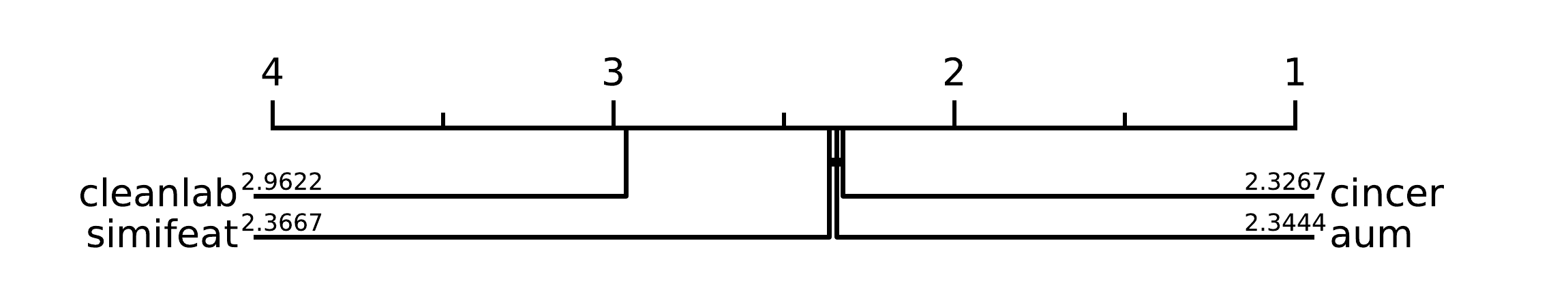} \\
  \textit{image} & \textit{tabular} \\
  \includegraphics[width=0.5\textwidth,trim={1.2cm 0.4cm 1.3cm 0.5cm},clip]{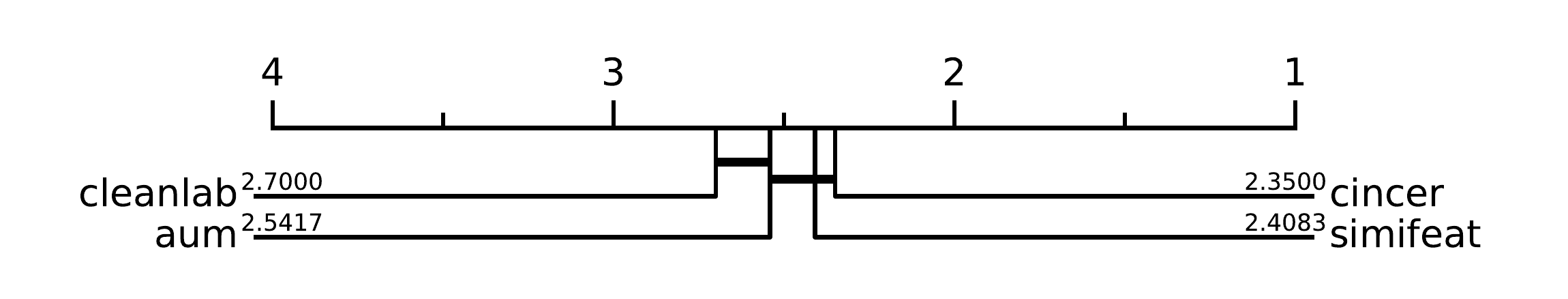} & 
  \includegraphics[width=0.5\textwidth,trim={1.2cm 0.4cm 1.3cm 0.5cm},clip]{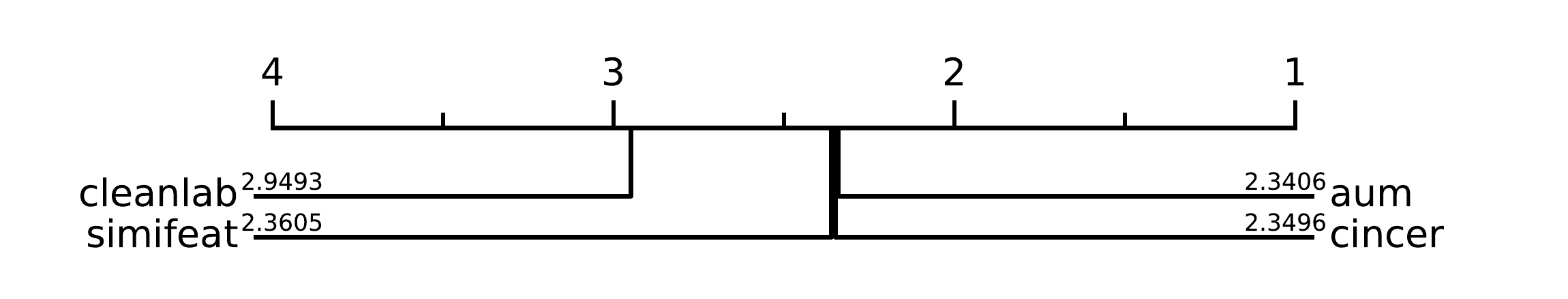} \\
  \textit{text} & \textit{time-series} \\
  \multicolumn{2}{c}{(\textit{ii}) Rankings by the weighted $F_1$ of downstream models trained on their cleaned data} \\

\end{tabular}
\caption{Rankings of cleaning methods segmented by data modality.}
\label{fig:CD_3}
\end{figure}

\begin{figure}[h!]
\setlength{\tabcolsep}{1.5pt}
\centering
\begin{tabular}{cc}

  \includegraphics[width=0.5\textwidth,trim={1.2cm 0.5cm 1.3cm 0.5cm},clip]{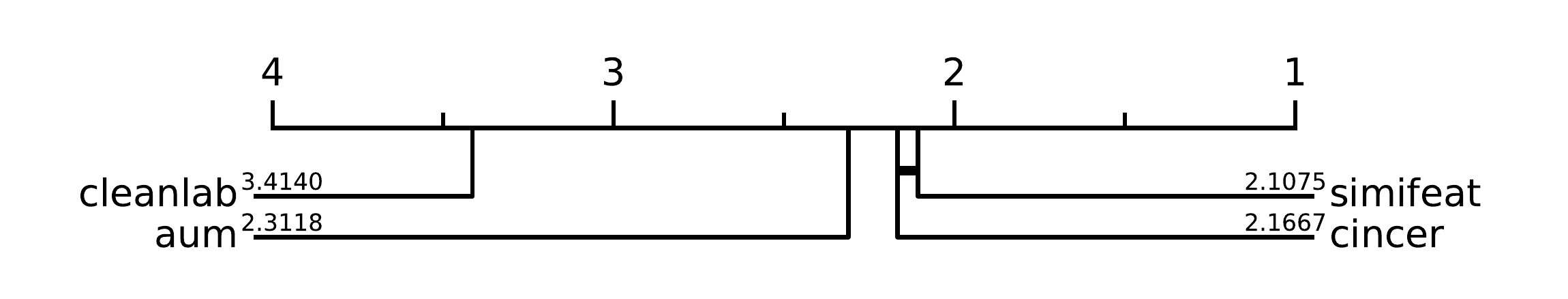} &
  \includegraphics[width=0.5\textwidth,trim={1.2cm 0.4cm 1.3cm 0.5cm},clip]{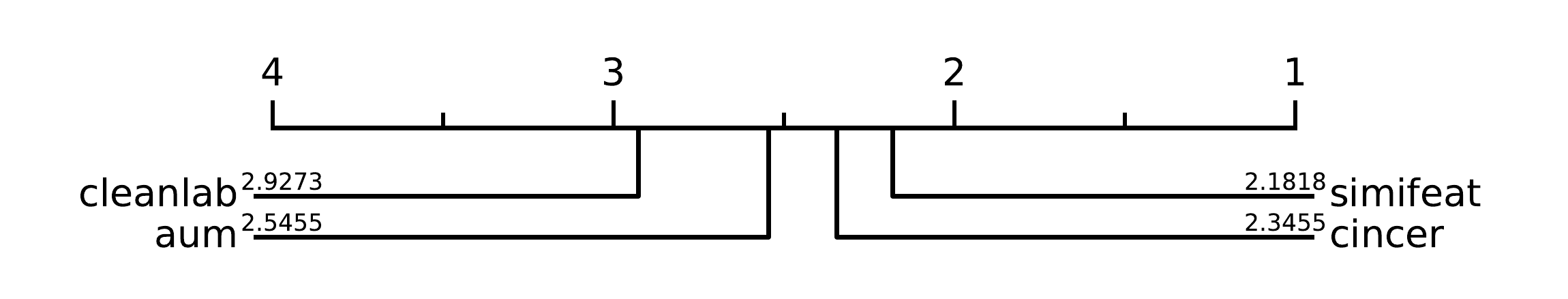} \\
  \textit{asymmetric} & \textit{class dependent} \\
  \includegraphics[width=0.5\textwidth,trim={1.2cm 0.4cm 1.3cm 0.5cm},clip]{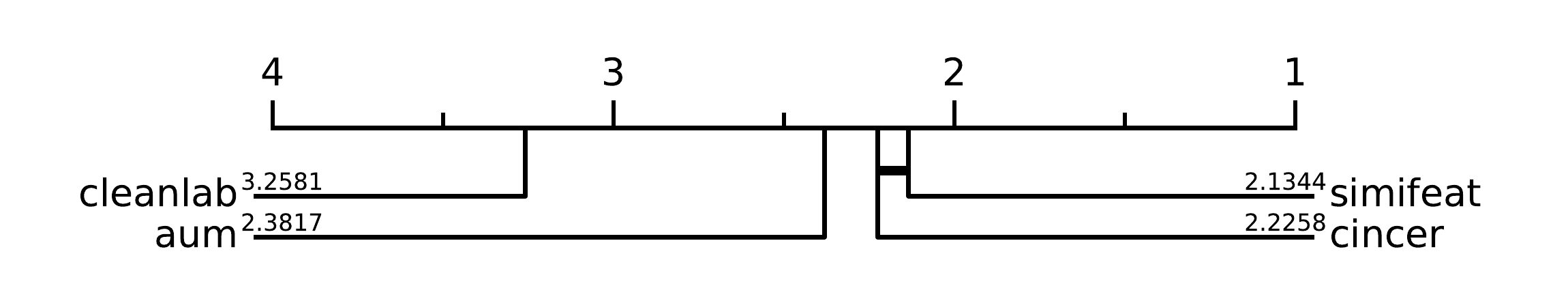} & 
  \includegraphics[width=0.5\textwidth,trim={1.2cm 0.4cm 1.3cm 0.5cm},clip]{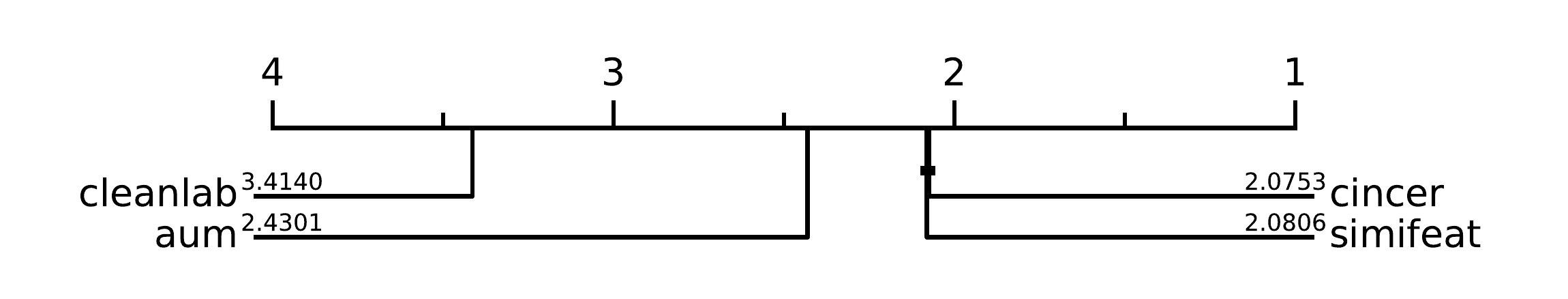} \\
  \textit{instance dependent} & \textit{uniform} \\
  \multicolumn{2}{c}{(\textit{i}) Rankings by the ability to identify labeling errors measured using weighted $F_1$.} \\
  
  \includegraphics[width=0.5\textwidth,trim={1.2cm 0.5cm 1.3cm 0.5cm},clip]{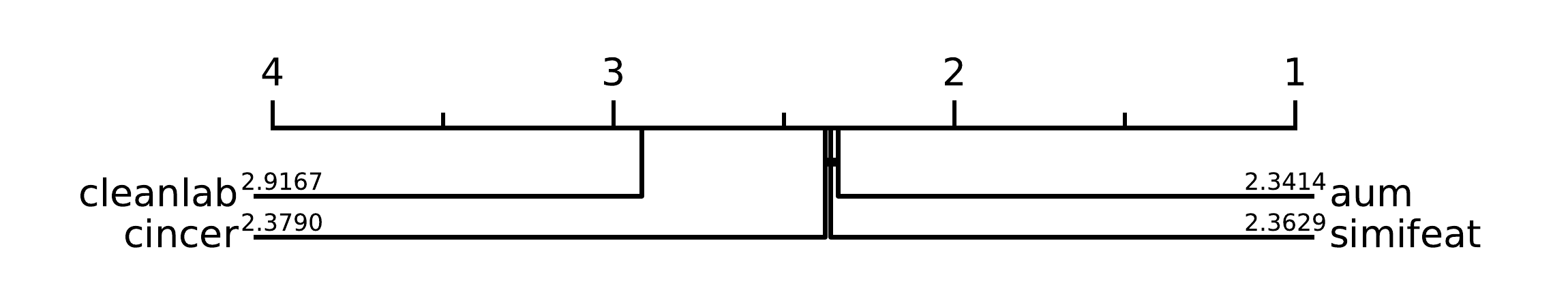} &
  \includegraphics[width=0.5\textwidth,trim={1.2cm 0.4cm 1.3cm 0.5cm},clip]{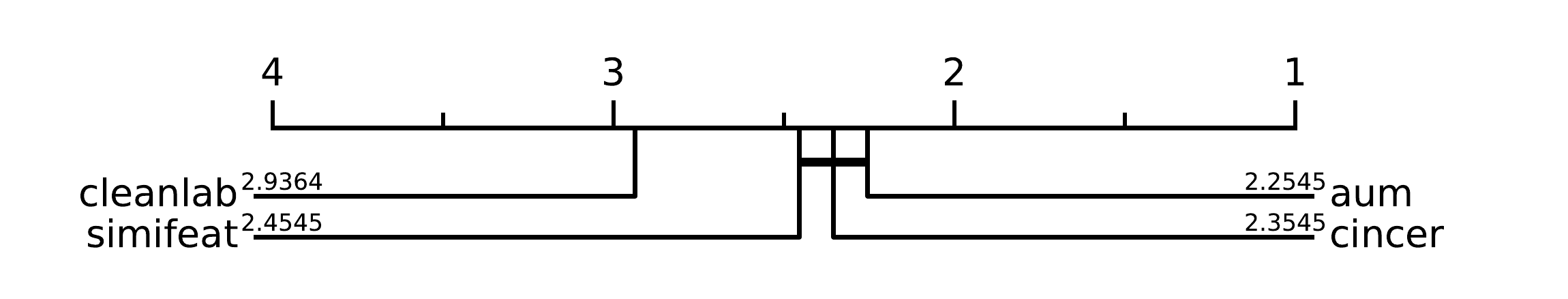} \\
  \textit{asymmetric} & \textit{class dependent} \\
  \includegraphics[width=0.5\textwidth,trim={1.2cm 0.4cm 1.3cm 0.5cm},clip]{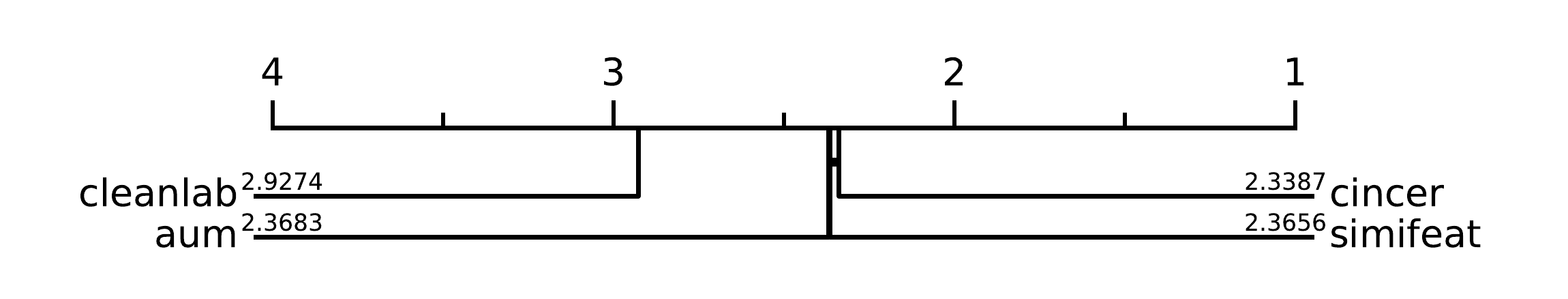} & 
  \includegraphics[width=0.5\textwidth,trim={1.2cm 0.4cm 1.3cm 0.5cm},clip]{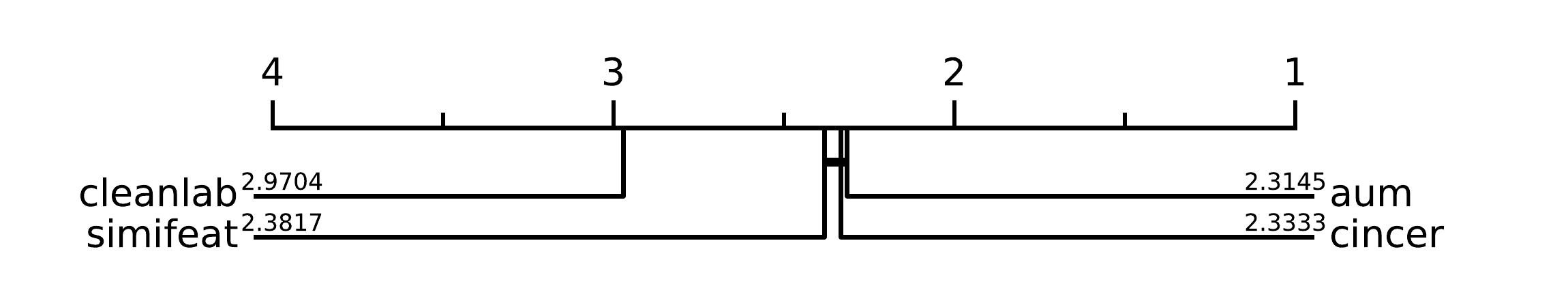} \\
  \textit{instance dependent} & \textit{uniform} \\
  \includegraphics[width=0.5\textwidth,trim={1.2cm 0.4cm 1.3cm 0.5cm},clip]{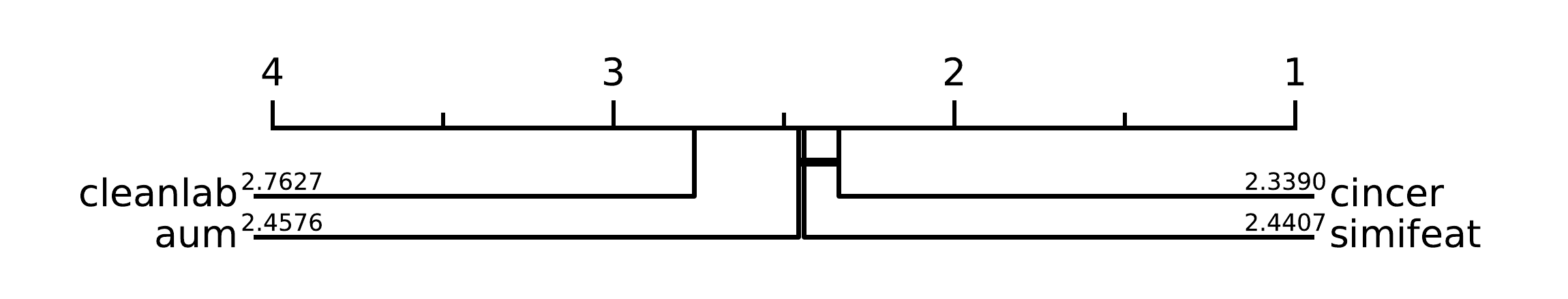} & \\
  \textit{no noise} & \\
  \multicolumn{2}{c}{(\textit{ii}) Rankings by the weighted $F_1$ of downstream models trained on their cleaned data.} \\

\end{tabular}
\caption{Rankings of cleaning methods segmented by synthetic noise type.}
\label{fig:CD_2}
\end{figure}

\vfill

\pagebreak

\subsubsection{Effects of Cleaning Methods on Data Distribution}

\begin{table}[h!]
\centering
\resizebox{0.4\textwidth}{!}{
\begin{tabular}{c c c c c}
\Xhline{1pt}
\textbf{Datasets} & \texttt{\textbf{AUM}} & \texttt{\textbf{CIN}} & \texttt{\textbf{CON}} & \texttt{\textbf{SIM}}\\
\midrule
\textbf{Average} & 0.011 & 0.133 & 0.574 & 0.061 \\
\textbf{Std Dev} & 0.001 & 0.115 & 0.248 & 0.038 \\
\textbf{Median} & 0.011 & 0.095 & 0.478 & 0.057 \\
\Xhline{1pt}
\end{tabular}}
\caption{Proportion of data points cleaned by each cleaning method,  averaged over noise type, noise rate, random seed(s), and downstream model architecture and datasets. Confident Learning removes 57\% of training data points on average, explaining its poor performance. All other methods remove <13\% data points. It is probably correct that for confident learning the downstream models are not seeing enough data or trained long enough for models to converge. But we believe that this might be a problem of the cleaning method, more than the experiment design.}
\label{tab:prop_samples_cleaned_1}
\end{table}

\begin{table}[h!]
\centering
\resizebox{0.4\textwidth}{!}{
\begin{tabular}{c c c c c}
\Xhline{1pt}
\textbf{Datasets} & \texttt{\textbf{AUM}} & \texttt{\textbf{CIN}} & \texttt{\textbf{CON}} & \texttt{\textbf{SIM}}\\
\midrule
\textbf{Average} & 0.002 & 0.011 & 0.047 & 0.003 \\
\textbf{Std Dev} & 0.003 & 0.011 & 0.036 & 0.004 \\
\textbf{Median} & 0.000 & 0.008 & 0.053 & 0.001 \\
\Xhline{1pt}
\end{tabular}}
\caption{Difference in proportion of data points belonging to the minority class before and after label cleaning, averaged over noise type, noise rate, random seed(s), and downstream model architecture and datasets. Barring Confident Learning, the other cleaning methods do not have a major impact on class imbalance.}
\label{tab:prop_samples_cleaned_2}
\end{table}

\pagebreak
\section{Sources and Licenses}

All experimentation datacards to reproduce results can be found \href{https://drive.google.com/drive/folders/1RHczHDUUilTOhcPyF5JSDvkO-rhiUKgb?usp=sharing}{here}.

\begin{table}[h!]
\centering
\resizebox{0.85\textwidth}{!}{
\begin{tabular}{r c l c}
\Xhline{1pt}
\textbf{Cleaning Methods and Datasets} & \textbf{Reference} & \textbf{License} & \textbf{Source}\\
\midrule
SimiFeat & \citep{zhu2022detecting}  & \href{https://creativecommons.org/licenses/by-nc/4.0/}{CC BY-NC 4.0} & \href{https://github.com/UCSC-REAL/SimiFeat}{Link} \\

AUM & \citep{aum_ranking} & \href{https://en.wikipedia.org/wiki/MIT_License}{MIT} & \href{https://github.com/asappresearch/aum}{Link}\\

CINCER & \citep{cincer} & \href{https://en.wikipedia.org/wiki/MIT_License}{MIT} & \href{https://github.com/abonte/cincer}{Link} \\

Confident Learning & \citep{northcutt2021confident} & \href{https://www.gnu.org/licenses/agpl-3.0.en.html}{GNU AGPL v3.0} & \href{https://github.com/cleanlab/cleanlab}{Link}\\
\midrule

CIFAR-10N & \citep{cifar_10n} & \href{https://creativecommons.org/licenses/by-nc/4.0/}{CC BY-NC 4.0} & \href{http://www.noisylabels.com/}{Link} \\

CIFAR-10H & \citep{peterson2019human} & \href{https://creativecommons.org/licenses/by-nc-sa/4.0/}{CC BY-NC-SA 4.0} & \href{https://github.com/jcpeterson/cifar-10h}{Link} \\

Clothing-100K & \citep{Clothing1M, aum_ranking} & \begin{tabular}{@{}l@{}} Non-commercial research \\ and educational purposes \end{tabular} & \href{https://github.com/asappresearch/aum/tree/master/examples/paper_replication}{Link 1}, \href{https://github.com/Cysu/noisy_label}{Link 2}\footnotemark[7] \\

NoisyCXR & \citep{noisycxr} & \href{https://nihcc.app.box.com/v/ChestXray-NIHCC/file/249502714403}{Unrestricted use} \footnotemark[8] & \href{https://github.com/microsoft/InnerEye-DeepLearning/tree/1606729c7a16e1bfeb269694314212b6e2737939/InnerEye-DataQuality}{Link} \\

IMDb & \citep{IMDB} & \href{https://en.wikipedia.org/wiki/MIT_License}{MIT} & \href{https://www.kaggle.com/datasets/lakshmi25npathi/imdb-dataset-of-50k-movie-reviews}{Link} \\

TweetEval & \citep{tweeteval} & \href{https://en.wikipedia.org/wiki/MIT_License}{MIT} & \href{https://huggingface.co/datasets/tweet_eval}{Link} \\

Credit Card Fraud Detection & \citep{credit_card_dataset} & \href{https://opendatacommons.org/licenses/dbcl/1-0/}{DbCL v1.0} & \href{https://www.kaggle.com/datasets/mlg-ulb/creditcardfraud}{Link} \\

Adult & \citep{uci} & \href{https://creativecommons.org/licenses/by-nc/4.0/}{CC BY-NC 4.0} & \href{http://archive.ics.uci.edu/dataset/2/adult}{Link} \\

Dry Bean & \citep{dry_bean_dataset} & \href{https://creativecommons.org/licenses/by-nc/4.0/}{CC BY-NC 4.0} & \href{https://archive.ics.uci.edu/dataset/602/dry+bean+dataset}{Link} \\

Car Evaluation & \citep{hierarchical_decision_model} & \href{https://creativecommons.org/licenses/by-nc/4.0/}{CC BY-NC 4.0} & \href{https://archive.ics.uci.edu/dataset/19/car+evaluation}{Link} \\

Mushroom & \citep{mushroom_dataset} & \href{https://creativecommons.org/licenses/by-nc/4.0/}{CC BY-NC 4.0} & \href{https://archive.ics.uci.edu/dataset/73/mushroom}{Link} \\

COMPAS &\citep{COMPAS_dataset} & \href{https://opendatacommons.org/licenses/dbcl/1-0/}{DbCL v1.0} & \href{https://www.kaggle.com/datasets/danofer/compass}{Link} \\\

Crop & \citep{dynamic_time_warping} & \href{https://www.gnu.org/licenses/gpl-3.0.en.html}{GNU GPL v3.0} & \href{http://www.timeseriesclassification.com/description.php?Dataset=Crop}{Link} \\

ElectricDevices & \citep{electric_devices_dataset} & \href{https://www.gnu.org/licenses/gpl-3.0.en.html}{GNU GPL v3.0} & \href{https://timeseriesclassification.com/description.php?Dataset=ElectricDevices}{Link} \\

MIT-BIH & \citep{mit_bih} & \href{https://opendatacommons.org/licenses/by/1-0/}{ODC-By v1.0} & \href{https://www.physionet.org/content/mitdb/1.0.0/}{Link} \\

PenDigits & \citep{pendigits_dataset} & \href{https://creativecommons.org/licenses/by-nc/4.0/}{CC BY-NC 4.0} & \href{http://www.timeseriesclassification.com/description.php?Dataset=PenDigits}{Link}\\

WhaleCalls & \citep{UEA} & \begin{tabular}{@{}l@{}}Copyright \copyright \ 2011 by Cornell University \\ and Cornell Research Foundation, Inc. \footnotemark[9] \end{tabular} & \href{https://www.timeseriesclassification.com/description.php?Dataset=RightWhaleCalls}{Link}\\
\Xhline{1pt}
\end{tabular}}
\caption{Licenses for cleaning methods and datasets.}
\label{tab:license_clean}
\end{table}

\footnotetext[7]{Dataset can be downloaded by contacting \url{tong.xiao.work@gmail.com}}
\footnotetext[8]{We acknowledge the NIH Clinical Center (\url{clinicalcenter.nih.gov}) and National Library of Medicine \url{www.nlm.nih.gov}) for providing this dataset.} 
\footnotetext[9]{Data courtesy of and copyrighted by Cornell University and the Cornell Research Foundation.}

\end{document}